\documentclass{article}
\usepackage{fancyhdr} 
\usepackage[english]{babel}

\usepackage[letterpaper,top=2cm,bottom=2cm,left=3cm,right=3cm,marginparwidth=1.75cm]{geometry}

\usepackage{natbib} % A common package for better citation styles
\usepackage{amsmath}
\usepackage{amssymb}
\usepackage{booktabs} 
\usepackage{graphicx}
\usepackage[colorlinks=true, allcolors=blue]{hyperref}
\usepackage{float}

\title{Sharp Minima Can Generalize: A Loss Landscape Perspective On Data}
\author{
  Raymond Fan\textsuperscript{1}\\[-2pt]\small\texttt{raymond@aileap.org}
  \and
  Bryce Sandlund\textsuperscript{1}\\[-2pt]\small\texttt{bryce@aileap.org}
  \and
  Lin Myat Ko\textsuperscript{1}\\[-2pt]\small\texttt{light@aileap.org}
}

\date{%
  \vspace{0.6em}
  {\large % or \Large, \LARGE, etc.
    \textsuperscript{1}\ 
    \raisebox{-0.5ex}{\includegraphics[height=1.2em]{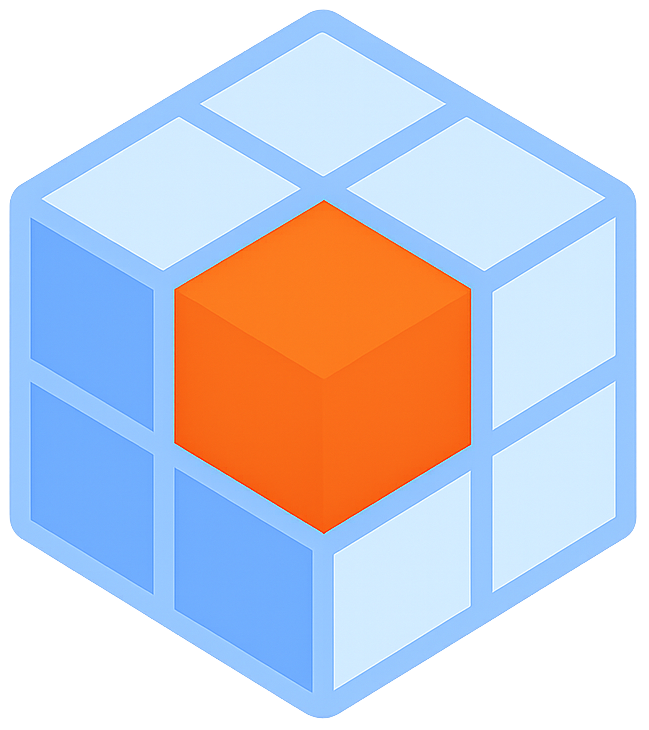}}\hspace{0.1em}
    AI Leap%
  }
}

\begin{document}
\maketitle

\begin{abstract}
The volume hypothesis suggests deep learning is effective because it is likely to find flat minima due to their large volumes, and flat minima generalize well. This picture does not explain the role of large datasets in generalization. Measuring minima volumes under varying amounts of training data reveals sharp minima which generalize well exist, but are unlikely to be found due to their small volumes. Increasing data changes the loss landscape, such that previously small generalizing minima become (relatively) large.
\end{abstract}

% Sharp Minima Can Generalize Better: How Data Affects Loss Landscapes
% The volume hypothesis suggests deep learning is effective because it is biased towards finding flat minima thanks to their large volumes, and flat minima generalize well. This picture does not explain why large datasets are necessary. Measuring the volumes of minima as we decrease dataset sizes reveals there exist sharp minima which are unlikely to be found by deep learning due to the bias to large volume minima. Increasing data changes the loss landscape, such that previously small generalizing minima become (relatively) large.

% A carefully chosen set of vectors displays a slice of the loss landscape for a given dataset, which shows minima generated by training on (A) all of MNIST, (B) 10\% of MNIST, and (C) 1\% of MNIST. When training on the entirety of the dataset, the minima obtained from training on smaller fractions of the dataset disappear. However, when trained on only 1\% of the dataset, all three minima appear viable. Despite this, the probability of finding A from gradient descent seems astronomically low. We study this phenomenon in this work.

\begin{figure}[h!]
    \centering
    % Include the combined figure
    \includegraphics[width=\textwidth]{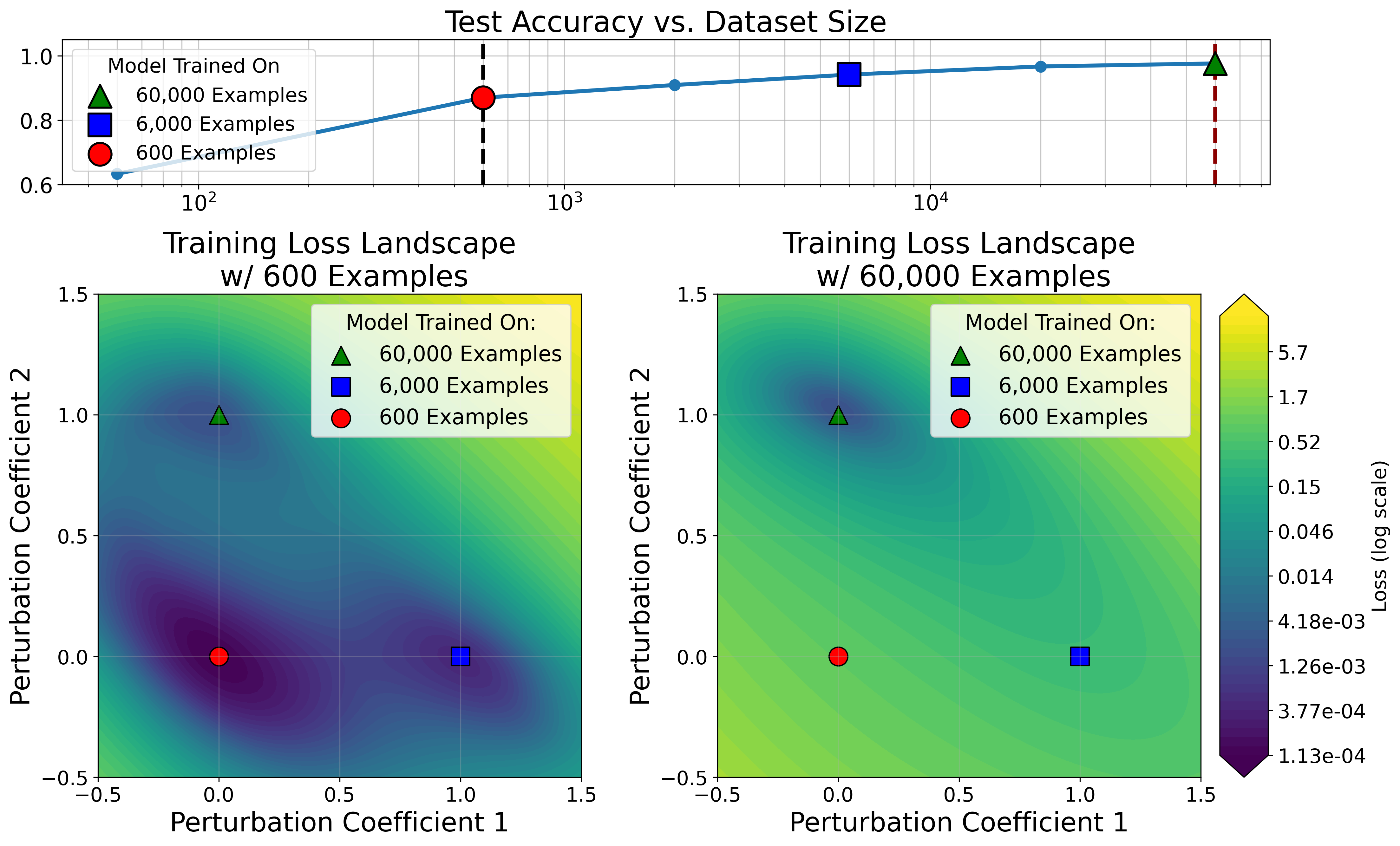}
    \caption{\textbf{Top: }Models trained on larger fractions of MNIST achieve higher test accuracy. \textbf{Bottom: }2D slices of the training loss landscape, containing models obtained by training on A: 100\% of MNIST (60,000 examples), B: 10\% of MNIST, and C: 1\% of MNIST. Only model A is a viable minima on the landscape from all training data (right). Yet in the 1\% training data landscape
    %But the loss landscape from 1\% of the dataset 
    (left---the same data we train C on) all three models appear viable minima. Training on this landscape however never yields minima like A or B.
    %Empirically, we never obtain models like A or B when training on such a small dataset.
    The volume hypothesis suggests this is because the volume associated with model C's minima is much larger than A or B.}    
    %On the bottom right landscape (using all training data), only model A is a viable minima. On the bottom left landscape (generated from 1\% of the dataset), all minima appear viable. However the probability of finding A from gradient descent seems astronomically low. The volume hypothesis states this is because the volume of C is vastly larger than A or B. Data then improves generalization by making generalization minima occupy the largest volumes. We study this phenomenon in this work.}
    \label{fig:combined_loss_landscape}
\end{figure}

\vfill
\begin{center}
\small Code is available at: \url{https://github.com/rfangit/minima-volume-project}
\end{center}
\vspace{2em}

\section{Introduction}

Generalization in neural networks refers to the phenomenon where trained models tend to perform well on unseen test sets. This is surprising since neural networks have enough parameters to fit any dataset, even to random labels or random noise~\cite{zhang_understanding_2017}. Existing guarantees on generalization require limiting model capacity, forcing low complexity solutions to fit the data~\cite{vapnik1998statistical, goos_rademacher_2001}; these bounds generally do not apply to overparameterized neural networks.

%Training on larger datasets is a well known method for producing models that generalize well, quantified by their performance on unseen test sets. %\cite{Do ImageNet Classifiers Generalize to ImageNet?, other papers by Ludwig Schmidt}
%These generalizing models also minimize the training loss on smaller datasets, but we never obtain them from training on those datasets. Why is more data needed to generalize?

%How does more data improve generalization?

%To answer this question, we first need to understand why models generalize at all. Existing notions like the bias-variance tradeoff, describe limiting model capacity so they can only match the training data by adopting a low complexity approximation to the true data. Such bounds do not apply to neural networks, which have enough parameters to fit to any dataset, even to random labels or random noise. %\cite{Random Noise Samy Bengio Paper}

This has inspired the volume hypothesis, which proposes generalization arises from the loss landscape. The hypothesis states the volume of parameter space occupied by generalization minima is significantly larger than the volumes of other minima, and thus any procedure that minimizes the training loss %(particularly if the minima found correlates to the \textit{volume} of said minima, believed to be the case in SGD) 
is likely to find good minima. The link between volume and generalization lies in the flat minima hypothesis, which argues minima which are flat in parameter space (and thus larger \textit{volume})  should generalize well~\cite{Hochreiter1997FlatM,hinton_keeping_1993,keskar_large-batch_2017}.
%correspond to robust representations that should generalize well.%\cite{Tons of citations for this!}

% Flatness in parameter space imply the network has learned robust features that cleanly separate the dataset outputs by a wide margin, meaning

% minima which are flat in parameter space should generalize well. %\cite{Tons of citations for this!}

The volume hypothesis has been tested, where experiments show the volume of minima obtained by training on a `poisoned' dataset (containing extra samples with incorrect labels) are significantly smaller than the volumes of minima obtained on the base dataset~\cite{huang_understanding_2020}. The hypothesis also implies minima found by randomly guessed parameters will generalize similarly to those found by gradient descent, which has been verified on small variants of MNIST~\cite{chiangLOSSLANDSCAPESARE2023, peleg_bias_2025}.

A theory of generalization also needs to explain why we only get the level of generalization observed and not better. In this work, we focus on the role of data: empirically, training on larger datasets results in models with better generalization. Why do we never obtain these models when training on smaller datasets?

The volume hypothesis suggests this is because the minima from training on a small dataset has the largest volume \textit{in the loss landscape generated by the small dataset}. These minima are counterexamples to the flat minima hypothesis. %Increased data changes the loss landscape and results in different minima being larger. We test this explanation by training models in a variety of dataset sizes and evaluating how their volumes change in different datasets.
Increasing dataset size changes the loss landscape and thus which minima are large. We test this explanation by training models in a variety of dataset sizes, evaluating how their volumes change.

\section{Background}

Generalization in neural networks can be explained via loss landscapes with two assumptions:

%A loss landscape perspective on generalization in deep neural networks postulates the effectiveness of deep learning is due to the following two facts:
%The volume hypothesis suggests that the effectiveness of deep learning is because:

\begin{enumerate}
    \item \textbf{Flat Minima Hypothesis: }Minima whose parameters can be perturbed without significantly increasing training loss generalize better.
    \item \textbf{Volume Hypothesis: }Gradient descent is biased towards finding minima with large volumes, which correspond to flat minima.
\end{enumerate}

Together, these imply gradient descent is likely to find minima that generalize well. But this explanation is incomplete, since it ignores deep learning is only effective on large datasets.

To understand the role of data, we examine each assumption in detail.

%Data effects the loss landscapes, which affects both the flatness of different minima and also the minima found by gradient descent. To understand where it's effect can manifest, we examine the two listed assumptions in more detail.

%To understand how data is essential, we first examine the two assumptions listed.

\subsection{Flat Minima Hypothesis}
% , for some definition of flatness,
The idea that flat minima yield better generalization has been explored extensively~\cite{Hochreiter1997FlatM, keskar_large-batch_2017, jiang_fantastic_2019, stutz_relating_2021, ibayashi_why_nodate}.
Empirical studies comparing the sharpness and generalization from minima found with different optimization methods (e.g., varying batch sizes~\cite{keskar_large-batch_2017}) support this connection. Here we mention two theoretical explanations.

First, from the viewpoint of model compression, models of low Kolmogorov complexity are expected to generalize better~\cite{wilson_deep_2025}. A flat minimum can be stored with lower numerical precision without loss of performance, implying lower description length and thus complexity~\cite{rissanen_universal_1983, hinton_keeping_1993}. %Related works discuss the minimum description length, Occam's razor and Solomonoff induction. %relation to complexity, kolmogorov, mdl. Get citations for this

Second, flat minima enforce a wide margin criterion~\cite{huang_understanding_2020}. In classification tasks, model parameter perturbations will shift decision boundaries. 
For a flat minimum, such perturbations do not alter the classification of nearby data points. This implies the model learns robust classifications in a way not captured by the loss.

%They are more confident about their predictions in a way not captured directly by the loss. % Colloquially, this means our model is highly confident in its predictions in a way not fully captured by the loss.

But for many measures of flatness, one can find arbitrarily sharp minima thanks to the scale invariance of neural networks~\cite{dinh_sharp_2017}. A potential remedy to this problem is to consider layer-wise normalization methods~\cite{li_visualizing_2018, stutz_relating_2021}.

Outside of issues with defining a flatness metric, both experimental and theoretical work has established sharp minima can generalize well~\cite{wen_sharpness_nodate}, and there exist experimental results asserting flatness appears negatively correlated with generalization~\cite{andriushchenko_modern_2023}. 
It seems plausible that flat minima in low data settings may not generalize well.

%    \item Gradient descent is highly biased towards finding minima with large volumes, which are flat. (volume hypothesis)

\subsection{Volume Hypothesis}

The volume hypothesis builds on the observation that a minima twice as flat along every parameter occupies a volume $2^N$ times as large. Flat minima thus can occupy much more volume in high-dimensional parameter spaces. Any method of optimizing a large neural network is more likely to find flat minima due to volume \cite{huang_understanding_2020}. This may explain why neural networks with more parameters tend to generalize better, as the volumes give a soft inductive bias towards simple solutions \cite{wilson_deep_2025}.

% ChatGPT wrote this - I feel like it's better....
%Empirical evidence supports this view. Minima found by training on large datasets containing incorrectly labeled examples tend to occupy significantly smaller volumes than those found on only the smaller correctly labeled dataset~\cite{huang_understanding_2020, scherlis_estimating_2025}. Likewise, minima obtained by alternative optimization methods (e.g., randomly chosen parameters that happen to minimize the test loss) generalize similarly well to those found by gradient descent, consistent with a volume-based bias~\cite{chiangLOSSLANDSCAPESARE2023, peleg_bias_2025}.

The hypothesis suggests minima found from training are likely to have larger volumes than minima not naturally obtained. This has been tested by measuring volumes of minima from training on poisoned datasets of incorrect labels. Poisoned minima had significantly smaller volumes, suggesting the hypothesis is a plausible explanation for why these minima which generalize extremely poorly are not naturally found \cite{huang_understanding_2020, scherlis_estimating_2025}.

The hypothesis also predicts that minima found via other optimization techniques (e.g., randomly chosen parameters that happen to minimize the test loss) will behave similarly to those found from gradient descent, which has been confirmed in experiments with small versions of MNIST \cite{chiangLOSSLANDSCAPESARE2023, peleg_bias_2025}.

Currently, there is no comparison with minima produced by larger datasets of correct labels. Note that the volume hypothesis does not rule out the possibility that there exist flatter minima not found from gradient descent, since the individual volumes of minima need not be particularly large if there are many of them.

%However, the volumes of minima produced by larger datasets of correct labels have not been compared. 

%It is possible these minima to be larger despite being not found by gradient descent. 
%We note that this is entirely possible even under the volume hypothesis's implication that gradient descent is largely biased towards large volumes, since the individual volumes of minima need not be particularly large if there are many of them. % This will need rewrites

\subsection{Generalization With Data}

Our results show either hypotheses can fail in low data contexts, suggesting a more nuanced picture:

\begin{enumerate}
    \item For a given loss landscape, flat minima generalize better than most sharp minima.
    \item Gradient descent is modestly biased towards minima occupying larger volume in a given landscape.%, though seldom (if ever) finds the largest.
    \item There exist sharp minima which generalize better than the flattest minima, but are unlikely to be found by gradient descent. Increasing dataset size reshapes the loss landscape, making previously sharp minima become (relatively) large minima which can be found via gradient descent.
\end{enumerate}

In this view, finding minima which generalize extremely well is like searching for `needles in a haystack'. Desirable generalization minima are sharp, and there exist many other poor sharp minima. Our training algorithms avoid this issue by only selecting flat minima (which generalize well but not exceptionally so) in the current loss landscape.

Larger datasets change the loss landscape and remove the current flattest minima, forcing our algorithms to search for the new `flat' minima.

\section{Key Results}

Our key results are:

\begin{itemize}
%\begin{enumerate}
    \item We extend previous work on measuring minima volumes to the case of minima obtained from training on larger datasets. For tasks like MNIST and CIFAR10, gradient descent shows a strong bias towards large volume minima at all dataset sizes. This supports the idea volumes are useful in understanding generalization.
    \item We find cases in swiss roll classification and modulo arithmetic where the volume of minima found are smaller than minima from training on larger datasets. We note a bias towards volume could be hidden by the fact that there are many small minima instead of a few larger minima, which is partially supported by the relatively small differences in volumes and the large variability in volume estimation.
    %and this is supported by the small differences in volume. These examples 
    %This is supported by the relatively small differences in volume and the large variation in volumes from random model initializations.
    \item We show the flat minima hypothesis does not explain the improvement in generalization from large datasets. Minima found from larger datasets are actually `sharp' minima in smaller datasets. The trend between flatness and generalization seems to depend on how minima are acquired (e.g., larger batch sizes in earlier work versus larger datasets here). 
    \item We find minima volumes almost always shrink with increased dataset sizes. Improved generalization occurs because previously-flat minima shrink faster in volume, so the previously-sharp minima are now relatively flat. The shrinkage is more severe when adding poisoned data, which suggests an intuitive picture where there exist many sharp minima that generalize poorly, and a few sharp minima which generalize well.% and bad minima, and a few sharp but good minima.
    \item We find an unexpected power-law relationship between minima volume and dataset size in MNIST, CIFAR10 and other image classification tasks that holds across three orders of magnitude.
    \item We show that from a volume perspective, the grokking phenomenon observed in modulo arithmetic involves a steady progression from a flat minima to a sharp minima. 
\end{itemize}

\section{Minima Volumes and Estimation}

The volume hypothesis suggests the probability of obtaining a minima via gradient descent with the observed amount of test loss is similar to the probability of obtaining such minima with randomly chosen parameters. This probability is directly proportional to the volume of the minima.
%The probability of a minima from randomly chosen parameters is directly proportional to its volume. 
%The high probability of 
Generalization in deep learning %then
follows from the existence of flat generalizing minima which occupy significantly higher volumes.

Here, volume refers to the size in parameter space of a region with sufficiently low training loss. The choice of threshold is arbitrary, but our results are generally robust to different threshold choices, see Appendix~\ref{app:robustness}. % worth mentioning issues here with the scale invariance and other strange facts with how volumes are well connected?

Note that gradient descent is known to have additional implicit biases that favor generalization~\cite{arora_implicit_nodate, galanti_sgd_nodate}. But prior experiments with the volume hypothesis suggest non-gradient biases can explain much of generalization, which we focus on in this work. %We only assert the ability of non-gradient biases 

%the volume hypothesis remains valuable if it explains a substantial component of this bias.

%Note there is evidence that gradient descent has biases that help generalization \cite{arora_implicit_nodate, galanti_sgd_nodate}, which imply it is more likely to find good minima than the volume hypothesis. But the volume hypothesis is still useful as long as it explains a significant part of generalization.

The feasibility of the volume hypothesis can be tested by measuring the volumes of a variety of minima.
%To test the feasibility of this hypothesis, one can measure the volumes of a variety of minima.
%One way to test the feasibility of this hypothesis is to measure the volumes of a variety of minima. 
If the volume of minima found by gradient descent are significantly larger than other candidate minima, that supports the idea that volumes explain much of generalization.

\begin{figure}[h!]
    \centering
    % Include the combined figure
    \includegraphics[width=0.95\textwidth]{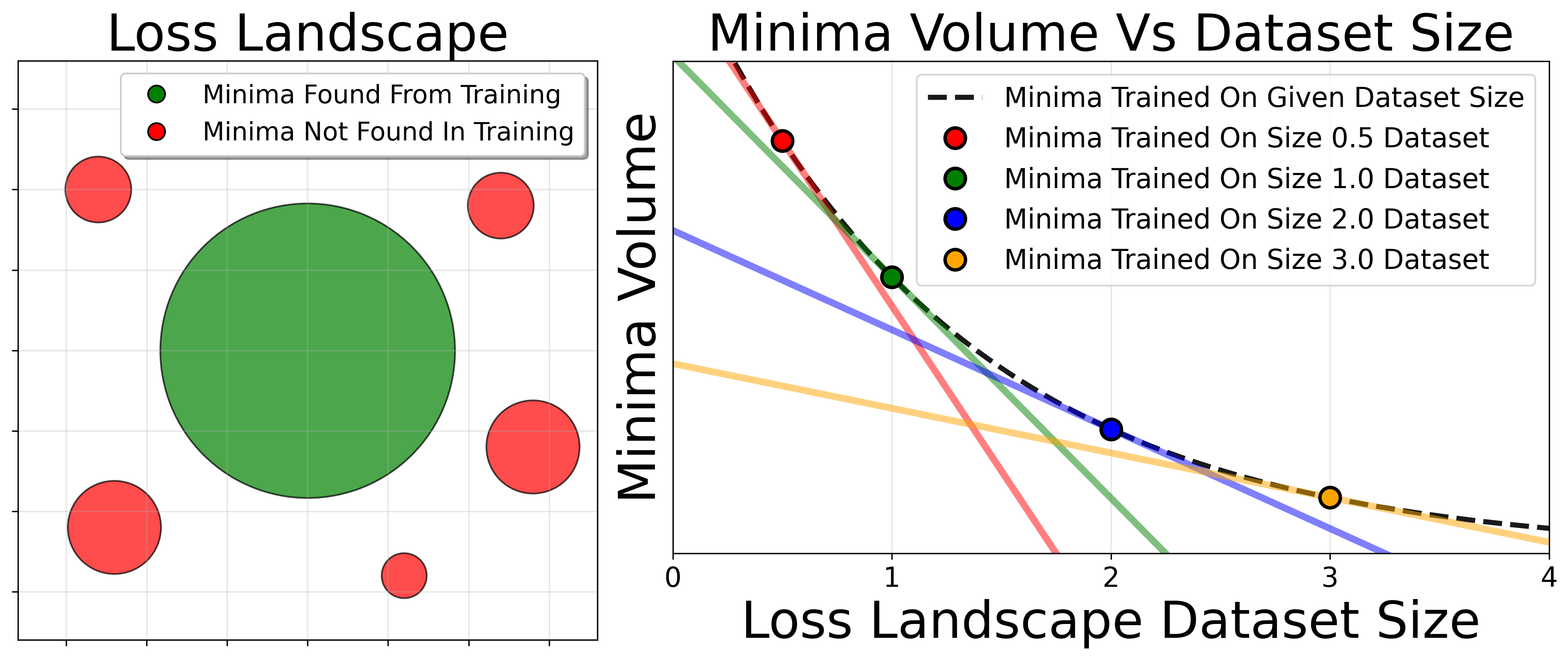}
    \caption{\textbf{Left: }Cartoon for a scenario where minima found on a dataset is larger than \textit{all other} minima. Previous experiments observed this case when comparing to minima from a poisoned dataset~\cite{huang_understanding_2020, scherlis_estimating_2025}. \textbf{Right: }Minima found at a given dataset size (e.g., green dot) is larger than minima found at other sizes (blue and yellow curves). Note the hypothesis does not predict how absolute minima volume scales with dataset size. If there are simple scaling relations like the lines shown here, one could imagine special algorithms targeting minima which generalize well via their scaling properties.}
    \label{fig:strong_volume_hypothesis}
\end{figure}
    %the hypotheses have predictions when plotting minima volume against dataset size and indicating where each minima was obtained (dots) results in a minima volume vs dataset size graph.    
    %If other candidate minima are minima obtained from training on other dataset sizes, then 
    %The minima found at a given dataset must always be the larger than the minima which were found at larger or smaller datasets. (right) The hypothesis has no prediction for how minima volumes behave in datasets other than the ones they were trained in, but if they obey some simple scaling with dataset size as shown, one could develop algorithms to search for minima which generalize better based off volume.
    
\begin{figure}[h!]
    \centering
    % Include the combined figure
    \includegraphics[width=0.95\textwidth]{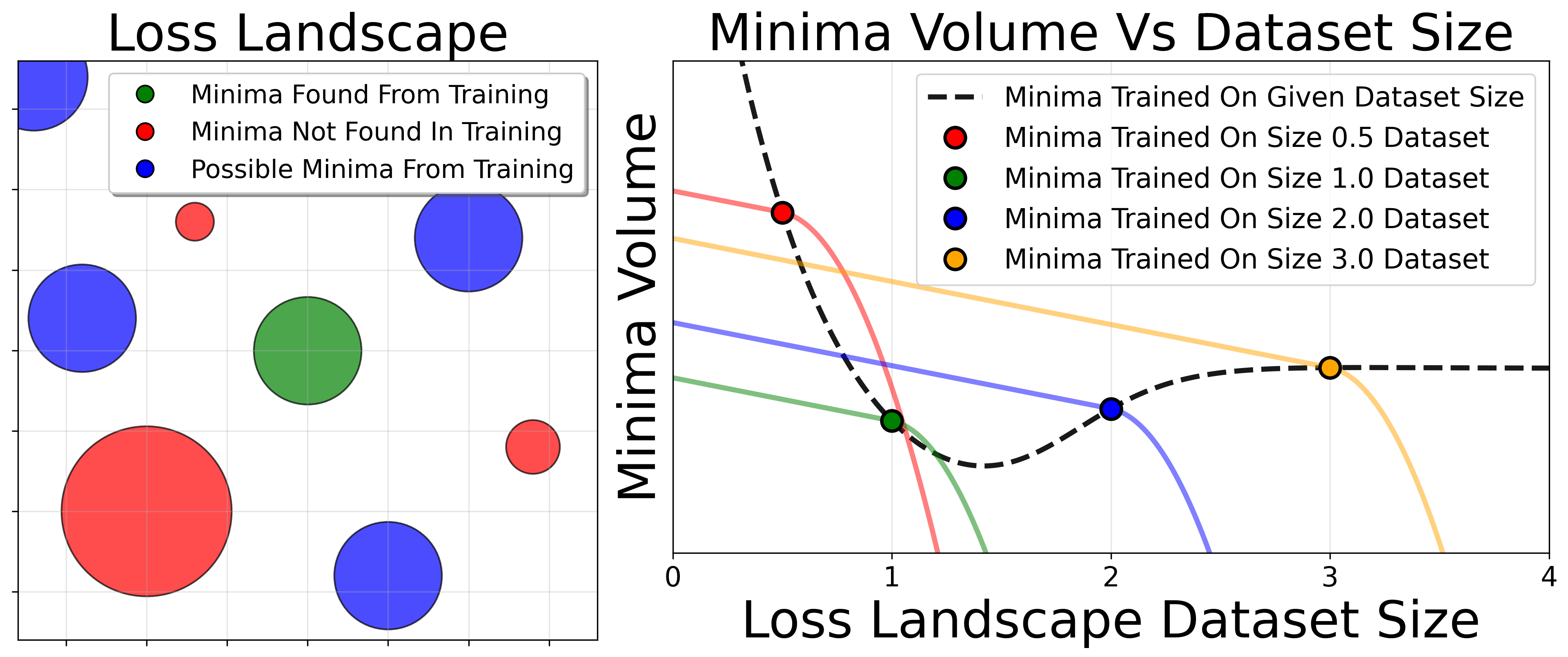}
    \caption{\textbf{Left:} Illustrating how generalization can be described by volume dynamics even if the minima found is smaller than other candidate minima. The found minima belongs to a popular class of minima with similar test loss, such that the volume of the class is largest overall. \textbf{Right:} Minima found at a given dataset size (e.g., green dot) are smaller than minima obtained at larger dataset sizes (blue and yellow curves). Volume-data scaling for individual minima here are inspired by our experimental results.} %maybe the colors here need to be swapped, don't have a blue minima but a purple instead?
    %Strong volume hypothesis implies the one found is largest, and implies the distribution of volumes is on the right. Bottom: Weak volume hypothesis.}    
    %On the bottom right landscape (using all training data), only model A is a viable minima. On the bottom left landscape (generated from 1\% of the dataset), all minima appear viable. However the probability of finding A from gradient descent seems astronomically low. The volume hypothesis states this is because the volume of C is vastly larger than A or B. Data then improves generalization by making generalization minima occupy the largest volumes. We study this phenomenon in this work.}
    \label{fig:weak_volume_hypothesis}
\end{figure}

Previous experiments compared a minima found with gradient descent on a dataset A with a variety of `poisoned' minima, obtained from training on A + P where P is a poisoned dataset containing new samples with incorrect labels \cite{huang_understanding_2020, scherlis_estimating_2025}. Poisoned minima are consistent with the base training dataset, but generalize poorly and have significantly smaller volumes. The volume hypothesis can thus explain why such minima never occur in practice.

This study examines minima obtained by training on A + B, where B is additional correctly labeled data. Unlike the poisoning experiments, we are interested in the behavior of minima as dataset sizes increase. The volume hypothesis suggests minima change in volume when landscapes change such that the minima found in the new landscape are largest, see Fig~\ref{fig:strong_volume_hypothesis}.

Note the volume hypothesis can be true even if this is not the case. The hypothesis argues that generalization can be mostly explained by volume; however, the volumes of individual minima can be small as long as there are many distinct minima with similar levels of generalization, see Fig~\ref{fig:weak_volume_hypothesis}.

Minima volumes as a function of dataset size is also interesting. Intuitively, we expect the volumes of minima to shrink with more data, since variations in parameters that previously left the loss unchanged may be heavily penalized by new data. But this is not necessarily true, except in trivial cases where the loss is not normalized by sample count.

If interesting relationships between volume and dataset size exist---as illustrated in Fig~\ref{fig:strong_volume_hypothesis}---then one could design specialized algorithms to find minima that generalize better. However, useful scaling relationships are not guaranteed, see Fig~\ref{fig:weak_volume_hypothesis}. Our results suggest the latter case.

\subsection{Basin Volume Estimation}

Previously, Huang et al.~\cite{huang_understanding_2020} used Monte Carlo basin volume estimation to measure the volume of minima. %We discuss the method here.

%Past work has used a Monte Carlo basin volume estimation to measure the volume of minima. We discuss the approach here and possible issues that we address empirically in our experiments.

%Measuring the volume of minima is difficult due to the high dimension of neural network parameter space. Past work has used a Monte Carlo basin volume estimation technique. We discuss the method here and describe several potential issues which we address empirically with our experiments.
The method measures the (star-convex) basin volume of a minima---the region of linearly-connected parameter space below an arbitrary loss threshold. For any random direction $\vec{\theta}$ in parameter space, one can find a distance $r(\vec\theta)$ from the minima to the edge of the basin. An $n$-dimensional basin volume is given by

\begin{equation}
V = V_{\text{unit ball}, n} \times \int r^n(\vec\theta) \, d\vec\theta,
\end{equation}
where $V_{\text{unit ball}, n}$ is the volume of the unit ball in n dimensions, and the integral is over all directions in $n$ dimensions \cite{scherlis_estimating_2025}. This integral can be approximated by the expected value from $K$ random directions

\begin{equation}
\int r^n(\vec\theta) \, d\vec\theta = \mathbb{E} [ r^n(\vec\theta) ] \approx \frac{1}{K} \sum_{i=1}^K r^n(\vec\theta_i).
\end{equation}

Using this approach, Huang et al. tested the volume hypothesis for swiss roll classification and the street view house numbers (SVHN) dataset. They found poisoned minima have significantly smaller volumes, which could explain why they never occur in practice.

%They generated minima consistent with the training dataset but with poor generalization by creating additional data with incorrect labels, and compared their volumes to the volumes of minima obtained normally via the training data. They found large differences in the volume that could explain why poisoned minima never occur in practice.

\begin{figure}[h!]
    \centering
    % Include the combined figure
    \includegraphics[width=0.42\textwidth]{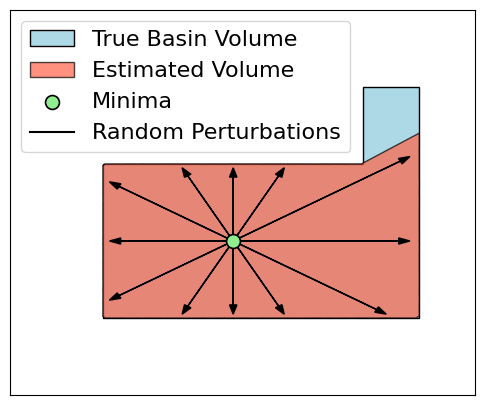}\hspace{0.03\textwidth}
    \includegraphics[width=0.42\textwidth]{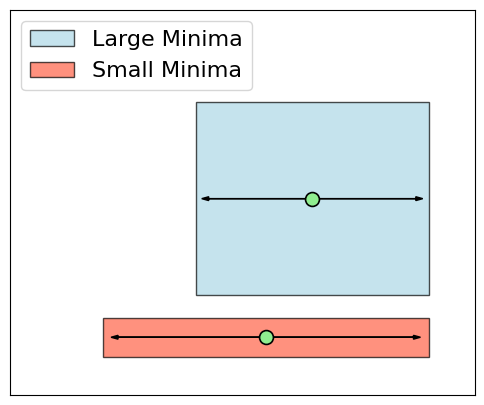}
    \caption{\textbf{Left: }Monte Carlo basin volume estimation measures distances to the basin boundary along random directions, yielding a star-convex estimate (red) that underestimates the true volume (blue).
    %Monte Carlo basin volume estimation involves moving in random directions and tracking the distances to the edge of the basin. Repeating this many times only allows you to estimate the volume of a convex basin (red), which is always an underestimate of the true basin volume (blue).
    \textbf{Right: }Random directions can align with flat axes in a small basin, making it appear larger than other minima. This issue depends on the shape of the minima and is exacerbated by high dimensionality.
    %Larger minima can be incorrectly identified as smaller if the randomly chosen vectors happen to align with a flat axis in a small minima. The severity of this problem will depend on the shapes of minima and is exacerbated by the extremely high-dimensional nature of the Monte Carlo integral.
    } 
    \label{fig:basin_volume_estimation}
\end{figure}

\begin{figure}[h!]
    \centering
    % Include the combined figure
    \includegraphics[width=0.95\textwidth]{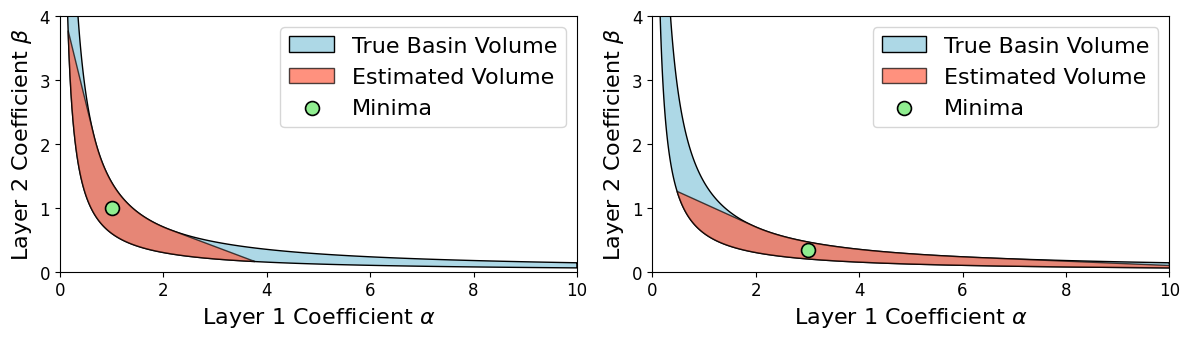}
    \caption{Multiplying one layer by a factor $\alpha$ and the following layer by $\beta = 1/\alpha$ leaves a neural network unchanged. This scale invariance is problematic for flatness measures as they may not return the same values for the minima on the left as the right, despite being identical models. Note this invariance also implies minima volumes are infinite. Interestingly, for the two plots above, star convex basin volume is actually identical, which suggests it may be more robust than other flatness measures. (See Appendix~\ref{app:analytical_basin_volume})} 
    \label{fig:scale_invariance}
\end{figure}

This approach has several shortcomings:

\begin{itemize}
    \item \textbf{Star-Convex Basin Volume:} It computes the star-convex basin volume of minima, or the region reachable via straight lines from the basin center. This is always an underestimate of the true minima volumes, which are often infinite, see Fig~\ref{fig:basin_volume_estimation}.
    \item \textbf{High Dimension Scaling:} In high dimensions, the Monte Carlo integral is dominated by the largest radius found. A small minima might appear larger than another minima because a randomly-chosen vector happens to align with a rare but flat direction, see Fig~\ref{fig:basin_volume_estimation}.
    \item \textbf{Scale Invariance:} Multiplying the weights of a layer by $\alpha$ and the following layer by $1/\alpha$ results in an identical network. But there is no guarantee that basin volumes are unchanged. This is an issue for most measures of flatness \cite{dinh_sharp_2017}, see Fig~\ref{fig:scale_invariance}.
\end{itemize}

%Furthermore, neural network parameter space is scale invariant: multiplying the weights of a layer by $\alpha$ and the following layer by $1/\alpha$, results in an identical network, but there is no guarantee that basin volumes are unchanged. This is a problem with most measures of network flatness. Note this scale invariance also implies the volumes of minima are infinite, whereas basin volume estimates are always finite.

%Finally, for extremely high dimensions (such as those in neural networks) the Monte Carlo method is dominated by the largest radius found. One can imagine cases where a smaller minima appears larger than an actually larger minima because a randomly chosen vector happens to align with a rare but flat direction. %The basin volume integral is dominated by these outliers, and there is no guarantee that more sampling results in more accurate results.

To mitigate these flaws, Scherlis et al.~\cite{scherlis_estimating_2025} proposed refined sampling strategies and volume measures related to the probability density of the initialization distribution, which avoids the infinite-volume issue.

%To address these flaws, Scherlis et al.~\cite{scherlis_estimating_2025} discuss methods for choosing random vectors, and more reasonable volume measures (considering the probability of trained model parameters in terms of the initialization distribution, which removes the infinite volume issue).

Here, we point out basin volume estimation can be useful despite the previous shortcomings:

\begin{itemize}
    \item \textbf{Minima volumes can differ dramatically: } Numerical issues with basin volume are irrelevant in practice if differences in minima volumes are larger than the errors. This appears to be true for the minima studied in our experiments.
    %It's possible that the shape and size of minima of interest are so different that numerical errors associated with basin volume are irrelevant in practice. Empirically this appears to be the case, enabling us use basin volume regardless of it's theoretical issues.
    \item \textbf{Easy to compute and robust: }%Unlike other measures of flatness, 
    Basin volume estimation only require model perturbations and additional forward passes. % without requiring second order derivatives.
    It also appears more robust to the scale invariance issues with other flatness measures. For an analytical example where basin volumes are invariant to scale, see Appendix~\ref{app:analytical_basin_volume}. %of basin volume invariance to scale in Appendix XXX. 
    %where it remains invariant to scale invariance in Appendix XXX. 
    \item \textbf{Local volumes may be more meaningful: } True volume may not be very useful given that real basin volumes are possibly infinite and exhibit varying test loss across the basin (e.g. mode connectivity \cite{garipov_loss_2018}). In grokking, for example, the test loss varies dramatically along a trajectory of low train loss \cite{power_grokking_2022}. % add citation
    A local measure of volume may be more useful than a global measure.
\end{itemize}

\section{Minima Volume Results}

We measure basin volumes for MNIST, CIFAR10, swiss roll classification, and modulo arithmetic (base 97). For SVHN and Fashion MNIST, see Appendix~\ref{app:svhn_fmnist}. Our models are trained with AdamW with Pytorch defaults, though similar results occur for stochastic gradient descent (SGD) (Appendix~\ref{app:model_variations}) and sharpness-aware minimization (SAM).
% Needto describe these models

To reduce the influence of random initialization, we vary both model parameter seeds and dataset split seeds. Basin volume is defined as the region of parameter space where the training loss is below 0.1, except for modulo arithmetic, where typical loss values are much smaller, and a threshold of 0.01 is used instead.

%We choose the region where the loss is below 0.1 (an arbitrary value) as our basin, except for modulo arithmetic where standard loss values are much lower (with threshold 0.01 instead).

Due to the high number of parameters (and thus dimensions), volumes span many orders of magnitude. In our experiments, we focus not on absolute value (which almost always increases with more perturbation directions in our Monte Carlo estimate~\cite{scherlis_estimating_2025}) but instead on the relative sizes of different minima for a fixed network architecture. 

%To make our figures easier to follow, we also fix the loss landscape in all of our graphs, despite the found minimas coming from models trained on different datasets.

Each experiment uses %at least 
500 random perturbation directions (Monte Carlo samples) to ensure reliability. A volume measurement means 50,000 forward passes over the dataset with our chosen resolution. A perturbation direction involves a normally distributed random value for each parameter (weights and biases), scaled by the layer-wise filter norms (a standard practice to correct for possible invariances of the network  \cite{li_visualizing_2018}). For details, see  Appendix~\ref{app:filter_normalization}.

Aside from our experiments focused on datasets and the volume hypothesis, we also investigated variability in volumes found on a fixed dataset. Here, we visualized how much variability in minima volume can be generated purely by random seed initialization or batch sizes in training. We also found minima volume correlates with test accuracy when training with different batch sizes, similar to results by Keskar et al \cite{keskar_large-batch_2017}. These experiments are present in Appendix~\ref{app:batch and model seed}.

%To ensure our results are reliable, all experiments are reported for at least 500 random perturbation directions. A perturbation direction involves a normally distributed random value for each parameter (weights and biases), scaled by the layer-wise filter norms (a standard standard practice to correct for possible invariances of the network  \cite{li_visualizing_2018}). For details, see Appendix XXX.

%method to handle possible invariances of the network such as the scale invariance between layers \cite{li_visualizing_2018}). For details, see Appendix XXX.

%Each perturbation means generating a normally distributed random variable for each model parameter (weights or biases), multiplied by the layer-wise filter norms (a standard method to handle possible invariances of the network such as the scale invariance between layers). See the appendix for the details. Additionally, plots with less perturbations are also in the appendix, and show similar trends.

Our experimental trends are also observed when using as few as 50 perturbation directions (Appendix~\ref{app:robustness}). The minima in our experiments appear well-behaved enough in shape such that it is easy to determine their relative sizes.

%The trends observed in our experiments also appear when considering only 50 random perturbations (see Appendix XXX). This suggests the minima in our experiments are well behaved in shape such that it is easy to determine their relative sizes.

%Additionally, in the appendix are plots produced using only the first 50 and first 10 random perturbations. We find the same trends observed in 500 perturbations are found with only 50 or 10 perturbations%, with exceptions for .... (swiss roll?), 
%although the absolute values of the volumes are smaller. This suggests that the minima considered in our experiments are relatively easy to distinguish, with differences in volume becoming apparent very easily.

\subsection{Poisoned Minima Are Always Smaller}

\begin{figure}[h!]
    \centering
    % Include the combined figure
    \includegraphics[width=0.48\textwidth]{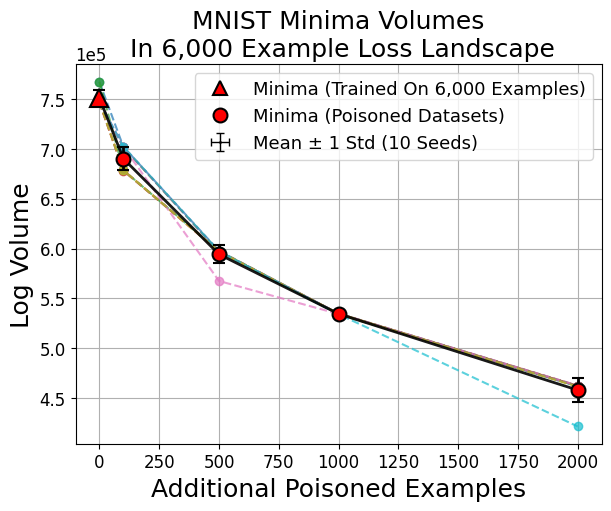}%
    \includegraphics[width=0.48\textwidth]{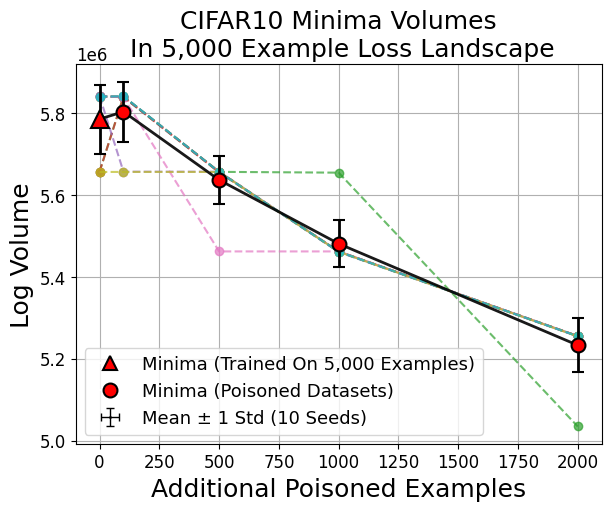}
    \caption{We extend the poisoning procedure to MNIST and CIFAR10. Adding incorrectly labeled examples to a subset of MNIST and CIFAR10 (6,000 and 5,000 examples, respectively) generates poisoned minima that minimize the training loss on the base dataset. 
    Poisoned minima typically have significantly less volume than the minima found normally. A minor exception occurs for small amounts of poisoning on CIFAR10 for some of our model/data seeds. This can be attributed to low test accuracy of CIFAR10 and numerical error from basin volume estimation. Identical volumes across seeds are likely the result of insufficient resolution in our measurement procedure.
    %Poisoned minima have significantly less volume than the minima found normally, except for CIFAR10 with small amounts of poisoning. This may be explained by the low test accuracy of CIFAR10 and numerical error from basin volume estimation. Identical volumes across seeds are likely the result of insufficient resolution in our measurement procedure.
    % I should probably address the fact that CIFAR actually doesn't have less volume with small amounts, probably because CIFAR is... so hard. Maybe the poisoned points are not enough.
    } 
    \label{fig:MNIST CIFAR poison}
\end{figure}

First, we extend the poisoning experiments of Huang et al.~\cite{huang_understanding_2020} to MNIST and CIFAR10 (Fig~\ref{fig:MNIST CIFAR poison}). %, and Fashion MNIST (Appendix XXX). 
In our experiments, minima from training on datasets with additional incorrectly labeled samples are significantly smaller in volume. Even very small amounts of additional poisoned data ($\approx 3\%$ of the original training dataset) result in noticeably smaller minima.
%These differences are dramatic enough that by poisoning only $\approx 3\%$ of the training dataset, the resulting minima volumes can be reliably distinguished from minima found without poisoning with near-100\% accuracy. 
The volume hypothesis is thus a plausible explanation for why the poorly-behaved `poisoned' minima do not appear in practice.

\begin{figure}[h!]
    \centering
    % Include the combined figure
    \includegraphics[width=0.48\textwidth]{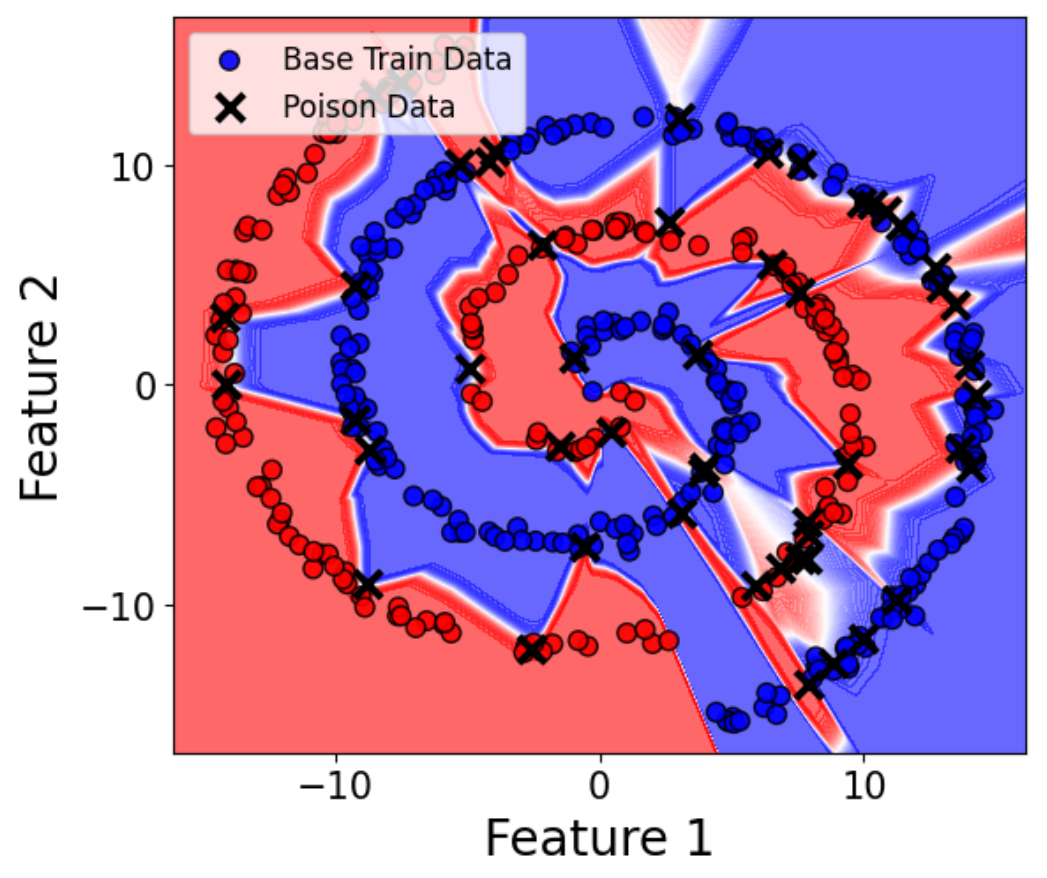}%
    \includegraphics[width=0.48\textwidth]{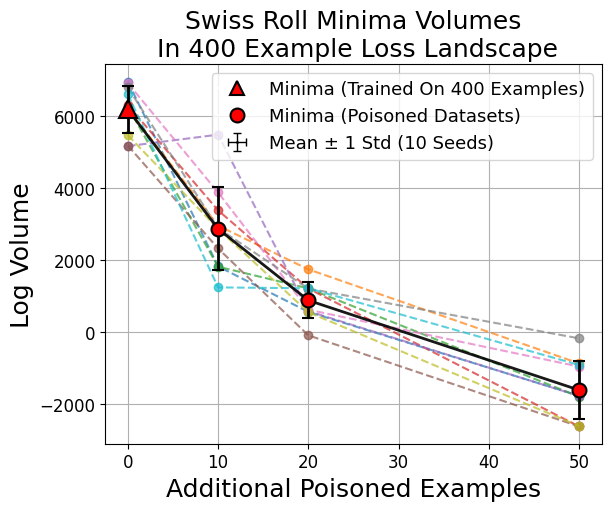}
    \caption{\textbf{Left:} A swiss roll classification problem is augmented with additional poisoned points (black `X's). `Poisoned' minima from training on this dataset fit the base training dataset but have poor generalization. \textbf{Right:} The volumes of poisoned minima (red circles) are smaller than the volume of the minima obtained by training on the base dataset (red triangle). Red points correspond to averages of experiments with different data generation and model initialization seeds purely to show overall trends.} 
    \label{fig:poison minima orig}
\end{figure}

\subsection{Small Datasets Always Find Large, Flat, Non-Generalizing Minima}

Our experiments show minima found in small datasets are significantly larger than minima found in larger datasets, in the training loss landscape of the small dataset. These minima generalize extremely poorly. In contrast, smaller minima from training on the larger datasets achieve vastly better generalization. The volume hypothesis seems to explain why these minima are not found, since their volumes are very small in comparison.

%This seems to correspond to the regime where neural networks are said to memorize their training data, with poor generalization.

\begin{figure}[h!]
    \centering
    % Include the combined figure
    \includegraphics[width=0.48\textwidth]{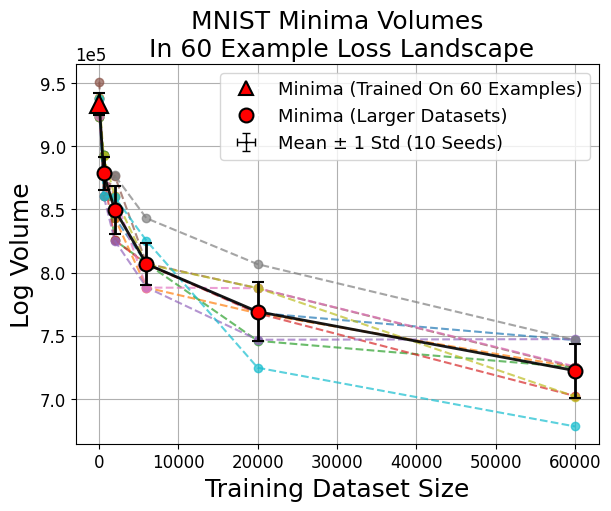}%
    \includegraphics[width=0.48\textwidth]{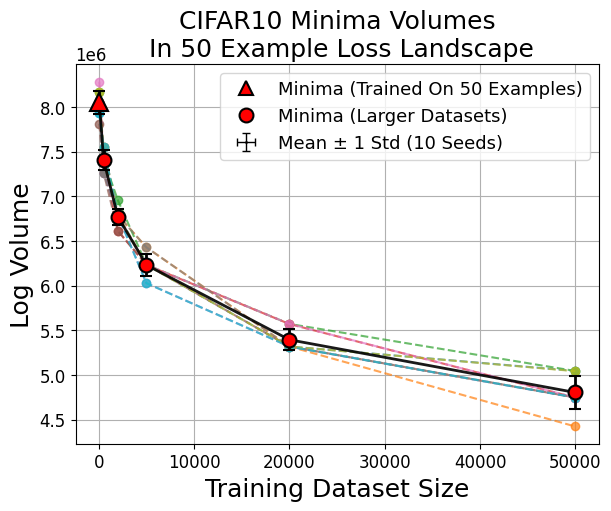}
    \includegraphics[width=0.48\textwidth]{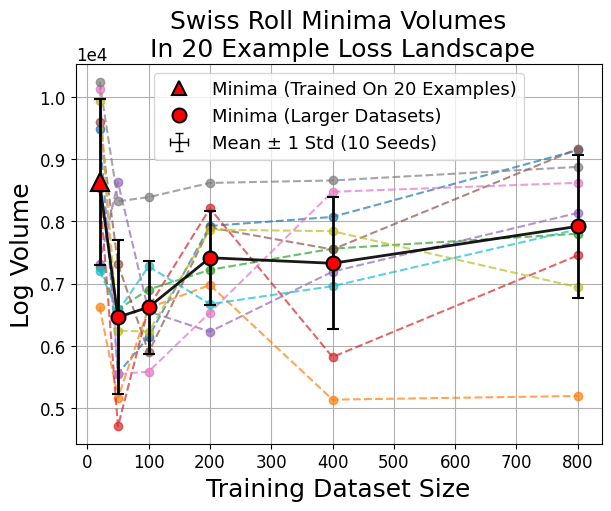}%
    \includegraphics[width=0.48\textwidth]{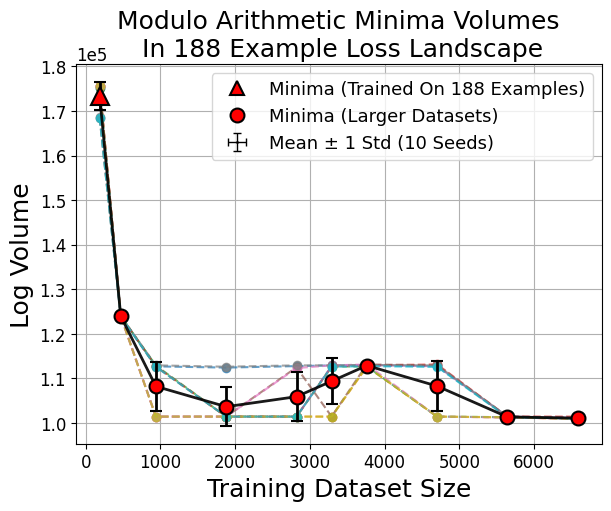}
    \caption{Volumes of minima obtained from training on larger amounts of data (red circles) evaluated in a loss landscape formed by a small fraction of the total training data. %Unlike previous experiments that show poisoned minima generalize poorly, we investigate minima from training on larger amounts of data that generalize well (red circles). 
    Minima found from training on the small dataset (red triangle) consistently have larger volumes.
    %The volumes of minima found from training on the small dataset (red triangle) are significantly larger than other minima. 
    This is in agreement with the volume hypothesis, but contradicts the flat minima hypothesis in low data loss landscapes since the small minima achieve superior generalization. Red points represent the average of 10 different seeds.} 
    \label{fig:Low Dataset Large Minima}
\end{figure}

% Things to emphasize
% volume seems like a good explanation for why we can't generalize at low data settings
% and the flat minima hypothesis does not work very well, since the flat minima 

%In contrast, minima obtained from training on the full dataset are sharper yet achieve vastly better generalization. Figure~\ref{fig:Low Dataset Large Minima} demonstrates this pattern across MNIST, CIFAR-10, Swiss Roll, and Modulo tasks: low-data solutions occupy significantly larger basins of attraction than their full-data counterparts.

%In this low data regime, we find examples of sharp minima (those found by training on larger datasets) which generalize better than flat minima. The volume hypothesis seems to explain why these minima are not found, since their volumes are very small. 

\begin{figure}[h!]
    \centering
    % Include the combined figure
    \includegraphics[width=0.48\textwidth]{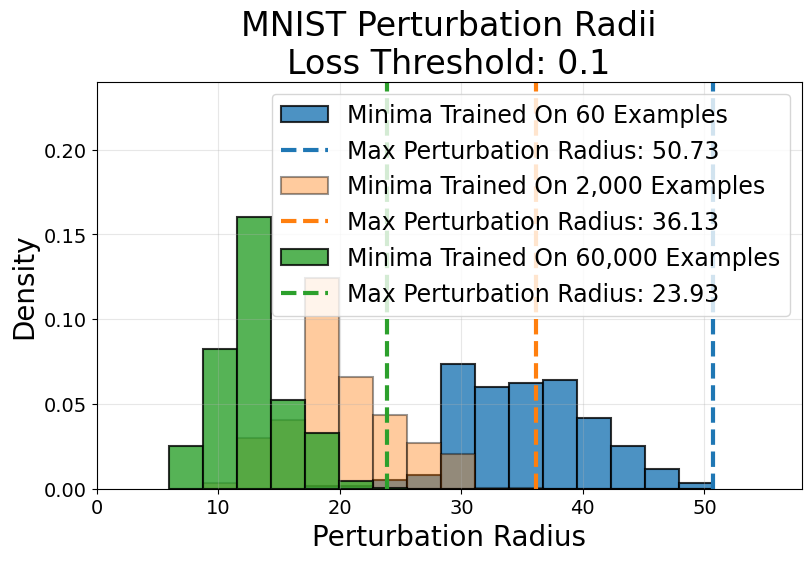}%
    \includegraphics[width=0.48\textwidth]{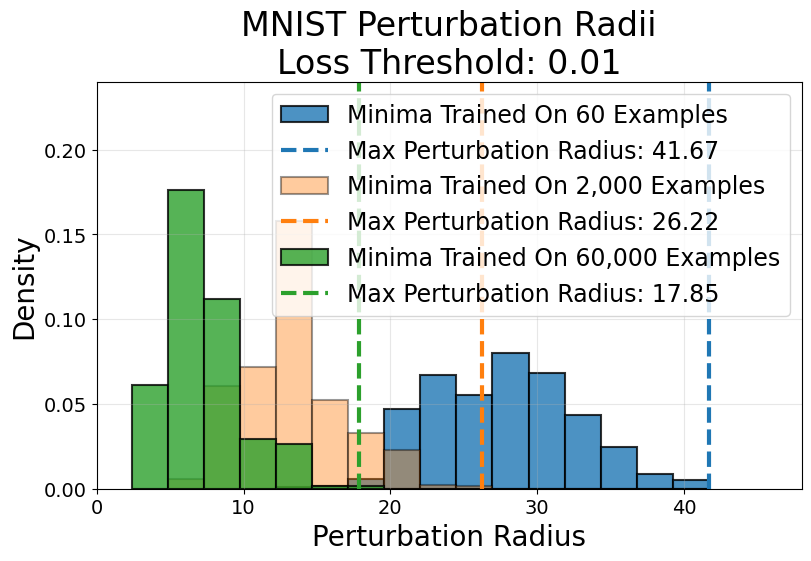}
    \caption{Distribution of distances random perturbations travel in neural network parameter space before reaching a loss threshold. In the 60 example loss landscape, MNIST minima trained on 60,000 examples are smaller than minima trained on 60 examples for most distance metrics (e.g., max, mean, min), regardless of loss threshold. See Appendix~\ref{app:robustness} for additional plots.} 
    \label{fig:MNIST Perturbation Histogram}
\end{figure}

This counterexample to the flat minima hypothesis is unlikely to be an artifact of our flatness measure. For MNIST and CIFAR10, random perturbations to low data minima travel further than any perturbations to the minima trained on the entire dataset; see Fig~\ref{fig:MNIST Perturbation Histogram} and Appendix~\ref{app:histogram_radii}.

%This possible counterexample to the flat minima hypothesis, is unlikely to result only from issues with basin volume estimation. For MNIST and CIFAR10, %and Fashion MNIST and SVHN?
%perturbations to low data minima travel further than any perturbations for the minima trained on the entire dataset, see fig~\ref{fig:MNIST Perturbation Histogram}

The volume hypothesis motivates our interest in basin volume, but Fig~\ref{fig:MNIST Perturbation Histogram} also suggests that most flatness measures based on perturbation radii (e.g., mean, median, or minimum) will also conclude the low data minima are flatter, with a possible exception for those based on special directions (e.g., along the optimizer). The difference in flatness is also less extreme for the swiss roll, where minima are of more comparable sizes, see Appendix~\ref{app:histogram_radii}.

%Any reasonable flatness measure related to how far random perturbations can travel before the loss becomes too high (such as the minimum, mean or median distance to the boundary) will conclude the low data minima is flatter. Here we only concern ourselves with the basin volume due to it's connection to the volume hypothesis.
% where only some sharp minima 

Training in low-data regimes is often described as memorization or overfitting. This seems an appropriate description for the minima that are consistently discovered in this regime, which occupy large volumes yet exhibit poor generalization.

%Our results indicate deep learning is guaranteed to find poor minima in low data settings. %I feel like I should mention this more.

%Our results here seem to suggest that low data settings are very 

%In our experiments, minima obtained in very low data settings are always larger than minima from other data settings. This appears to represent the memorization regime of neural networks.

% Needs rewrites.

%Small datasets seem to be an example where the implicit bias of large networks towards simple solutions (supposedly explained by the volume hypothesis) works against them. They do not learn useful representations, but memorize the inputs.

%The low data setting seems to be an example where the implicit bias of large networks towards simple solutions (supposedly from the volume hypothesis) works against them. The simple solution for small datasets seems to be memorizing the data instead of developing useful learned representations. In the parlance of the flat minima hypothesis, the way to easiest separate small datasets doesn't involve any useful internal representations.

%Interestingly, while minima from larger datasets increase in volume when evaluated on smaller datasets, they are noticeably smaller than the minima typically found. This suggests it may be worthwhile to develop algorithms that seek out sharp minima on small datasets, since the wide naturally found minima never generalize.

\subsection{Larger Datasets Are Problem-Dependent}

In larger datasets, volume-data relations are more complex.

\begin{figure}[h!]
    \centering
    % Include the combined figure
    \includegraphics[width=0.48\textwidth]{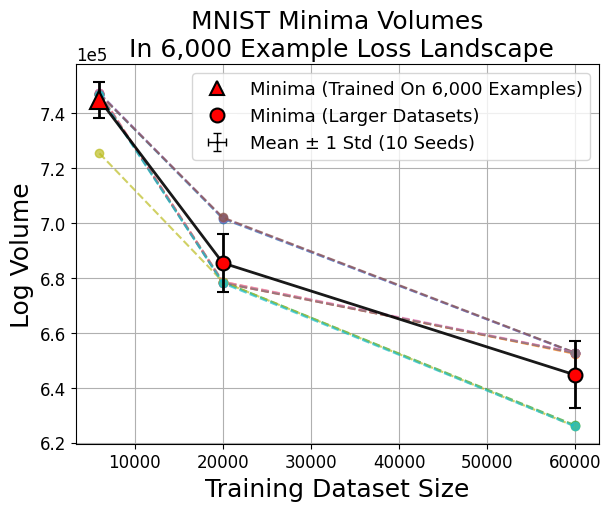}%
    \includegraphics[width=0.48\textwidth]{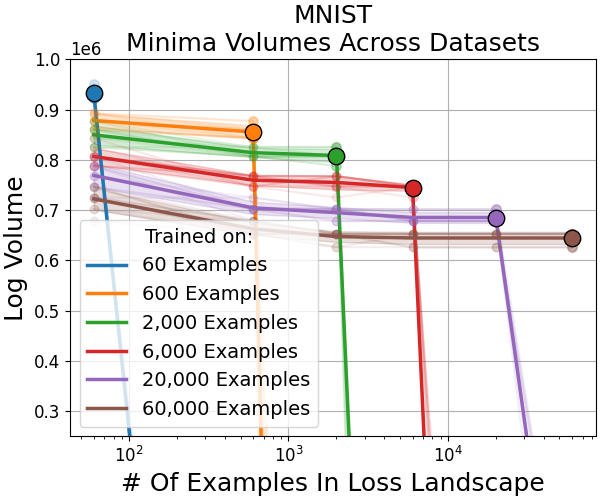}
    \includegraphics[width=0.48\textwidth]{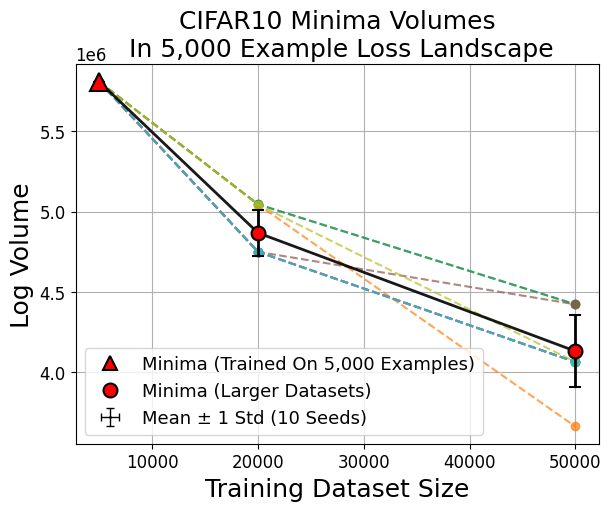}%
    \includegraphics[width=0.48\textwidth]{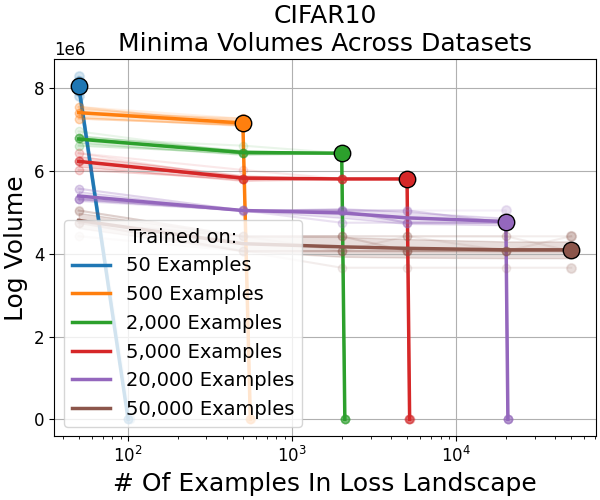}
    \caption{\textbf{Left:} In MNIST and CIFAR, the trend continues in larger datasets: minima found from training (red triangle) are still larger than other minima (red circles). %, even from training on larger datasets.
    \textbf{Right: }Volumes of minima in different loss landscapes. Lines track the behavior of minima trained on a particular amount of data. The minima found from training on a dataset is larger than any other minima in that dataset's loss landscape.  When encountering data that they have not been trained on, minima rapidly increase in loss and their volume shrinks to 0. There appears to be a power law between log volume and dataset size
    % The straight line between log volume and dataset size indicates a power law 
    that holds across three orders of magnitude. Similar trends apply in other image classification tasks; see Appendix~\ref{app:svhn_fmnist}.}
    \label{fig:MNIST CIFAR Large}
\end{figure}

%For larger dataset sizes, different problems show different relationships between minima volume and dataset size.

%\subsubsection{CIFAR and MNIST Follow A Strong Volume Hypothesis}
In MNIST and CIFAR10 (also SVHN and Fashion MNIST, see Appendix~\ref{app:svhn_fmnist}), minima found in a given loss landscape are always larger than minima from larger datasets.
%In MNIST and CIFAR (SVHN and Fashion MNIST in Appendix XXX), minima from training were always larger than minima from larger datasets, in the training landscape. 
This suggests the volume hypothesis can largely explain generalization (or lack thereof) in these problems: gradient descent finds large minima, which do not generalize as well as the sharp minima found from larger datasets. Increasing the dataset causes these large minima to rapidly shrink in volume. Training on the larger dataset finds a previously-sharp minima that has become the largest minima. The relationship between volume and dataset sizes is shown in Fig.~\ref{fig:MNIST CIFAR Large}.

This trend also occurs when considering different architectures or convolutional neural networks instead; see Appendix~\ref{app:architectures}.

Minima volume and the number of samples in the dataset appear to follow a power law. Note that for basin volume estimation
\begin{equation}
\log \left(V\right) \approx \log\left(\text{max}_{\vec\theta} \text{ } r^n(\vec\theta) \right) = n \log\left(r_{max}(\vec\theta) \right),
\end{equation}
where $n$ is the number of parameters in the model and $r_{max}(\vec\theta)$ the largest perturbation radius. Our curves suggest the following power law:
\begin{equation}
%n \log\left(r_{max}(\vec\theta) \right) = \alpha n \log(D) \implies r_{max}(\vec\theta) \propto D^{\alpha},
r_{max}(\vec\theta) \propto D^{\alpha},
\end{equation}
where $D$ is the number of samples in our dataset and scaling constant $\alpha < 0$. It is unclear from our experiments if this trend holds across more than the 3 orders of magnitude observed and if it has any relation to neural scaling laws \cite{kaplan_scaling_2020} or the manifold hypothesis.

\begin{table}[h!]
    \centering
    \begin{tabular}{l c c c c}
        \toprule
        Model & Scaling Constant $\alpha$ & Test Acc (100\%/0.1\%)\\% & Log Volume (100\%/0.1\%) \\
        \midrule
        MNIST, MLP & -0.1835 & 97.7\% / 67.5\% \\%& $9.33 \times 10^5$ / $6.44 \times 10^5$ \\
        MNIST, MLP (Deep) & -0.1804 & 97.6\% / 66.7\% \\%& $1.19 \times 10^6$ / $8.23 \times 10^5$ \\
        MNIST, MLP (SGD) & -0.1447 & 97.9\% / 68.1\% \\%& $9.13 \times 10^5$ / $6.80 \times 10^5$ \\
        MNIST, MLP (SAM) & -0.1485 & 97.8\% / 67.8\% \\%& $9.13 \times 10^5$ / $6.80 \times 10^5$ \\
        MNIST, CNN & -0.0853 & 99.1\% / 70.5\% \\%& $1.64 \times 10^6$ / $1.42 \times 10^6$ \\
        CIFAR10, MLP & -0.3394 & 52.8\% / 19.7\% \\%& $8.06 \times 10^6$ / $4.10 \times 10^6$ \\
        CIFAR10, CNN & -0.2339 & 75.5\% / 23.0\% \\%& $5.16 \times 10^6$ / $3.36 \times 10^6$ \\
        Fashion MNIST, MLP & -0.3747 & 88.2\% / 63.9\% \\%& $9.34 \times 10^5$ / $3.45 \times 10^5$ \\
        \bottomrule
    \end{tabular}
    \caption{Model, Scaling Constant, and Test Accuracy (trained on 100\%/0.1\% of the dataset).
    % \newline
    Models which perform better on low data settings seem to have smaller scaling constants. %Absolute volumes appear irrelevant.
    %Scaling constants are only observed in our experiments with image classification tasks. 
    For model details see Appendix~\ref{app:scaling_laws}.}%, and log volumes. 
    %\caption{ Models which perform better on low data settings have smaller scaling constants. %Absolute volumes appear irrelevant.
    %Scaling constants are only observed in our experiments with image classification tasks. 
    %For model details see Appendix.
    %}
    \label{tab:scaling_constant_summary}
\end{table}

%It is unclear if such relationships are interesting (due to the nature of log-log plots), but our experiments do consider a modest range of dataset sizes.
Note the scaling constant (normalized by parameter count) is not independent of model details, see Table \ref{tab:scaling_constant_summary}. The relation describes the volume of minima in the landscape it is found in and cannot be used to inform a search for good minima in low data landscapes (where volumes generally seem to be larger). We note a trend where architectures or training methods that achieve better performance for a given amount of data have less severe scaling on volumes.

One possible explanation relates the geometry of the loss landscape to achieved test loss: training settings that achieve low test loss have a loss landscape \textit{more representative} of the true distribution. This suggests volumes, in addition to minima loss, will change less dramatically as additional data is added. We present evidence to support this view in Appendix~\ref{app:batch and model seed}.

Given that minima from large datasets are consistently sharper and achieve superior generalization, one might wonder if algorithms should be optimized for sharp minima. We note Fig~\ref{fig:MNIST CIFAR poison} and Fig~\ref{fig:MNIST CIFAR Large} suggest small amounts of poisoned data result in significantly smaller minima volumes than the same amount of correct data. Intuitively, it seems most sharp minima are poor at generalization. We show this is typically the case from minima found from our optimizers in a given loss landscape in Appendix~\ref{app:batch and model seed}.

%If true, then algorithms targeting sharp minima for generalization have to find "needles in a haystack" - most sharp minima are likely terrible. Deep learning side steps this problem because the minima it finds are volume-driven and it skips many of these sharp minima.

%are the scaling relationships? robust across different loss thresholds? Need to test. 

%In MNIST and CIFAR10, minima obtained from training are always much larger than minima from training on different data sizes. Plotting the volumes of minima as a function of dataset size results in curves consistent with the strong volume hypothesis, suggesting generalization (or rather, lack of in small datasets) in these problems can be largely explained by minima volumes. The same trends are observed when considering a CNN (fig...) or deeper architectures, see appendix. 

% more paragraphs for the CNN, and how the scaling seems less severe.
% also talk about how there is an apparent scaling law, with the coefficients. It's unclear if this can be connected to the manifold hypothesis.

\begin{figure}[h!]
    \centering
    % Include the combined figure
    \includegraphics[width=0.48\textwidth]{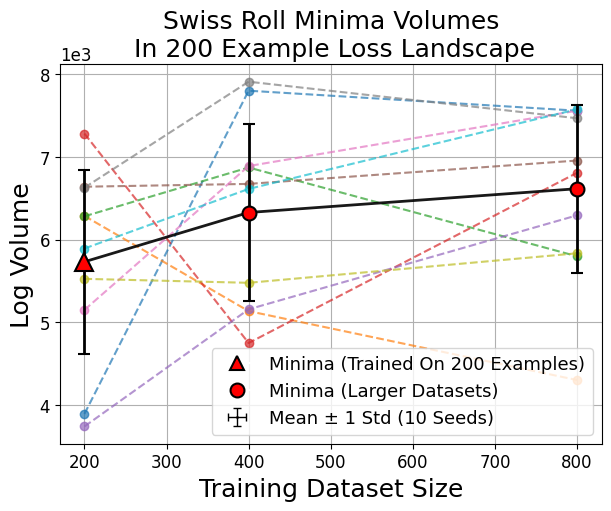}%
    \includegraphics[width=0.48\textwidth]{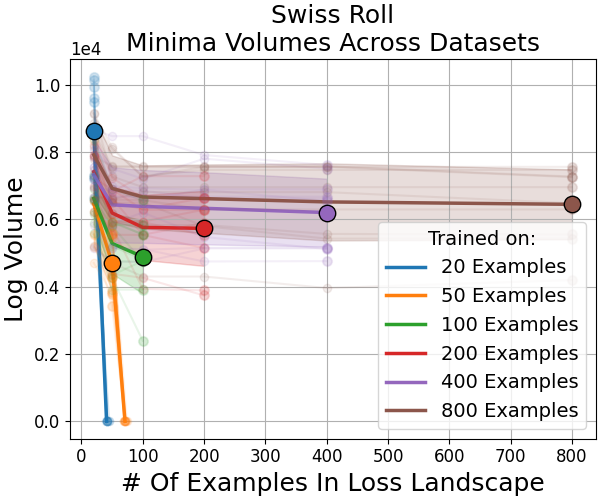}
    \includegraphics[width=0.48\textwidth]{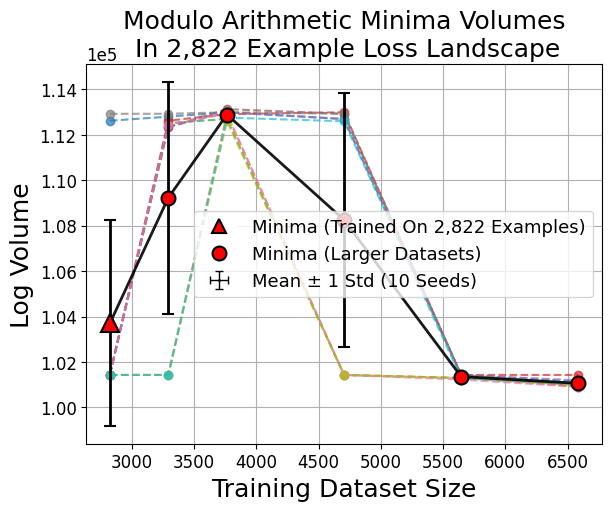}%
    \includegraphics[width=0.48\textwidth]{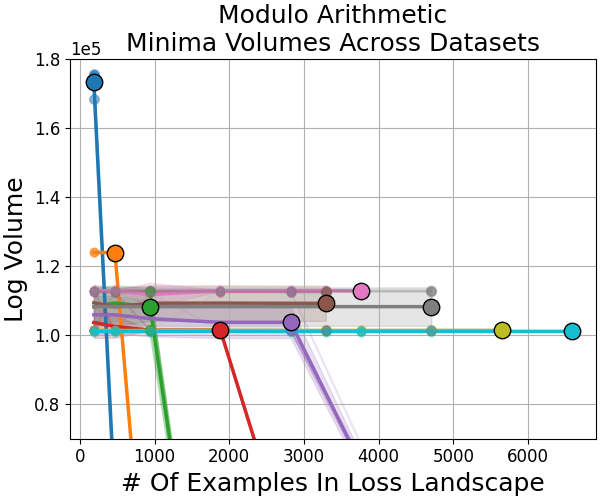}
    \caption{\textbf{Left:} For swiss roll and modulo arithmetic (base 97), the minima found by training (red triangle) were sometimes smaller than minima from larger datasets. This is not necessarily a counterexample to the volume hypothesis, since the overall volume of minima typically found at this dataset size could be large if there are many minima at this size. We also note that across our experiments, variability between seeds is more significant, suggesting the disparity in volumes is not that large. For modulo arithmetic, 2,822 examples is 33\% of the dataset.
    \textbf{Right: }Volumes of minima in different loss landscapes. Lines track the behavior of minima trained on a specific fraction of data, taken to different dataset size landscapes. Unlike before, the minima found from training are not always the largest on a dataset, and there is no sign of a power law between minima volume and dataset size.} 
    \label{fig:Swiss Modulo Small}
\end{figure}

%swiss and modulo arithmetic do not show the trends

These volume-data trends are also not universal. In swiss roll and modulo arithmetic, at sufficiently large dataset sizes we find minima from training that are no longer the largest by volume.

This agrees with the flat minima hypothesis—larger minima tend to generalize better—but is not necessarily a counterexample to the volume hypothesis. The hypothesis implies the total volume is what matters, not the volume of an individual minima. If there are many small but similar minima in the loss landscape, then by total volume we are biased towards finding these small minima. 

%This is consistent with the flat minima hypothesis since the larger minima generalize better, but is not necessarily a counterexample to the volume hypothesis. It is still possible that volume heavily influences the found minima, since what matters is not the volume of an individual minima but the volume associated with all the minima of the generalization typically observed from training. If there are many small but similar minima in the loss landscape, then by total volume we are highly biased towards finding them.

Note that compared to MNIST and CIFAR, the variation in volume from different seeds is comparable to the variation between minima from different data settings. This suggests the minima are similar in volume, in line with our explanation on minima number.

%We  point out that compared to the differences observed earlier, the variation from different random seeds comprises are degree of the variation between minima. This may be evidence that the 

%For these problems, there is no scaling law between volume and data. We point out that the spread in 

%only when the dataset is extremely small are the minima obtained the largest (in the small dataset). In general, in other dataset sizes the minima naturally found by training was smaller than the minima obtained in other dataset sizes. This trend can be explained by invoking the possibility of multiple minima (the weak volume hypothesis), which cannot be measured in practice. Again, the trends remain when considering deeper architectures, see appendix.

% also no such scaling law as found in MNIST, CIFAR10 and etc

\subsection{Minima (Almost) Always Shrink}

In almost all experiments, the volume of minima shrink as dataset sizes are increased. When encountering data they were not initially trained on, volumes shrink rapidly, often exceeding the chosen basin loss threshold. There was no clear trend to how fast volumes shrunk before this threshold (minima from a large dataset can shrink faster than others at first). The relative ordering of minima sizes generally remains the same in different landscapes except for the minima which vanish as they encounter data not seen in their training. Experiments focused on data outside of training sets, with much higher loss thresholds, are included in Appendix~\ref{app:batch and model seed}. Here we also see more examples of volumes growing with data.

This behavior indicates that volume alone cannot reliably predict generalization performance. All minima—large or small—experience drastic volume collapse when exposed to new data. %The decisive factor is not the absolute size of a minima, but how abruptly it disappears outside its training distribution.

%This suggests there is no clear way from volume alone to determine how well a minima will do on unseen test data, since minima of all sizes will abruptly and rapidly shrink in volume on new data. 

\begin{figure}[h!]
    \centering
    % Include the combined figure
    \includegraphics[width=0.48\textwidth]{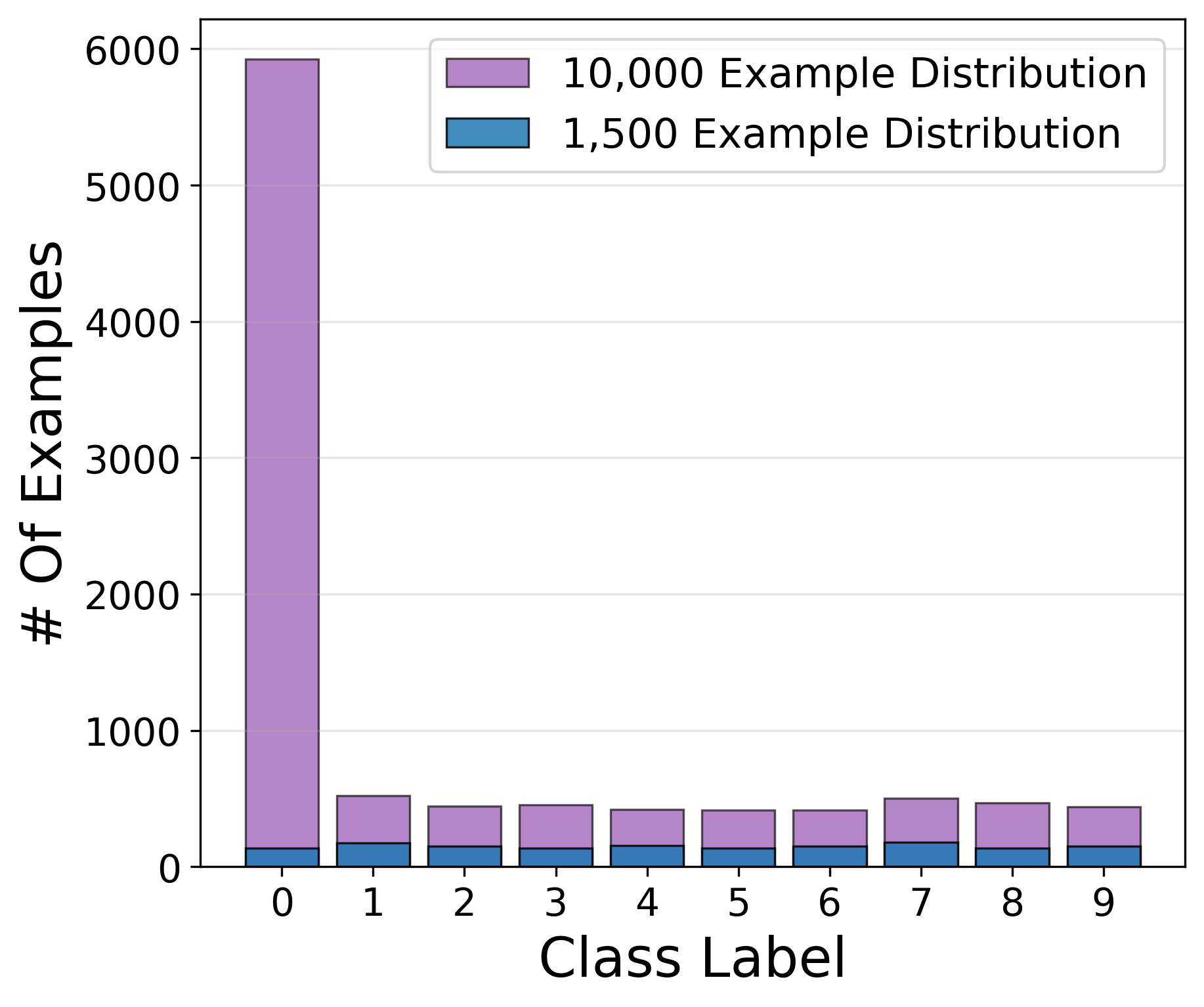}
    \includegraphics[width=0.48\textwidth]{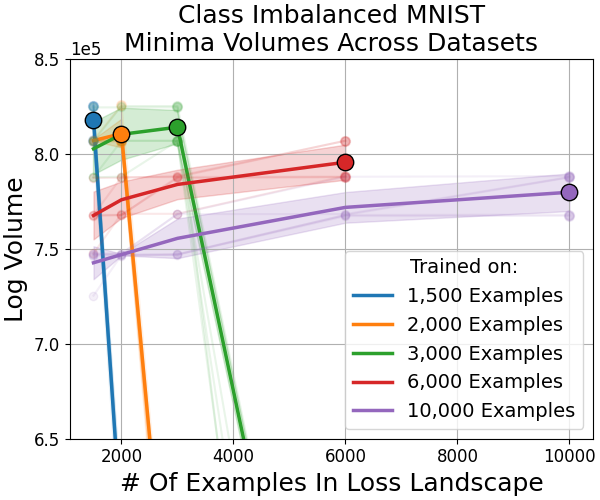}
    \caption{A rare example where minima volumes increase with dataset size. \textbf{Left:} Training on a class-imbalanced subset of MNIST (purple) results in a minima whose volume shrinks in a smaller balanced distribution (blue). \textbf{Right:} A minima increase in volume as the dataset is increased to the original imbalanced dataset (purple line).} 
    \label{fig:Class Imbalance Experiment}
\end{figure}

Theoretically, it is possible for volumes to increase with data. To find an example, we examined class-imbalanced datasets. In MNIST and CIFAR10, a model trained on an imbalanced dataset taken to a smaller, evenly class-balanced dataset shrunk in volume. Conversely, the volume increases as the dataset is increased in an imbalanced way, see Fig~\ref{fig:Class Imbalance Experiment}. Outside of this contrived setting, volumes almost always decreased. %Even the reverse of this procedure, obtaining a minima from a large balanced dataset and then evaluating on a smaller imbalanced dataset, resulted in larger volumes in the smaller dataset.

\section{Other Generalization Phenomena}

The volume hypothesis abstracts away training details, proposing generalization can be explained largely by loss landscapes. However, to better understand our results (which contradict either the flat minima or volume hypotheses in low data regimes), we study minima volumes under two training variations with well-established effects on generalization: sharpness-aware minimization and grokking.

% Remember: We want to support the big picture, that we do typically find flat minima, they generalize somewhat well (sam) but there are other ways to generalize with sharp minima (ALL OUR RESULTS + Grokking)

% To better understand our results, we look at some other details?

%The volume hypothesis does not incorporate any training details, as it suggests much of generalization can be attributed purely to loss landscapes. However, there are many training-dependent generalization phenomena. Here, we study sharpness-aware minimization and grokking to see how their effects may be independent of the volume-fueled generalization seemingly observed earlier. 

\subsection{Sharpness-Aware Minimization}

Our results have shown that minima obtained by training on larger datasets are likely to be `sharp' relative to the minima found normally by gradient descent on smaller datasets. %(with most experiments for AdamW, but similar results for SGD - see Appendix).

\begin{figure}[h!]
    \centering
    % Include the combined figure
    \includegraphics[width=0.48\textwidth]{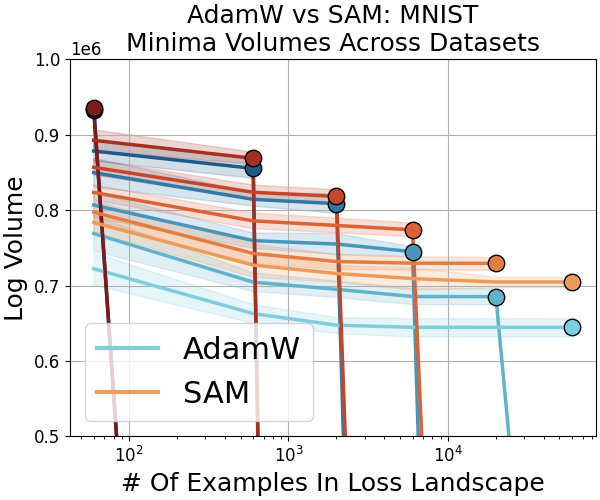}%
    \includegraphics[width=0.48\textwidth]{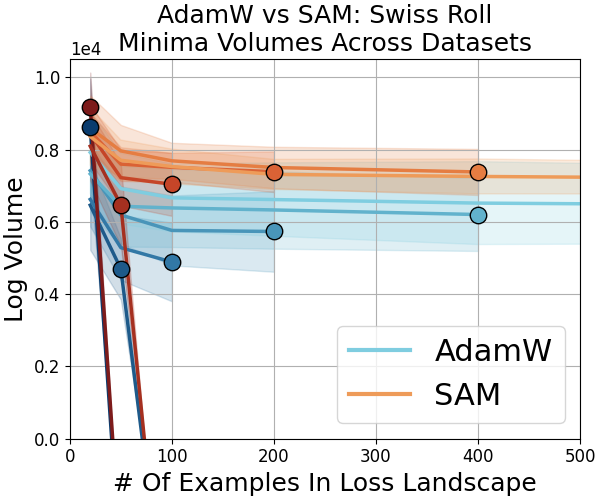}    \includegraphics[width=0.48\textwidth]{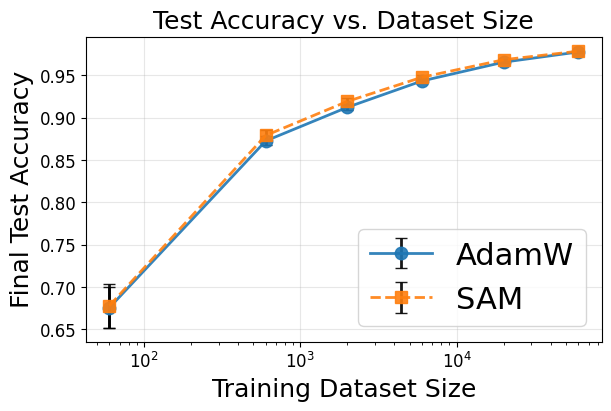}%
    \includegraphics[width=0.48\textwidth]{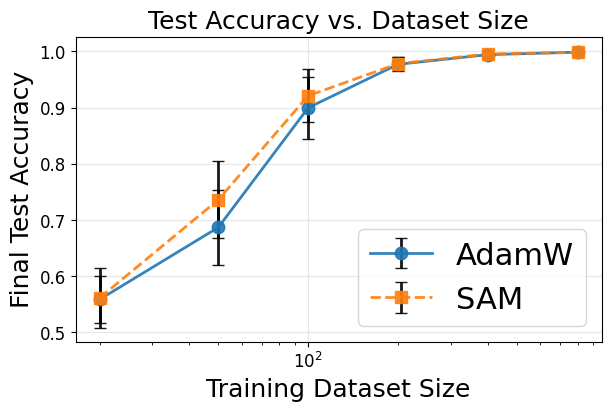}
    \caption{Minima found via sharpness-aware minimization (SAM) are larger in volume (top) and show modest improvements in test accuracy (bottom). Data-volume trends observed for AdamW-found minima also occur in minima found with SAM. Our results do not contradict the existing results suggesting flatter minima are better, but the improvement from large datasets seems different than the improvement from flatness.} 
    \label{fig:Sharpness Aware Minimization}
\end{figure}

This does not contradict prior findings that flatter minima often generalize better. Previously, we showed minima trained on larger datasets are sharper if the additional data is incorrectly labeled instead of correctly labeled, which suggests most sharp minima are bad. In Appendix~\ref{app:batch and model seed}, we also show that in a given loss landscape, flatter minima obtained from optimizer variation tend to have better generalization.

%We emphasize this does not contradict existing results for flat minima being good indicators of generalization. Previously we showed minima trained on larger datasets are sharper if the additional data is incorrectly labeled instead of correctly labeled, which suggests most sharp minima are bad.

%Here, we show our measure of flatness (basin volume) can agree with existing results arguing minima flatness improves generalization. 
Here, we show a similar result with sharpness-aware minimization (SAM).
Sharpness-aware minimization (SAM) is a modification of gradient descent that minimizes the \textit{maximum loss} in a nearby neighborhood. In searching for these flat minima, SAM %demonstrates strong empirical 
empirically improves generalization on a variety of tasks.
%results in improving generalization for a variety of tasks.

In our experiments, SAM achieves improved test accuracy and finds larger minima than vanilla AdamW. The largest increases in accuracy were found in swiss roll with small amounts of data, which is where we previously speculated the minima found were driven more by the number of minima rather than individual flatness. 

This link between volume and generalization is also seen if we vary batch sizes, where small batch sizes find larger volume minima with better generalization, in agreement with Keskar et al's results~\cite{keskar_large-batch_2017}, see Appendix~\ref{app:batch and model seed}.

Thus, the search for flatness to improve generalization is sensible (as it must be, with the existing empirical results). But our results imply the improved generalization from flatness is fundamentally different from the improvement obtained by training on larger datasets.

% 

%Previously we mentioned that minima trained on larger datasets are sharper if the additional data is poisoned, suggesting 

%With sharpness aware minimization, we find the volume

%We retrain with sharpness aware minimization and document our results. Did the volumes go up compared to the original?

\subsection{Grokking}

In grokking, training loss rapidly decreases and plateaus, followed by a much slower decrease in test loss. First observed with transformers in small algorithmic tasks such as modulo arithmetic \cite{power_grokking_2022}, it can occur even with single hidden layer MLPs \cite{gromov_grokking_2023}. Our experiments use this setup, where transitions in test loss are less extreme but the models are simple.

Theoretically, the volume and flat minima hypotheses could explain this behavior. If the loss landscape consists of many small minima and rare large minima, volumes imply training will almost always initially find small minima (with poor generalization). Viewing training as a random walk over the landscape (which has previously helped analyze the effects of batch sizes \cite{hoffer_train_2018}), the model will eventually find and settle into a large minima with better generalization.

This random-walk view is inconsistent with existing results that suggest networks make consistent progress towards generalization over the long training period. Networks appear to initially memorize, build useful circuits throughout training, and finally prune memorized components \cite{nanda_progress_2023}. 

\begin{figure}[h!]
    \centering
    % Include the combined figure
    \includegraphics[width=0.48\textwidth]{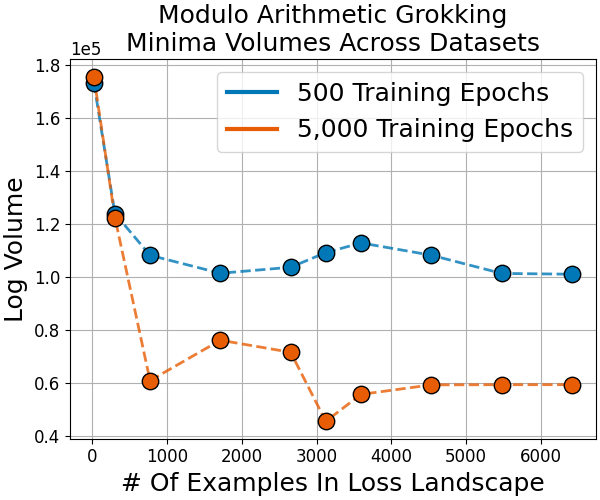}%
    \includegraphics[width=0.48\textwidth]{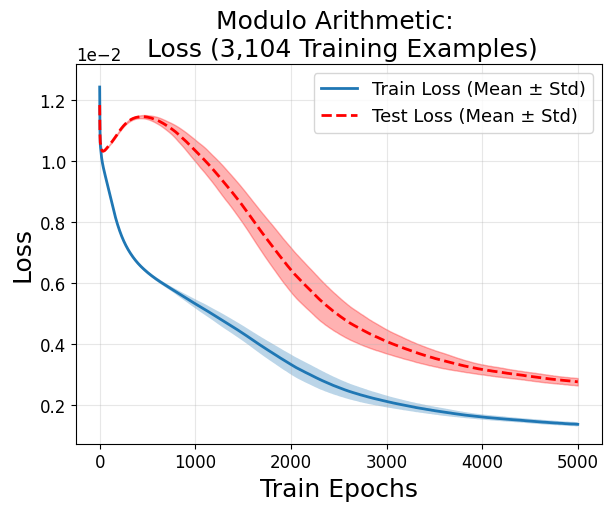}
    \includegraphics[width=0.48\textwidth]{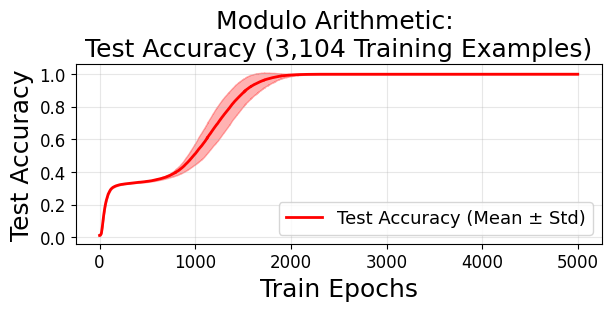}%
    \includegraphics[width=0.48\textwidth]{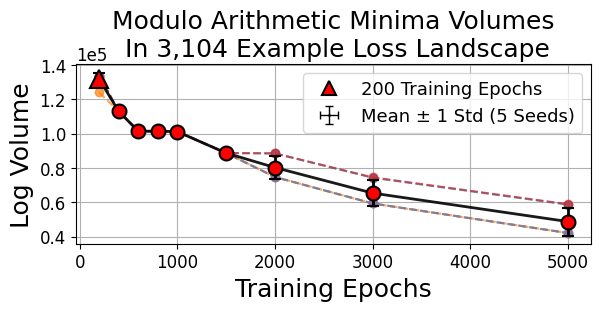}
    \caption{\textbf{Top Left: }Training our models for a large number of epochs reveals dramatic reductions in minima volume. The most interesting effects occur around $\approx 33\%$ of the training data, which is also where grokking occurs for this model architecture. \textbf{Top Right: }Losses over epochs for 33\% of the training data. \textbf{Bottom Left: }Test accuracy over epochs from 33\% of the training data. The test accuracy remains constant for many epochs before suddenly increasing, as is usual in the grokking. \textbf{Bottom Right: }The volumes decrease throughout training even when accuracy initially plateaus.}
    %\textbf{Top Right: }For 33\% of the training data, the volumes steadily shrink throughout training. \textbf{Lower Right: } The test accuracy abruptly increases over training.} % should probably plot the test loss and the train losses too 
    \label{fig:Grokking Volumes}
\end{figure}

We see results consistent with this latter picture. In modulo arithmetic, extending training from 500 epochs to 5,000 epochs produces almost no change in training loss, but volumes shrink significantly and test accuracy increases dramatically. Plotting volumes across training epochs reveals a steady and consistent decrease, see Fig \ref{fig:Grokking Volumes}. The large reduction in volume from increased training appears unique to modulo arithmetic; see Appendix~\ref{app:mnist_epochs} for experiments with MNIST.

These results are consistent with the volume hypothesis but serve as another counterexample to the flat minima hypothesis. The larger-volume minima are easier to find and thus are found earlier in training with fewer training epochs. However, generalization is achieved when parameter values descend into a narrow valley, occupying small volume.

Why do these sharp minima generalize better than flat minima? In this case, prior work studying grokking in modulo arithmetic suggests the final model implements simple solutions that capture the cyclic nature of the problem \cite{nanda_progress_2023, gromov_grokking_2023}. Our solutions are evidently sharp in parameter space (measured by basin volume), but flatness is only loosely-related to complexity, which was the original motivation for flat minima. Experiments with compression indicate that Kolmogorov complexity decreases in grokking \cite{DEMOSS2025134859}, which may explain why the sharp minima generalize better---they are not flat, but still simple networks.

\section{Conclusion}

The volume hypothesis attributes generalization in deep learning largely to the geometry of loss landscapes. Landscapes possess minima with dramatically different volumes, and gradient descent %optimizers such as AdamW are 
is biased towards minima with large volumes. Large volume minima are flat and supposedly generalize well. Data is absent in this picture. To understand its role, we measured volumes of minima obtained by training on varying amounts of training data.%, tracking how volumes change depending on the amount of data in the loss landscape.

We measured volumes with a Monte Carlo basin volume estimation technique. While this approach has theoretical shortcomings, empirically, minima in our experiments appear sufficiently well-behaved that their relative volumes can be reliably distinguished.

We find minima, obtained from training on larger datasets, occupy small volumes in the loss landscape of smaller datasets. The prediction that gradient descent tends towards large minima is mostly supported by our experiments, where the minima found from training on a given dataset are larger than minima not found. Our results instead find several counterexamples to the flat minima hypothesis, where sharp minima (from training on larger datasets) generalize better than the flat minima in a given landscape.

This suggests the following picture of generalization: deep learning is biased towards large volume minima which appear to generalize well, but the best-generalizing solutions may be sharp. The volume bias means these minima are only found by increasing the amount of data, which changes the loss landscape such that previously-large minima shrink in volume. The previously-sharp generalizing minima are now the largest in volume.

In image classification, we observe a strong bias toward flat minima and a striking power-law relationship between dataset size and minima volume spanning three orders of magnitude. Inductive biases and training paradigms (e.g., CNNs for images, SAM) appear to modify this scaling. Methods that generalize better show smaller differences in minima volume across dataset sizes. We speculate this may be linked to the manifold hypothesis and neural scaling laws. It is unknown whether similar structure appears in domains such as language modeling.

We also find cases where gradient descent finds relatively `sharp' minima in swiss roll classification and modulo arithmetic at large dataset sizes. But the differences in volume are relatively small compared to image classification tasks (and comparable to noise from our Monte Carlo approach). We argue these cases can be explained by an argument about the number of minima---numerous sharp minima may occupy substantial volume collectively.

Our experiments do not disagree with the existing work connecting generalization and flatness. As an example, we show sharpness-aware minimization (SAM) improves both test accuracy and basin volume, in line with existing ideas on flatness. But the generalization from SAM and from training on larger datasets appear different in nature. SAM finds flat minima while large datasets find sharp minima.

% In the past, there has been a contest to find generalization measures. The best performers were not methods around flatness but instead took advantage of data augmentation or domain specific methods. Our results suggest these contests can easily disqualify flatness measures by using minima trained on larger amounts of data, which will likely be sharp in the loss landscape provided to contestants.

Turning to grokking modulo arithmetic,
%As a final experiment to show sharp minima can generalize better, we look at grokking modulo arithmetic. 
we find the volumes of minima shrink throughout the grokking process. While outside the scope of our study, this can be explained by the initial motivation for the flat minima hypothesis in information theory. Flat minima can be stored with less precision, resulting in a low minimum description length and lower complexity. But the models found at the end of grokking, while not flat in neural network parameter space, are algorithmically simple and have been shown to admit low minimum description length \cite{DEMOSS2025134859}. 

\subsection{Future Outlook}

Our results highlight an interesting gap in generalization research. Previous studies focused on flatness based off experiments on minima with different optimizers. Yet empirically, large datasets are essential to models used today, which our work suggests likely correspond to `sharp' minima in smaller datasets.

We are not the first to suggest sharp minima can generalize~\cite{wen_sharpness_nodate, andriushchenko_modern_2023}. What we add is a link between sharp minima and the known value of large datasets via the volume hypothesis, which our experiments suggest is a plausible explanation for deep learning. While not a true theory of generalization, these are useful connections for such a theory.

Our findings suggest that data-efficient deep learning may require algorithms that consider `sharp' minima.
%consider sharp minima, or at least, minima that are sharp in low data
%, unregularized, standard loss landscapes. 
The volume hypothesis and results with poisoned minima suggest this is a difficult task. Minima realistically reachable with only general optimizer tweaks may be very limited, which would explain why existing research has suggested prioritizing flat minima instead of sharp minima.

But note none of the flat minima observed here for our high-dimensional problems perform as well as `sharp' minima (nor as poorly, to the credit of deep learning). And this trend seems likely to continue, given the generality of improvements from large data sizes, the volume hypothesis, and the apparent existence of minima scaling laws observed here. Since very good flat minima haven't been found, the search for flatness may not be the best route to improving generalization. 
%Since very good flat minima haven't been found, it is worth looking at sharp minima instead.

%this with the empirically observed improvements from large datasets, and the volume hypothesis which appears to describe results from deep learning.

% our experiments highlight an interesting gap here, not the first to argue that sharpness is the best, but we look at a useful slice of minima and provide strong mechanistic explanation for it from the volume hypothesis.

%Interesting directions for future work include:
%\begin{itemize}
%    \item Investigating how architectural and inductive biases shape volume dynamics.
%    \item Explaining the volume scaling law observed for image classification.
%    \item Examining potential volume scaling laws in large-scale domains such as language or audio.
%    \item Refining volume estimation techniques, such as those proposed by Schleris et al \cite{scherlis_estimating_2025}.
%    \item Studying volume dynamics in other cases of grokking to assess whether shrinking volumes occur universally, and if volume can help predict the onset of grokking.
%\end{itemize}
\section{Contributions and Acknowledgments}
Raymond Fan initially conceived the idea, developed main experiments and open-source code. Additional experiments were contributed by Bryce Sandlund. All authors contributed input in designing the experiments. Raymond Fan and Bryce Sandlund drove the writing of the paper, with contributions from Lin Myat Ko. We thank ML Collective and Jason Yosinski for useful discussions.

\section{Reproducibility Statement}

The code to reproduce results can be found at \url{https://github.com/rfangit/minima-volume-project}. All experiments were run with prespecified random seeds, so all plots are reproducible from the respective scripts in our codebase.

As additional aid for researchers interested in minima volume, we provide a google colab tutorial that measures volumes of MNIST minima, which can be freely modified for other experiments: \url{https://colab.research.google.com/drive/1JNbk8Sau-M31mLVOQv19GR2dlwW7xwLd}.

\bibliography{MinimaVolumeBib} % Note: No .bib extension

\clearpage
\appendix      % Switch to Appendix mode
% appendix.tex

\section{Neural Network Parameter Space}
\label{app:parameter_space}
Here we discuss properties of neural network parameter space and our perturbations (including the filter normalization procedure).

Neural networks parameter space is high dimensional and contains many nontrivial invariances, which complicates analysis. One well-known invariance is scale invariance. This can be illustrated with a simple toy model:

\begin{equation}
f(x, w_1, w_2) = w_2 \left(w_1 x\right) .
\label{EQ: toy model}
\end{equation}

Here, the model depends only on the product ($w_1 w_2$). For any scalar ($\alpha > 0$), the parameter pair ($\alpha w_1, w_2 / \alpha$) produces the same outputs and identical loss. Consequently, scale invariance creates continuous regions of equivalent solutions in parameter space.

In networks with ReLU activations, scale invariance holds for positive weights, still resulting in infinitely large solution regions. For more complex architectures, such as convolutional networks, scale transformations must be applied consistently across entire layers to preserve the output (e.g., every filter in a layer of the convolutional network needs to be rescaled the same way).

As an example of invariances aside from scale invariance, some input pixels in image classification tasks may be effectively ignored by the network. Parameters in layers corresponding to these ignored inputs can vary freely without affecting the loss. Since the lottery ticket hypothesis suggests large networks learn to utilize only a subset of their parameters~\cite{frankle_lottery_2019}, this scenario of many irrelevant parameters may be quite significant.

\subsection{Analytical Example of Basin Volume Scale Invariance}
\label{app:analytical_basin_volume}
Scale invariance is a problem for many flatness measures because one can create arbitrarily sharp minima (as measured by eigenvalues of the Hessian) by rescaling different layers~\cite{dinh_sharp_2017}. This appears to be less problematic for basin volume than other measures.

Consider basin volumes in the following toy model of scale invariance: we lie on a point located at $x = b, y = 1/b$ for scale factor $b > 0$. The boundaries of the basin are $y = \frac{1 \pm s}{x}$ for $1 > s > 0$. If the volumes are independent of the scale factor, then there is no dependence on $b$ in the volume estimated from this point. This problem is plotted in Fig~\ref{fig:scale_invariance}.

The basin volume estimated here corresponds to the minima volume of a 2 parameter model (Eq.~\ref{EQ: toy model}) where the minima lies at the center of the basin and the increase in loss is approximately symmetric in perturbations to the defining parameter (the product $a = x y = 1$ for our case where $w_1 = x, w_2 = y$). Despite its simplicity, it captures all the minimal aspects of scale invariance, and is sufficient to ruin simple eigenvalue-based metrics of flatness.

The volume estimated from this point can be computed directly, as long as we obtain equations describing the basin. The basin has the following properties:

\begin{enumerate}
    \item The lower boundary of the basin is given by $(x, \frac{1 - s}{x})$.
    \item At the left, there is an $x$ such that a line from $(b, 1/b)$ to $(x, \frac{1 - s}{x})$ touches a point on the upper boundary $(x_{c, 1}, \frac{1 + s}{x_{c, 1}})$. For x values $x_{c, 1} > x > x_{i}$, the upper boundary of the basin is given by this straight line.
    \item For $x_{c, 2} > x > x_{c, 1}$, the basin boundary is given by the upper boundary $(x, \frac{1 + s}{x})$. $x_{c, 2}$ is a critical point similar to $x_{c, 1}$, but for the larger w values.
    \item  At the right, there is a $x_f$ that plays a similar role to $x_i$. 
\end{enumerate}

\begin{figure}[h!]
    \centering
    \includegraphics[width=0.48\textwidth]{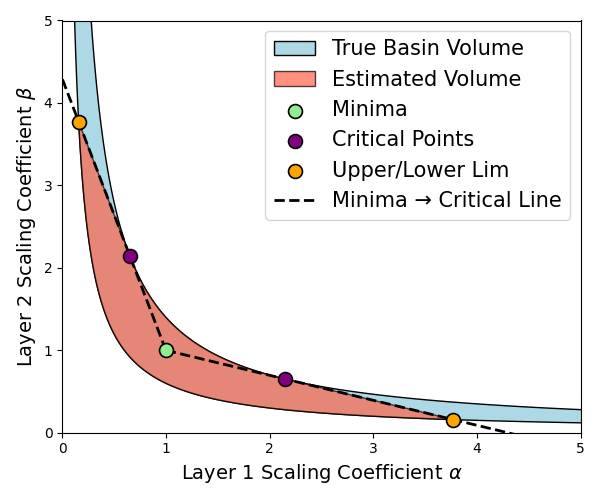}
    \caption{We compute the yellow and purple coordinates, and from that we can integrate for the total volume of the basin. Dashed lines emphasize how the boundary is linked to the region reachable by a straight line from the minima.} 
    \label{fig:scale invariance diagram}
\end{figure}

From calculus, the critical points are
\begin{equation}
x_c = b(1 + s) \left(1 \pm \sqrt{\frac{s}{1+s}} \right),
\end{equation}
where $\pm$ denotes the two critical points. Remember $x_{c, 1} < b$ denotes the critical point less than the original point's x value (given by the negative value), and $x_{c, 2} > b$ the critical point greater.

The slopes $m$ and intercepts $d$ of the two straight lines on the boundary of our basin are

\begin{equation}
m_{i} = - \frac{1 + s}{x_{c, 1}^2},\quad m_{f} = - \frac{1 + s}{x_{c, 2}^2},\quad d_{i} = 2\frac{a + s}{x_{c, 1}},\quad d_{f} = 2\frac{a + s}{x_{c, 2}}.
\end{equation}
The smallest and largest $x$ values in our basin are given by
$$
x_i = \text{min}(x) = x_{c, 1} \left(1 - \sqrt{\frac{2s}{1 + s}} \right),\quad x_f = \text{max}(x) = x_{c, 2} \left(1 + \sqrt{\frac{2s}{1 + s}} \right).
$$
The volume integral is 
$$
V = \int_{x_i}^{x_{c, 1}}\bigg( m_i x + d_i - \frac{1 - s}{x} \bigg) dx + \int_{x_{c, 1}}^{x_{c, 2}}\bigg(\frac{2 s}{x} \bigg) dx + \int_{x_{c, 2}}^{x_f}\bigg( m_f x + d_f - \frac{1 - s}{x} \bigg) dx.
$$

There is a dependence on the scaling factor $b$ in critical values $x_{c}$ and the maximum and minimum $x_f, x_i$. But carrying out this integral yields cancellations (from the slope and intercepts), resulting in
$$
V = 2\sqrt{2s (1 + s)} + 2s \log \frac{\left(1 + \sqrt{\frac{s}{1+s}} \right)}{\left(1 - \sqrt{\frac{s}{1+s}} \right)} - (1 - s)\log \frac{\left(1 + \sqrt{\frac{2s}{1 + s}} \right)}{\left(1 - \sqrt{\frac{2s}{1 + s}} \right)},
$$
which is independent of scale factor $b$. Thus the basin volume of this toy model of scale invariance is invariant to rescaling and is plotted in both plots of Fig~\ref{app:analytical_basin_volume} for $s = 0.2$, with $b = 1, 3$.

It is unclear if this holds in higher dimensions or more generally, but this toy model is sufficient to stop eigenvalue-based sharpness metrics.

\subsection{Perturbations and Filter Normalization}
\label{app:filter_normalization}
Let the initial parameters of our model be denoted by $\vec{W}_0$ (in our experiments, all weights and biases). A random normal variable is generated for each parameter, $\vec{P}$. We evaluate the loss of our model on a dataset by varying model weights along perturbation $\vec{P}$, specifically
\begin{equation}
\vec{W} = \vec{W}_0 + c \times \vec{P} \circ \vec{F},
\end{equation}
where $c$ is a scalar coefficient from $[0, 1]$, $\circ$ the element-wise multiplication between two vectors (the Hadamard product, which unlike the dot product yields a vector) and $\vec{F}$ the filter-normalization.

The filter-normalization procedure was suggested by Li et al.~\cite{li_visualizing_2018}. It handles scale and other invariances. To understand this procedure, consider the following model
\[
\vec{x} =
\begin{bmatrix}
x_1 \\[2mm]
x_2 \\[2mm]
x_3
\end{bmatrix}, 
\quad
\vec{w}_1 =
\begin{bmatrix}
w_{1,1} \\[2mm]
w_{1,2} \\[2mm]
w_{1,3}
\end{bmatrix}, 
\quad
f(\vec{x}; \vec{w}_1, w_2) = w_2 (x_1 w_{1,1} + x_2 w_{1,2} + x_3 w_{1,3}).
\]
The model is scale invariant in $\vec{w}_1, w_2$. Perturbations can be made similarly scale invariant if they are multiplied by a factor related to the overall scale of the weights. But naively scaling our perturbation by the scale of the corresponding weights has issues. Consider the following weights and perturbations
%However, there is an issue with 

% scale invariance is between different models, we don't want the volumes to change

%On top of scale invariance by multiplying $w_2$ and vector $w_1$ by scalar coefficients, if $w_1$ has components which are very small then the volume of the basin will seem unnaturally large. Eg, consider the following weights and perturbations

\[
\vec{w}_1 =
\begin{bmatrix}
1 \\[2mm]
1 \\[2mm]
0
\end{bmatrix}, \quad
\vec{P_1} =
\begin{bmatrix}
1 \\[2mm]
0 \\[2mm]
0
\end{bmatrix}, \quad
\vec{P_2} =
\begin{bmatrix}
0 \\[2mm]
0 \\[2mm]
1
\end{bmatrix}
\]

If perturbations are scaled directly by the corresponding parameter, then $\vec{P_2}$ will be associated with enormous volumes since $\vec{w}_1 \circ \vec{P_2}$ does not affect model performance at all. 

Filter-normalization solves this issue by instead multiplying perturbations by the norm of the layer, or filter (e.g., in the context of convolutional neural networks). In our toy model, $\vec{w}_1$ is a layer that results in the following filter norm
\begin{equation}
\vec{W}_0 = 
\begin{bmatrix}
w_{1,1} \\[2mm]
w_{1,2} \\[2mm]
w_{1,3} \\[2mm]
w_2
\end{bmatrix} \implies 
\vec{F} = 
\begin{bmatrix}
\sqrt{w_{1,1}^2 + w_{1, 2}^2 + w_{1, 3}^2 }  \\[2mm]
\sqrt{w_{1,1}^2 + w_{1, 2}^2 + w_{1, 3}^2 }  \\[2mm]
\sqrt{w_{1,1}^2 + w_{1, 2}^2 + w_{1, 3}^2 }  \\[2mm]
\sqrt{w_2^2}
\end{bmatrix}.
\end{equation}
For a convolutional filter, e.g.
\begin{equation}
\mathbf{W}_{\text{conv}} = 
\begin{bmatrix}
w_{1,1} & 0 & w_{1,3} \\[1mm]
0 & w_{2,2} & 0 \\[1mm]
w_{3,1} & 0 & w_{3,3}
\end{bmatrix},
\end{equation}
the corresponding filter norm has the same shape, with all elements $N = \sqrt{w_{1,1}^2 + w_{1,3}^2 + w_{2,2}^2 + w_{3,1}^2 + w_{3,3}^2}$. 

The process of identifying the layers for novel architectures may be cumbersome (here, we only focus on MLPs and CNNs), but experiments without filter normalization (see Appendix~\ref{app:filter_normalization}) suggest our conclusions are mostly independent of the procedure. However, since it is well motivated theoretically and seems to reduce variability, we apply it to most of our experiments.

%Perturbations $\vec{P_2}, \vec{P_3}$ will be associated with enormous volumes, because they extend in directions that are irrelevant to model performance. The filter normalization...

% need to think a bit more - theres a problem here where the scale of weights don't matter that much
% it's because we want perturbations to be scaled between different models - it isn't displayable in this simple graph.

\newpage
\section{Robustness of Results}
\label{app:robustness}

Here, we show our results are not artifacts of random noise or due to special choices of hyperparameters such as loss thresholds.

\subsection{Histogram of Perturbation Radii}
\label{app:histogram_radii}

Fig~\ref{fig:MNIST Perturbation Histogram} showed the distribution of distances that random perturbations travel in MNIST in the low data loss landscape. Here are histograms for our other 3 main problems.

\begin{figure}[h!]
    \centering
    % Include the combined figure
    \includegraphics[width=0.48\textwidth]{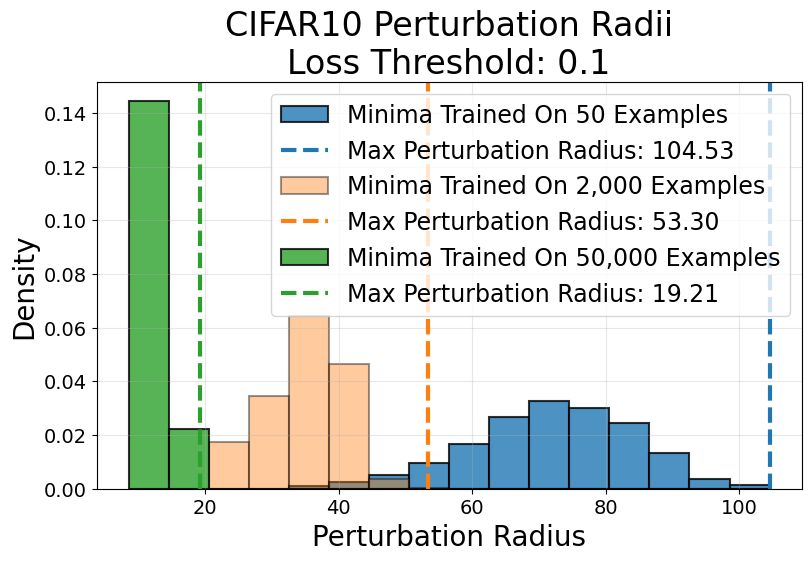}%
    \includegraphics[width=0.48\textwidth]{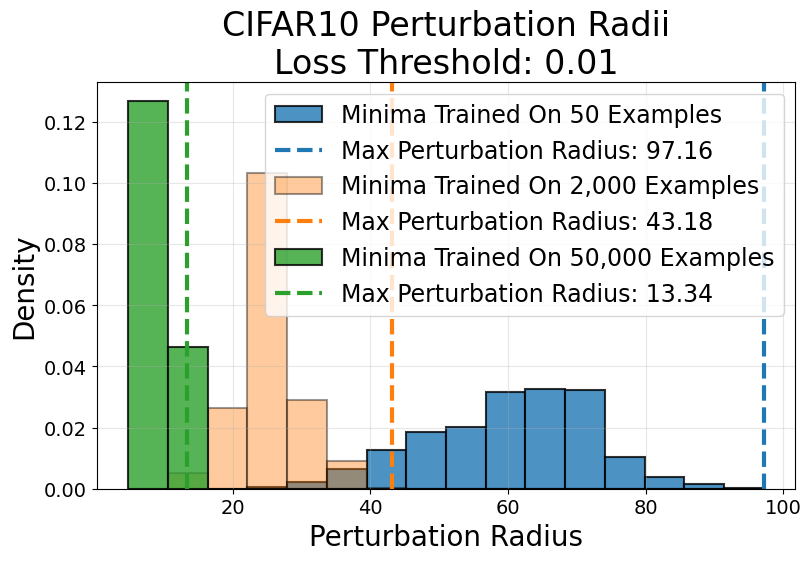}
    \includegraphics[width=0.48\textwidth]{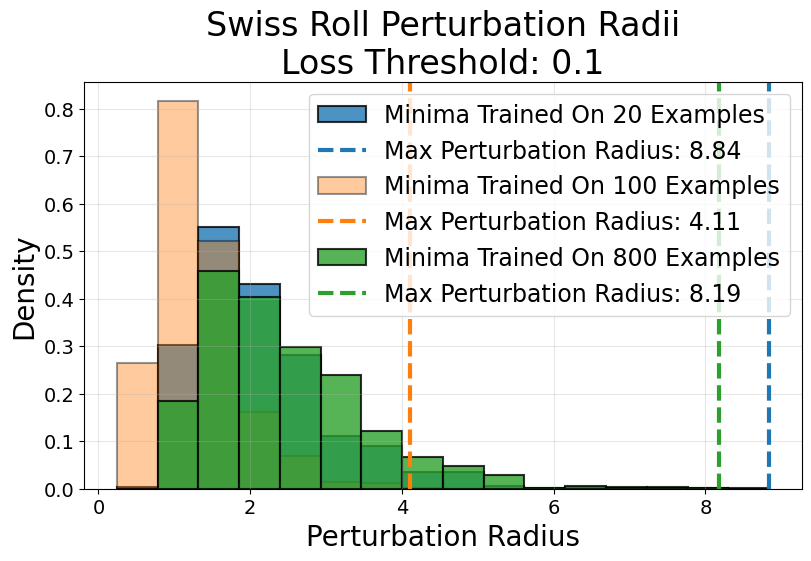}%
    \includegraphics[width=0.48\textwidth]{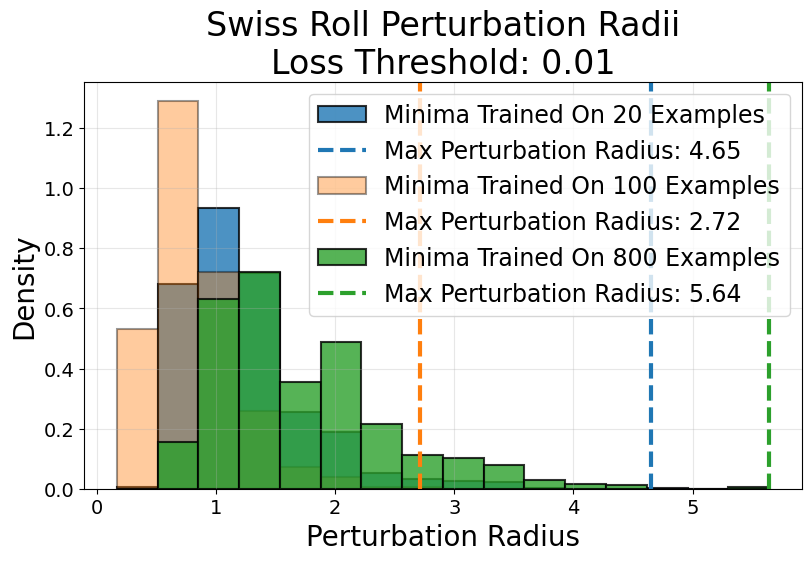}
    \includegraphics[width=0.48\textwidth]{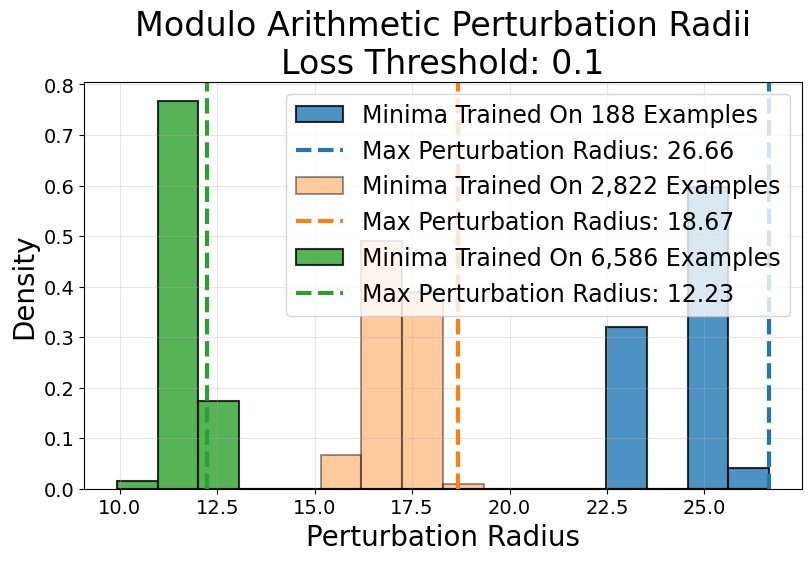}%
    \includegraphics[width=0.48\textwidth]{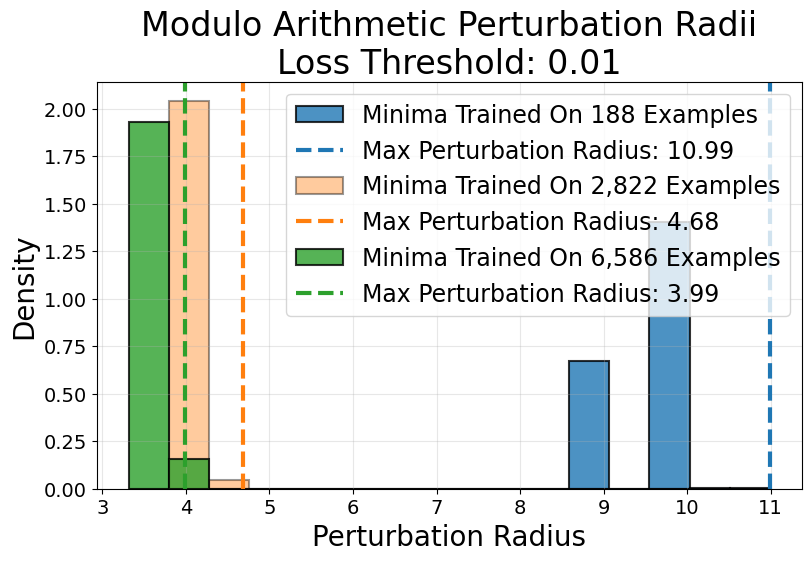}
    \caption{Distribution of distances random perturbations reach in neural network parameter space before hitting a loss threshold. The distances are evaluated in a loss landscape of 50 (CIFAR10), 20 (swiss roll) and 188 (modulo arithmetic) examples. The difference between minima trained on the largest and smallest datasets is evident in all cases except for the swiss roll, where only intermediate minima are significantly smaller. We note there is no possibility of mistaking the relative sizes of our CIFAR10 or modulo arithmetic minima, but swiss roll minima are less clear (and in fact, the choice of loss threshold can change which minima appears larger). } 
    \label{fig:Histograms CIFAR10 Swiss Modulo}
\end{figure}

\subsection{Results With 50 Perturbations}
\label{app:perturbations_50}

All main text experiments use 500 perturbations. Here, we display results from taking only 50 perturbations instead, which show the same trends as in the main text. Minima volumes appear to be relatively easy to measure, with clear differences apparent from a small number of perturbations.%This suggests minima volumes are relatively easy to measure.

\begin{figure}[h!]
    \centering
    % Include the combined figure
    \includegraphics[width=0.45\textwidth]{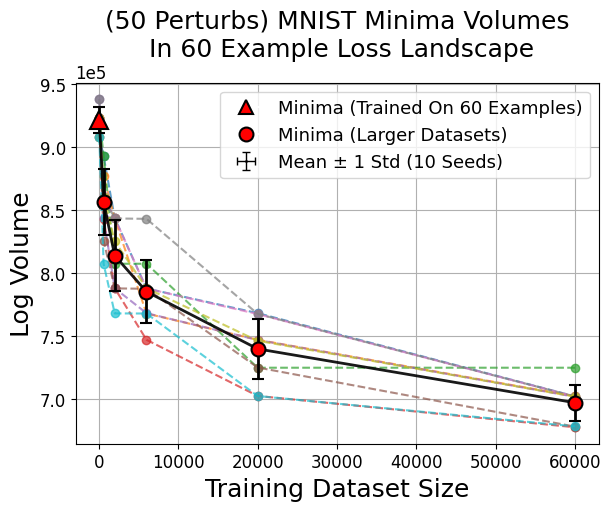}%
    \includegraphics[width=0.45\textwidth]{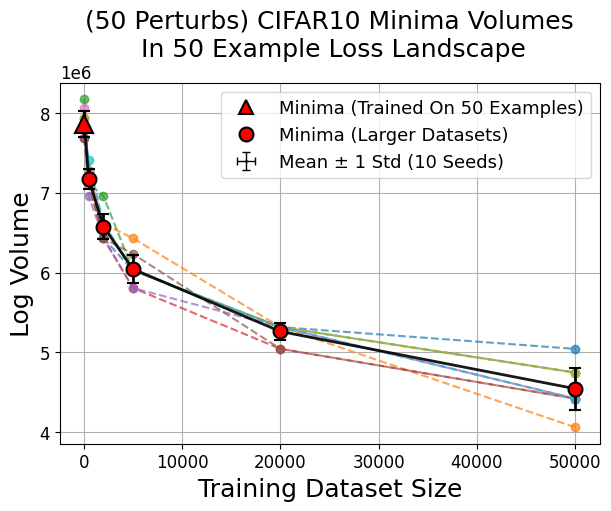}
    \includegraphics[width=0.45\textwidth]{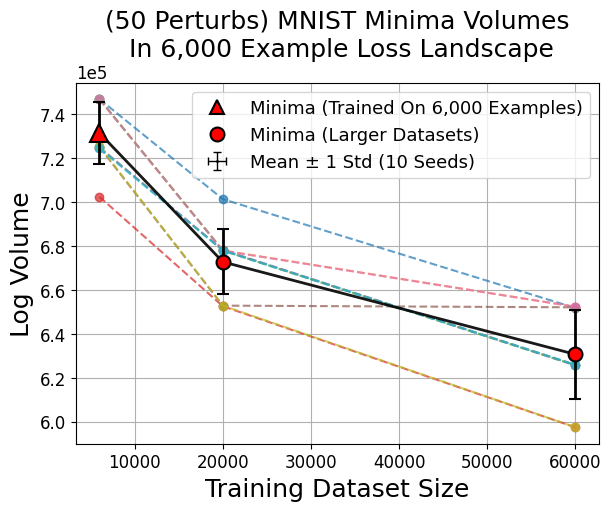}%
    \includegraphics[width=0.45\textwidth]{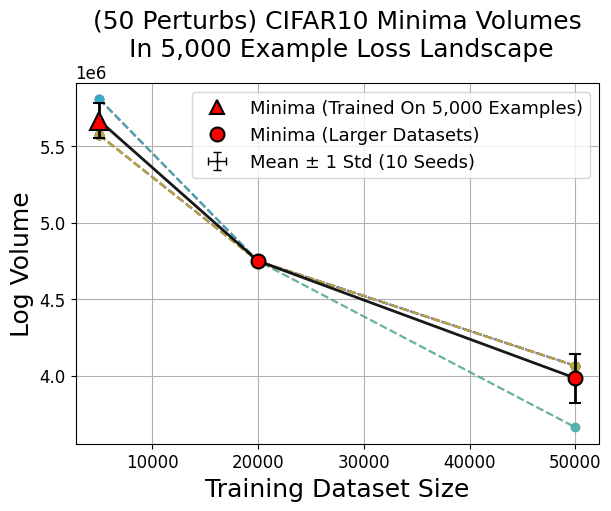}
    \includegraphics[width=0.45\textwidth]{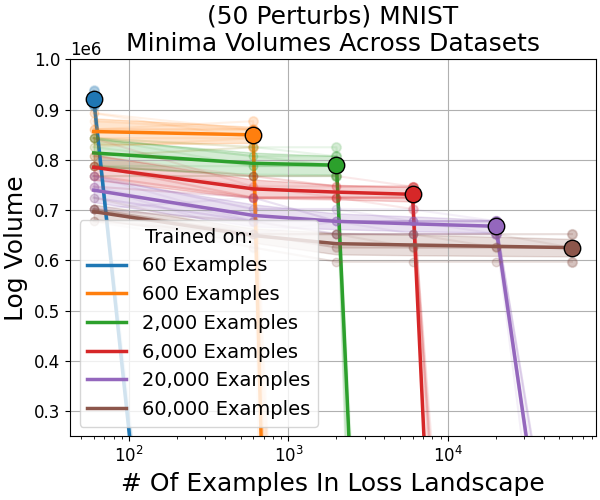}%
    \includegraphics[width=0.45\textwidth]{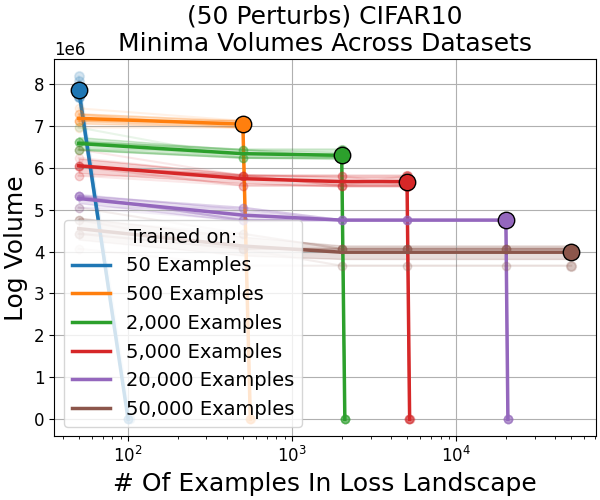}
    \caption{Results for minima volumes measured with only 50 perturbations. All volumes are smaller than with 500 perturbations (e.g., CIFAR10's largest volume is below $8 \times 10^6$ when it was larger before), which is to be expected. Similar trends are present as before.} 
    \label{fig:50 perturbations MNIST CIFAR}
\end{figure}

\begin{figure}[h!]
    \centering
    % Include the combined figure
    \includegraphics[width=0.45\textwidth]{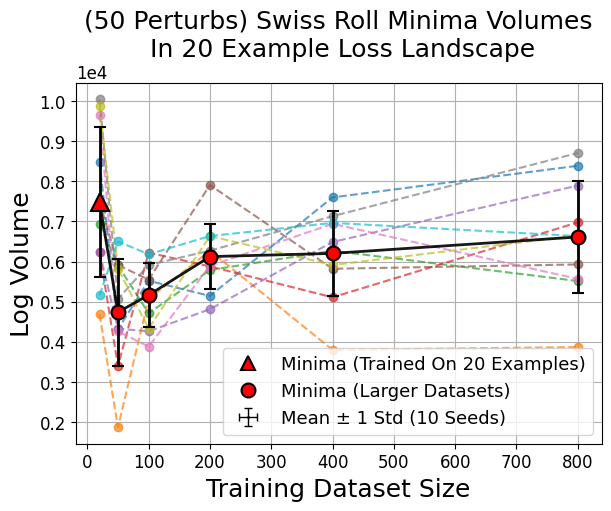}%
    \includegraphics[width=0.485\textwidth]{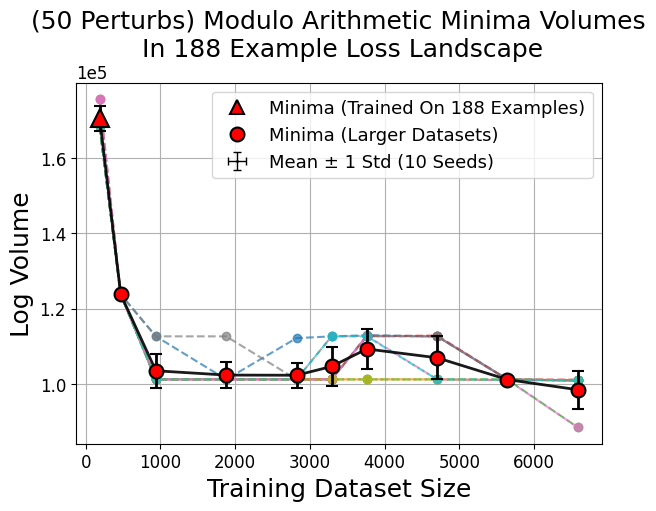}
    \includegraphics[width=0.45\textwidth]{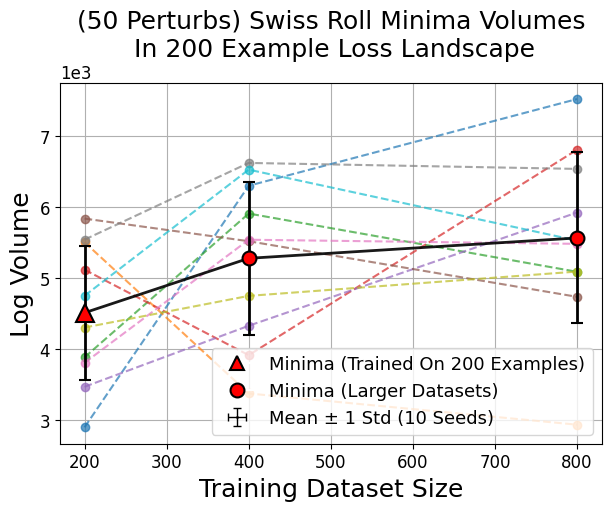}%
    \includegraphics[width=0.485\textwidth]{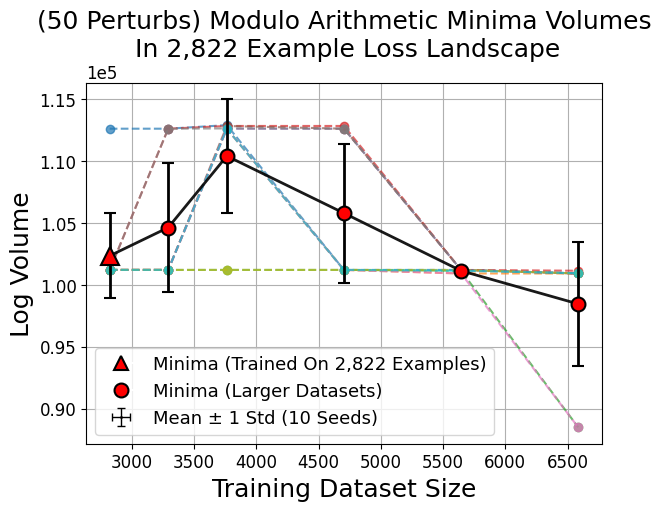}
    \includegraphics[width=0.45\textwidth]{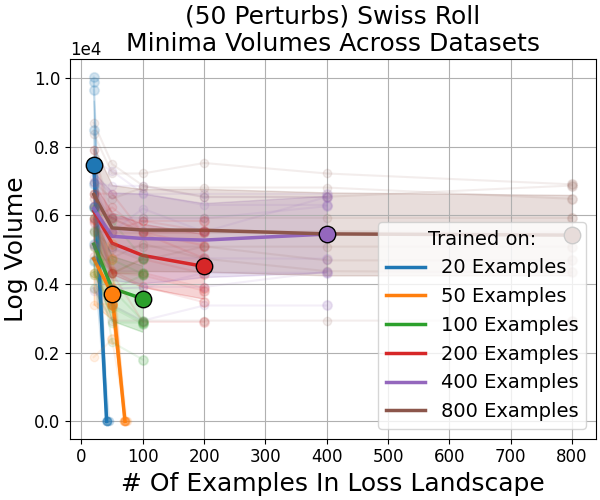}%
    \includegraphics[width=0.45\textwidth]{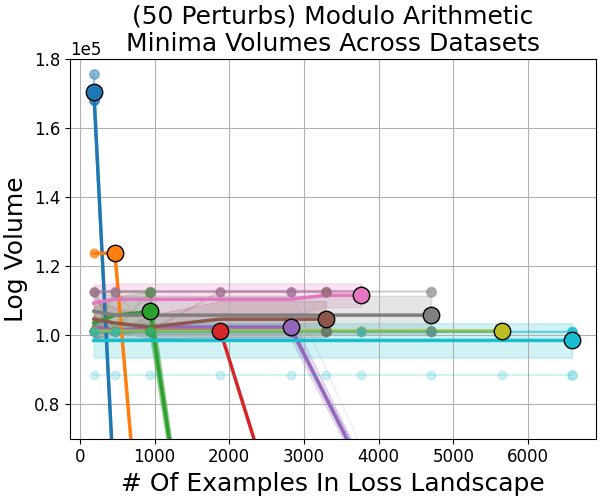}
    \caption{Results for minima volumes measured with only 50 perturbations. All volumes are smaller as expected. Similar trends are present as before.} 
    \label{fig:50 perturbations swiss modulo}
\end{figure}
\newpage 
\subsection{Different Loss Thresholds}
\label{app:loss_thresholds}

In the main text, we measured the volumes of basins with a loss threshold of 0.1 (except for modulo arithmetic where a threshold of 0.01 was used instead, since it has much smaller loss values for our network and loss function). This choice is arbitrary. Here, we show results for a smaller loss threshold, which shows the same trends.

\begin{figure}[h!]
    \centering
    % Include the combined figure
    \includegraphics[width=0.45\textwidth]{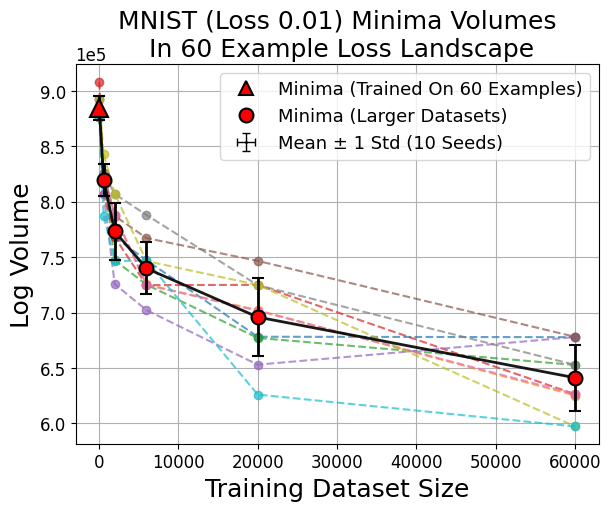}%
    \includegraphics[width=0.45\textwidth]{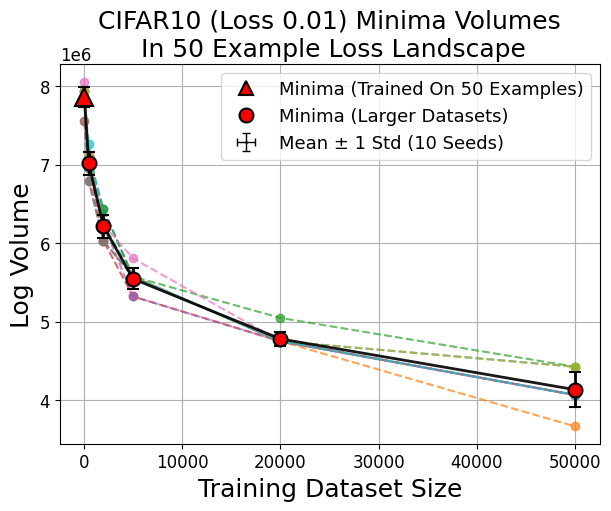}
    \includegraphics[width=0.45\textwidth]{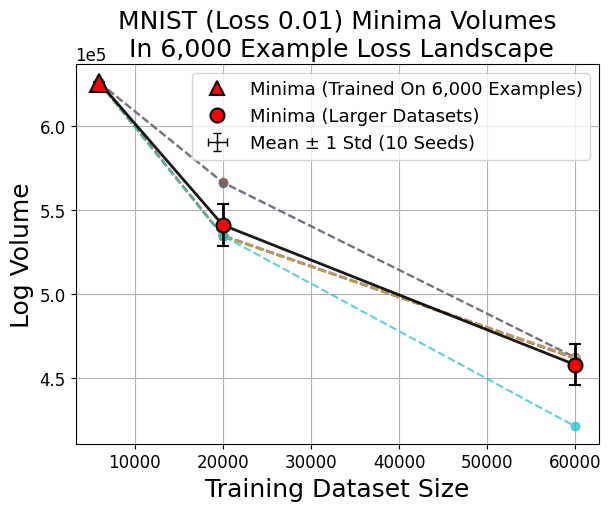}%
    \includegraphics[width=0.45\textwidth]{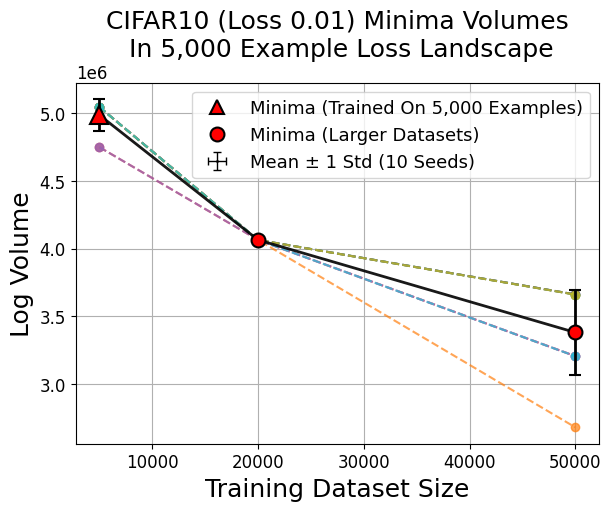}
    \includegraphics[width=0.45\textwidth]{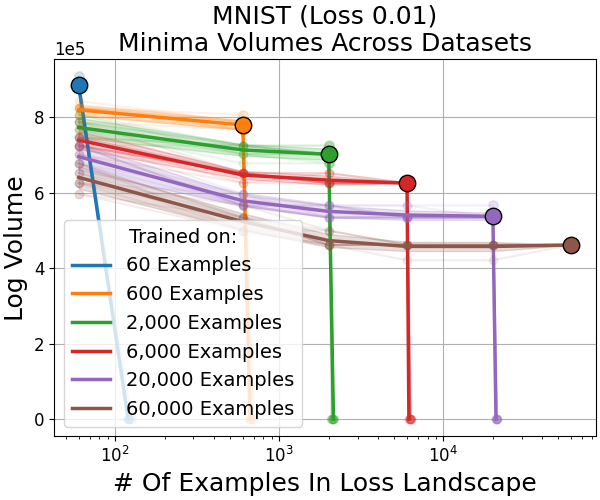}%
    \includegraphics[width=0.45\textwidth]{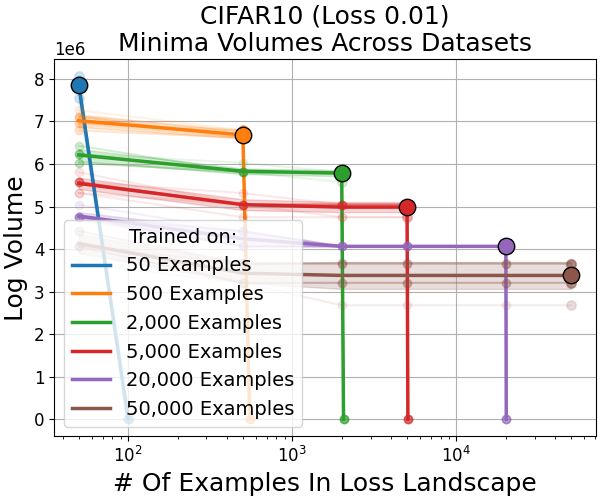}
    \caption{Results for minima volumes measured with a loss threshold of 0.01 instead of 0.1. All volumes shrink, but the same trends as before are observed. For MNIST and CIFAR, the volume-data power law is steeper, with volumes now shrinking even faster as data is added. See Appendix~\ref{app:scaling_laws} for details.} 
    \label{fig:MNIST CIFAR Loss 0.01}
\end{figure}

\begin{figure}[h!]
    \centering
    % Include the combined figure
    \includegraphics[width=0.45\textwidth]{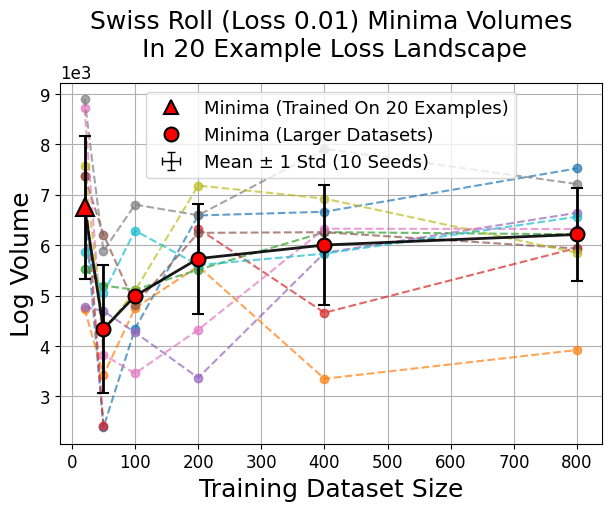}%
    \includegraphics[width=0.485\textwidth]{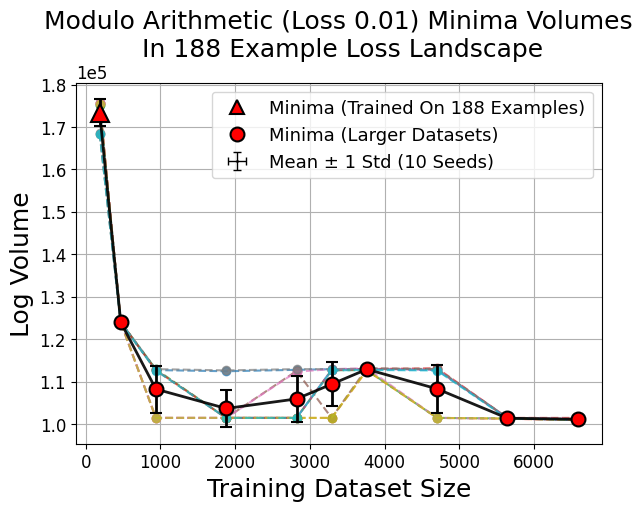}
    \includegraphics[width=0.45\textwidth]{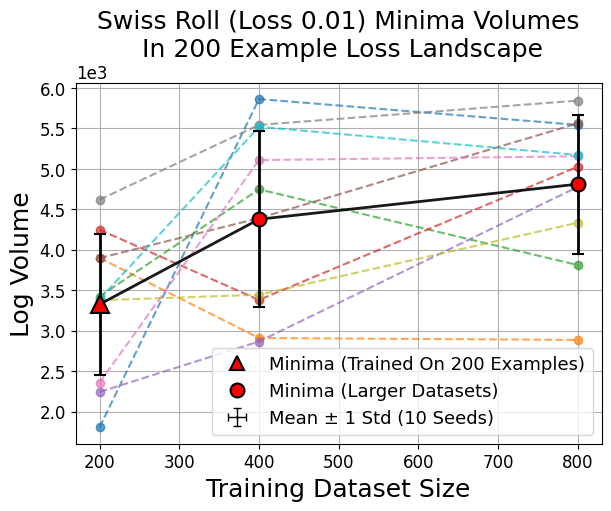}%
    \includegraphics[width=0.485\textwidth]{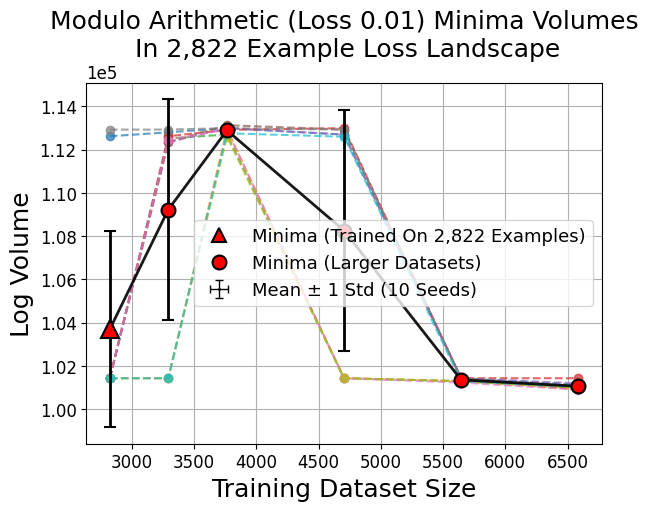}
    \includegraphics[width=0.45\textwidth]{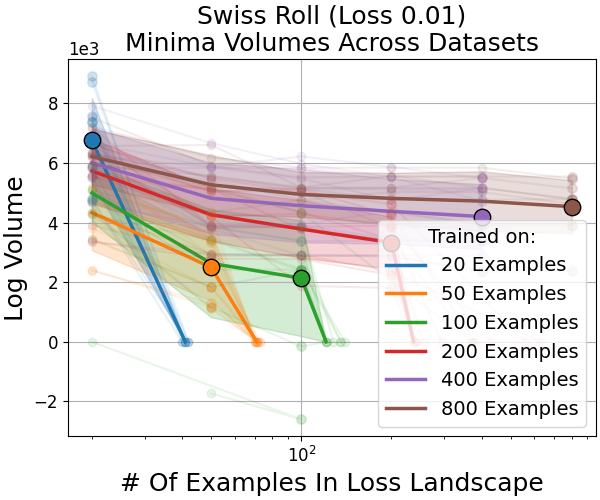}%
    \includegraphics[width=0.45\textwidth]{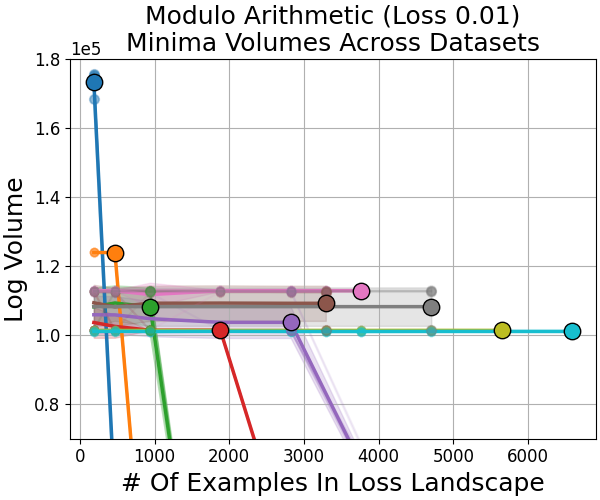}
    \caption{Results for minima volumes measured with a loss threshold of 0.01 instead of 0.1 for swiss roll, and 0.001 instead of 0.01 for modulo arithmetic. All volumes shrink, but the same trends as before as observed. An issue with extremely small loss thresholds is that models may struggle to reach the desired loss, resulting in significantly more noise in volumes (e.g., swiss roll).} 
    \label{fig:swiss modulo Loss 0.01}
\end{figure}

\clearpage 
\section{Modified Architectures and Training Methods}
\label{app:architectures}

Here we display results for different architectures and different training methods. Our results are robust to all these cases.

\subsection{Convolutional Neural Networks - MNIST, CIFAR}
\label{app:cnn_results}

\begin{figure}[h!]
    \centering
    % Include the combined figure
    \includegraphics[width=0.45\textwidth]{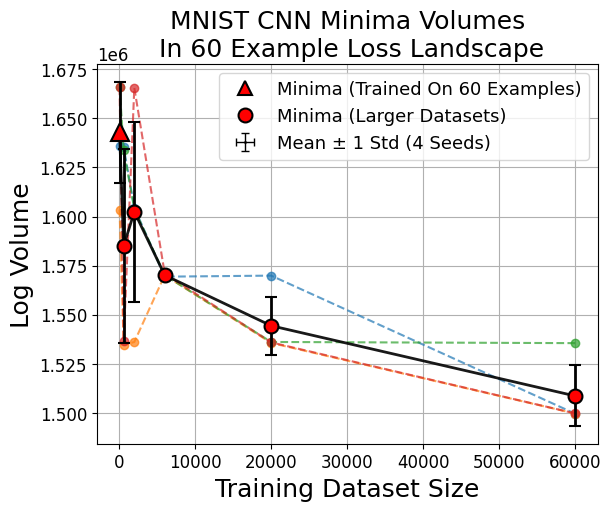}%
    \includegraphics[width=0.45\textwidth]{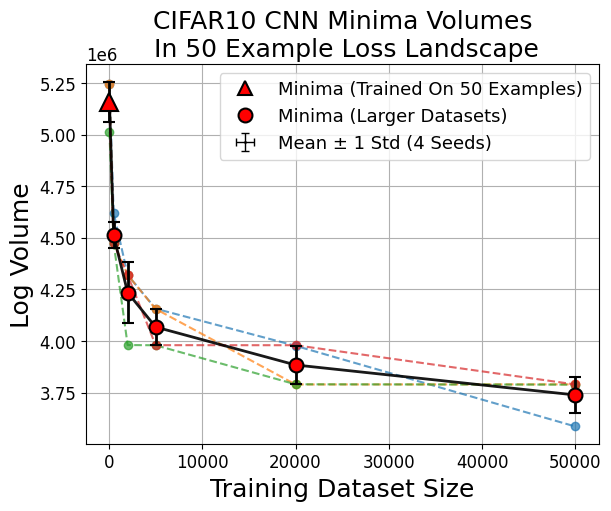}
    \includegraphics[width=0.45\textwidth]{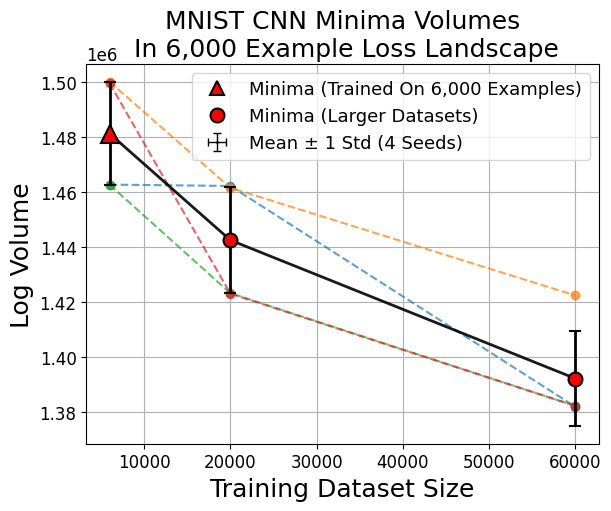}%
    \includegraphics[width=0.45\textwidth]{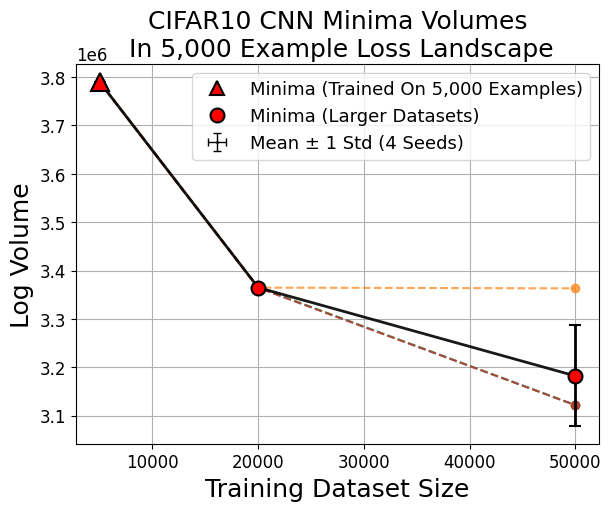}
    \includegraphics[width=0.45\textwidth]{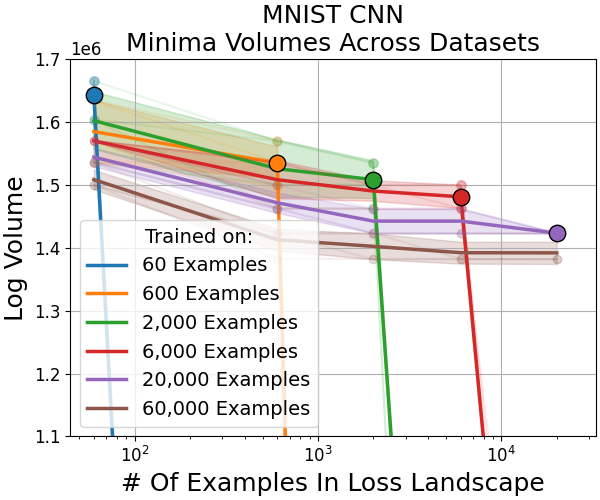}%
    \includegraphics[width=0.45\textwidth]{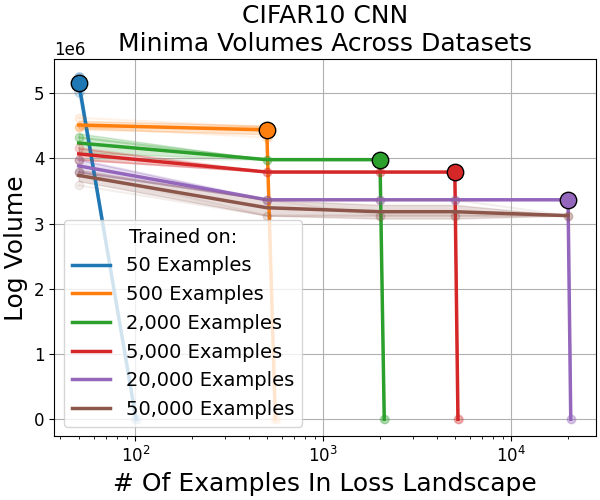}
    \caption{Results for minima volumes for a convolutional neural network instead of an MLP. Similar trends are observed. Due to computational limitations, we skip evaluating the volumes on the loss landscape formed from the entire dataset, and lower the number of random seeds. The disparity in volumes seems smaller than for an MLP, as quantified by the power law relating the volume found at a given dataset size. We speculate this suggests good inductive biases make previously sharp minima more attractive from a volume perspective.
    } 
    \label{fig:MNIST CIFAR CNN}
\end{figure}

\subsection{Larger Networks, No Filternorm, SGD}
\label{app:model_variations}

We display results for a larger network for both swiss roll and MNIST, and for MNIST without filter normalization and with SGD.

\begin{figure}[h!]
    \centering
    % Include the combined figure
    \includegraphics[width=0.45\textwidth]{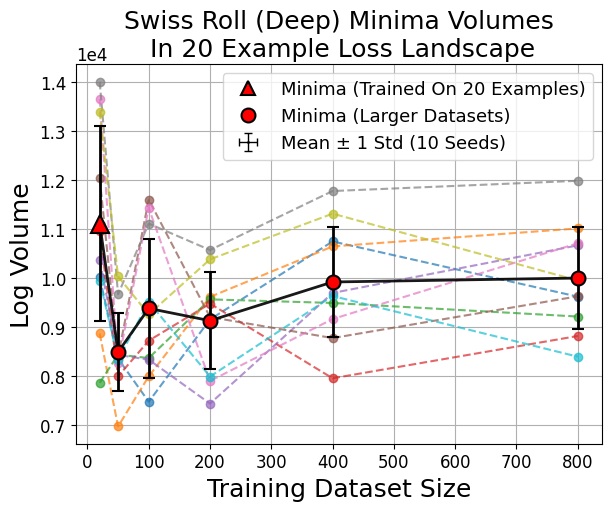}%
    \includegraphics[width=0.45\textwidth]{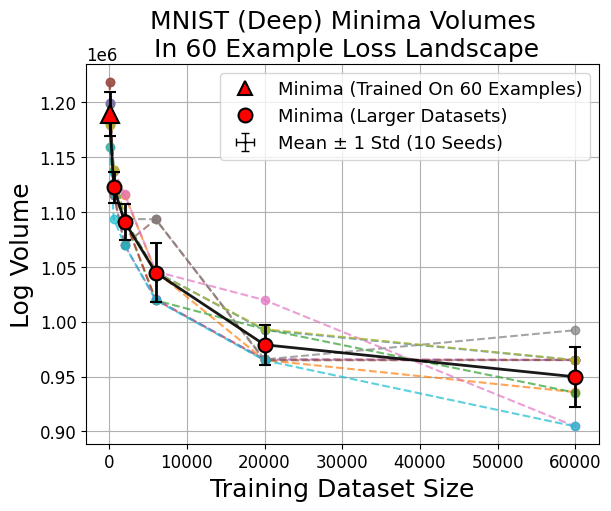}
    \includegraphics[width=0.45\textwidth]{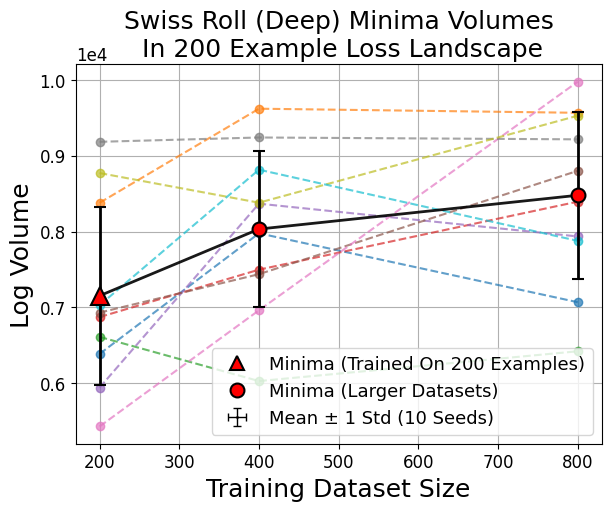}%
    \includegraphics[width=0.45\textwidth]{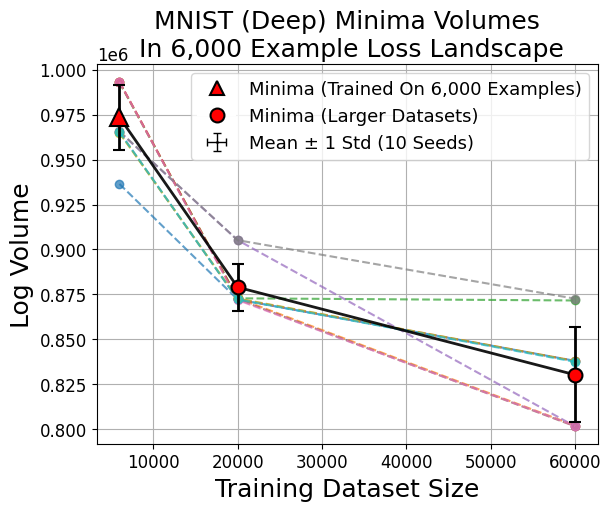}
    \includegraphics[width=0.45\textwidth]{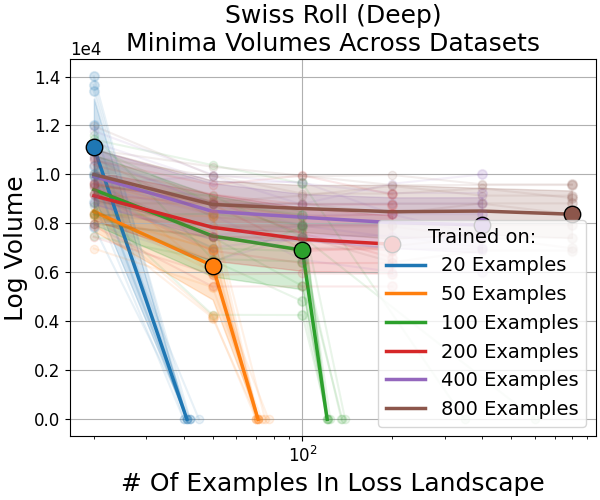}%
    \includegraphics[width=0.45\textwidth]{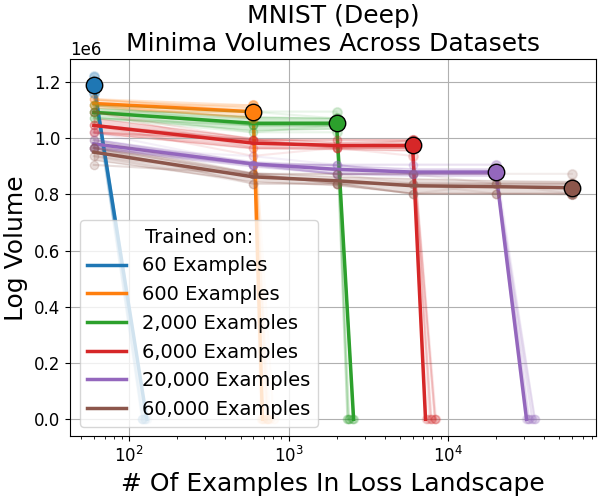}
    \caption{Results for minima volumes for larger networks than displayed in the main text. The increased number of parameters results in larger volumes overall. The same trends are observed as before. Specifically, the swiss roll increased from an MLP with 5 hidden layers of 32 neurons to 6 hidden layers of 32 neurons. MNIST went from 2 hidden layers of 256, 128 neurons to 3 layers of 256, 256, 128 neurons, respectively.
    } 
    \label{fig:swiss MNIST large model}
\end{figure}

\begin{figure}[h!]
    \centering
    % Include the combined figure
    \includegraphics[width=0.45\textwidth]{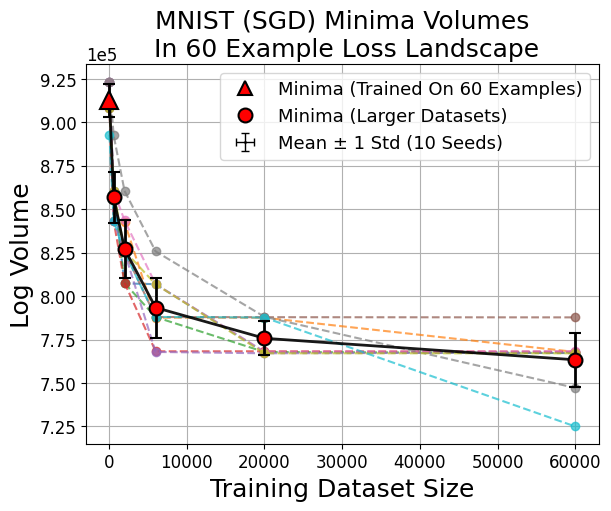}%
    \includegraphics[width=0.45\textwidth]{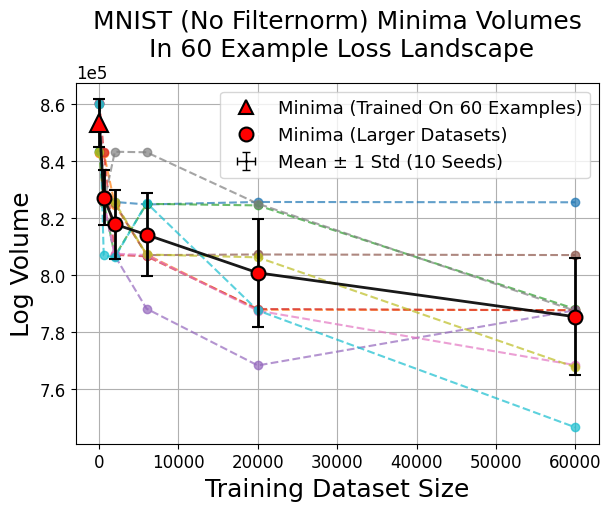}
    \includegraphics[width=0.45\textwidth]{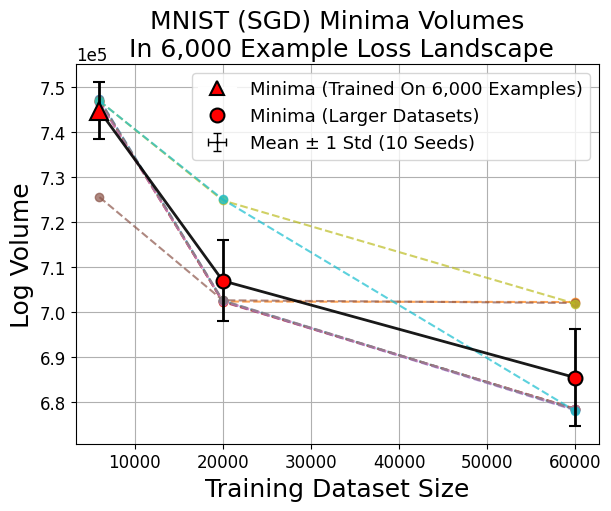}%
    \includegraphics[width=0.45\textwidth]{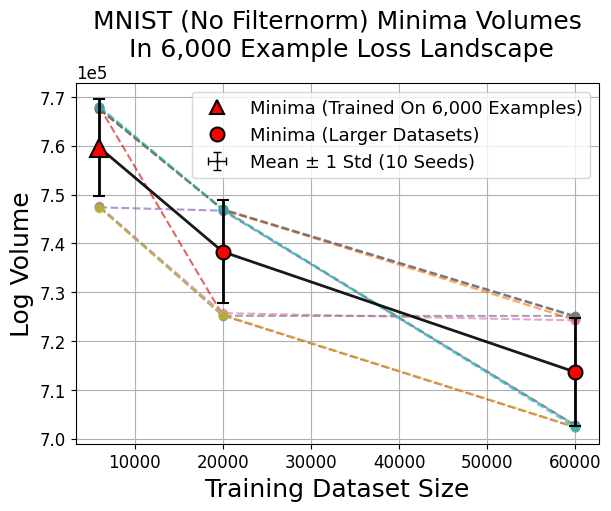}
    \includegraphics[width=0.45\textwidth]{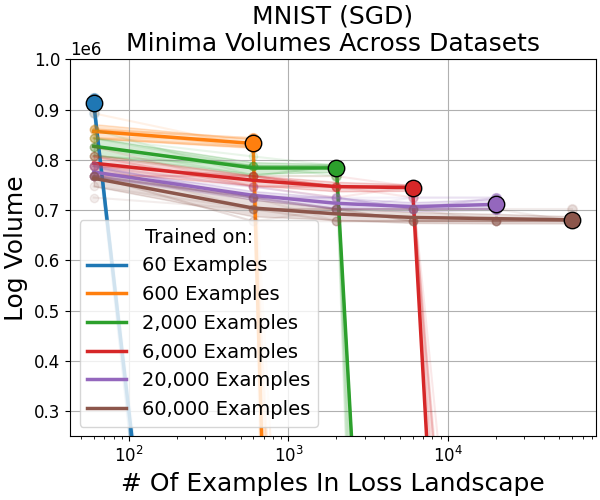}%
    \includegraphics[width=0.45\textwidth]{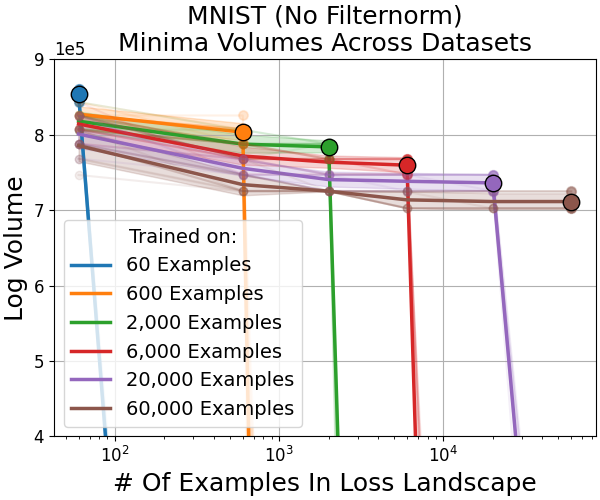}
    \caption{Results for minima volumes in MNIST with SGD (left) and without filter normalization (right). With SGD, volumes are generally smaller but the same trends appear. SGD tends to generalize better and appears likelier to find sharper minima as indicated by the volume-data power law coefficient (see Appendix~\ref{app:scaling_laws}). 
    When filter normalization is removed, the same qualitative trends persist but variability across experiments increases. This suggests that while filter normalization improves stability, our core findings are not an artifact of its use.
    } 
    \label{fig:MNIST filter SGD}
\end{figure}

\subsection{MNIST With More Training Epochs}
\label{app:mnist_epochs}

In modulo arithmetic, more training epochs resulted in a large decrease in volume. For comparison, training MNIST for additional epochs results in very slight changes in volume and negligible differences in test accuracy.

\begin{figure}[h!]
    \centering
    % Include the combined figure
    \includegraphics[width=0.45\textwidth]{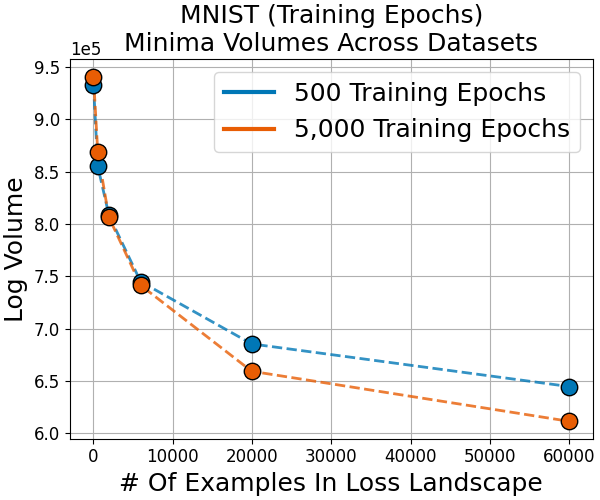}%
    \includegraphics[width=0.45\textwidth]{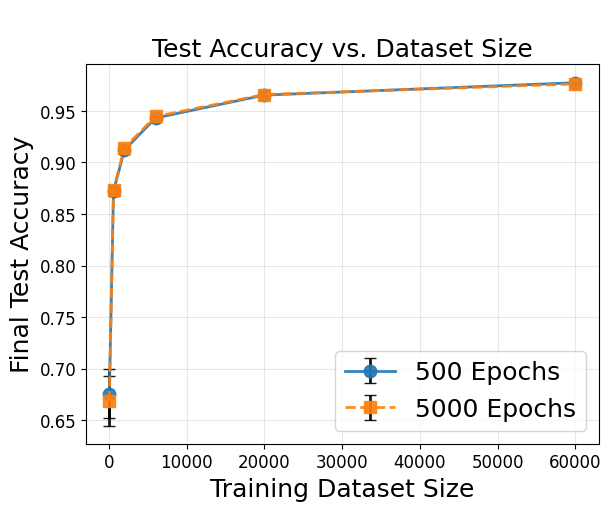}
    \caption{Training MNIST for significantly more epochs does not result in large volume changes, unlike in modulo arithmetic. It also results in negligible changes in test accuracy. 
    } 
    \label{fig:MNIST epochs}
\end{figure}

%\subsection{Swiss Roll -- Deeper Architectures}
%\label{app:swissroll_deep}
% We have this but it's not interesting. Let's skip.

\subsection{Modulo Arithmetic -- Grokking At Other Data Sizes}
\label{app:modulo_grokking}

In Fig~\ref{fig:Grokking Volumes}, we showed the loss and accuracy over epochs, along with the volumes for a modulo arithmetic model trained with $\approx 33\%$ of the data. Grokking also occurs at smaller training datasets, with the time to generalization increasing. Here, we display the curves and the volumes.

\begin{figure}[h!]
    \centering
    % Include the combined figure
    \includegraphics[width=0.4\textwidth]{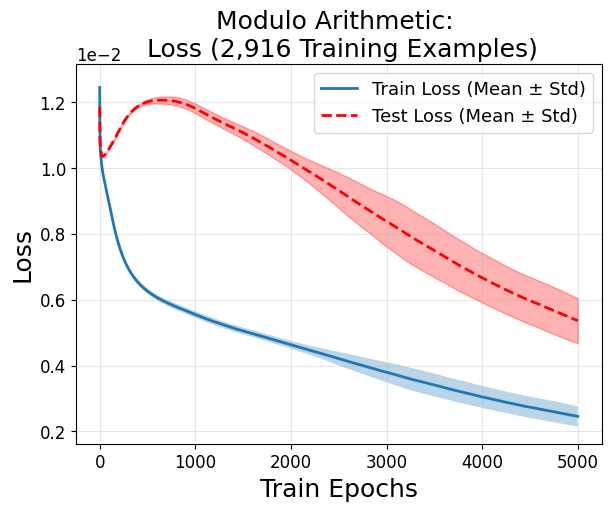}%
    \includegraphics[width=0.4\textwidth]{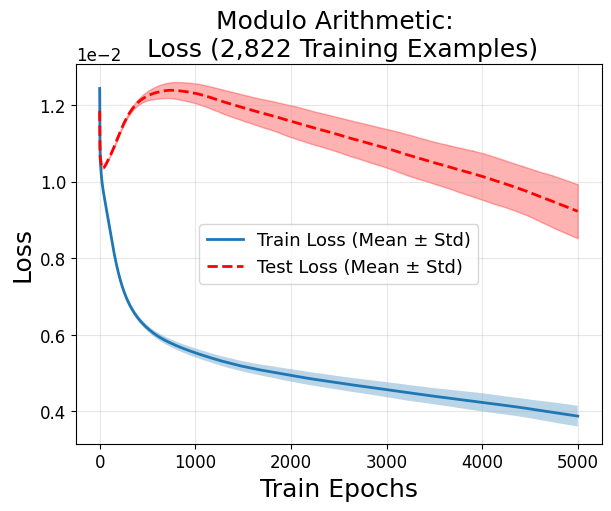}
    \includegraphics[width=0.4\textwidth]{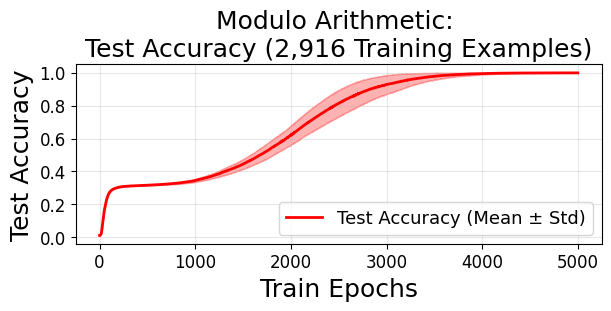}%
    \includegraphics[width=0.4\textwidth]{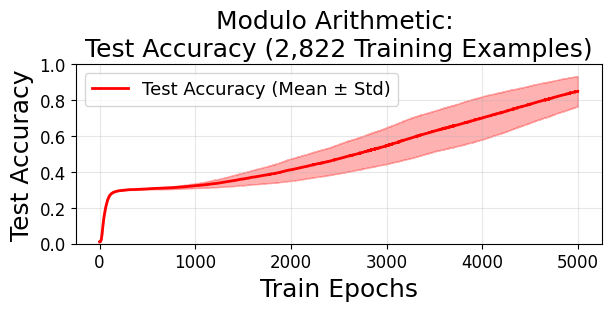}
    \includegraphics[width=0.4\textwidth]{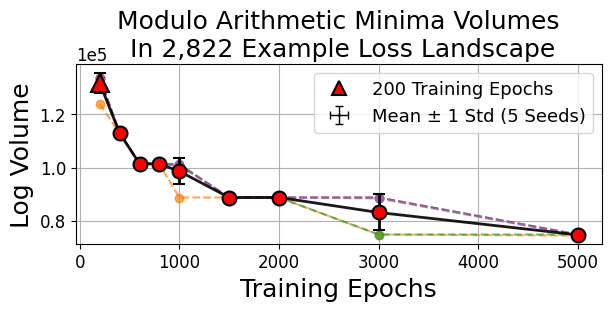}%
    \includegraphics[width=0.4\textwidth]{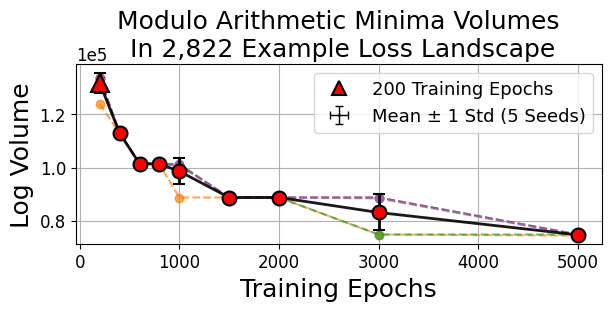}
    \caption{Training modulo arithmetic on $31 \%$ (left) and $30\%$ (right) of the data still results in grokking, with increased epochs to generalization. Volumes decrease steadily throughout the grokking process.
    } 
    \label{fig:modulo grokking other}
\end{figure}

\section{Batch Sizes and Fixed Data Experiments}
\label{app:batch and model seed}
In the main text, we largely studied the effect of dataset size on minima volume and test accuracy. Here, we vary batch size and test 50 random model parameter initializations on a fixed dataset.
% This is where the random seed work should go! I think the figure caption should mention how much variability an be attributed purely to random seeds, in contrast with earlier results.

% I'll add my batch size figures here showing that minibatches correlate with test accuracy.

\begin{figure}[h!]
    \centering
    % Include the combined figure
    \includegraphics[width=0.45\textwidth]{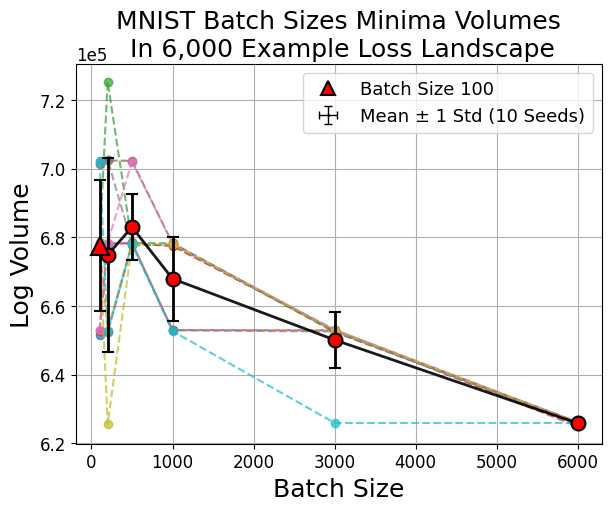}%
    \includegraphics[width=0.45\textwidth]{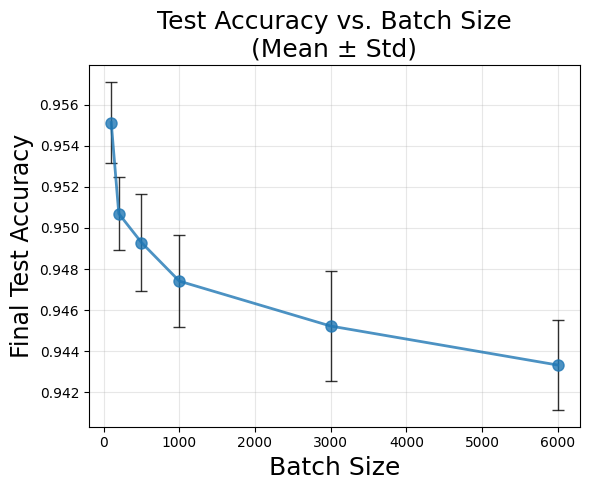}
    \caption{Training on MNIST with a fixed number of examples and different batch sizes (for 10 different model and data split seeds) yields a variety of volumes. In line with the work of Keskar et al \cite{keskar_large-batch_2017}, we find that large batch gradient descent tends to find smaller minima. We also find the test accuracy is inversely correlated with batch size. This trend is different when comparing minima from larger dataset sizes, where flatness and generalization appear inversely correlated instead.} 
    \label{fig:batch size}
\end{figure}

Figure~\ref{fig:batch size} reproduces the known effect of larger batch sizes typically finding sharp minima with worse test accuracy~\cite{keskar_large-batch_2017, he_control_nodate}. The increased noise of small-batch training pushes parameters into larger-volume minima, which, in a given loss landscape, tend to generalize better.

\pagebreak

\begin{figure}[h!]
    \centering
    % Include the combined figure
    \includegraphics[width=0.48\textwidth]{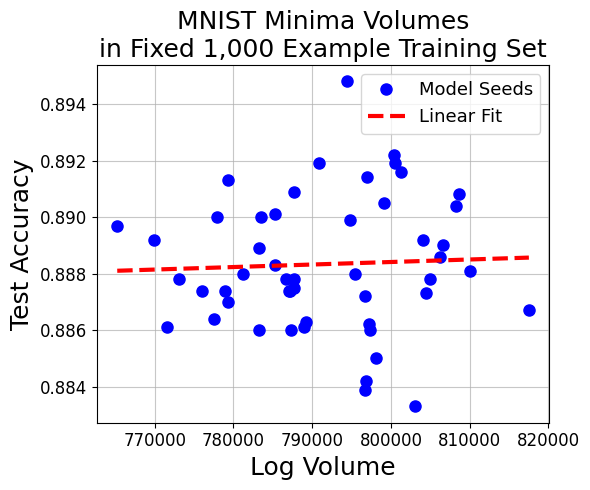}%
    \includegraphics[width=0.48\textwidth]{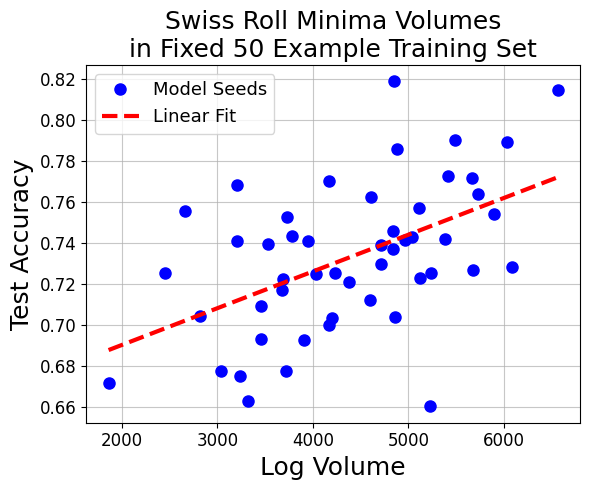}
    \includegraphics[width=0.48\textwidth]{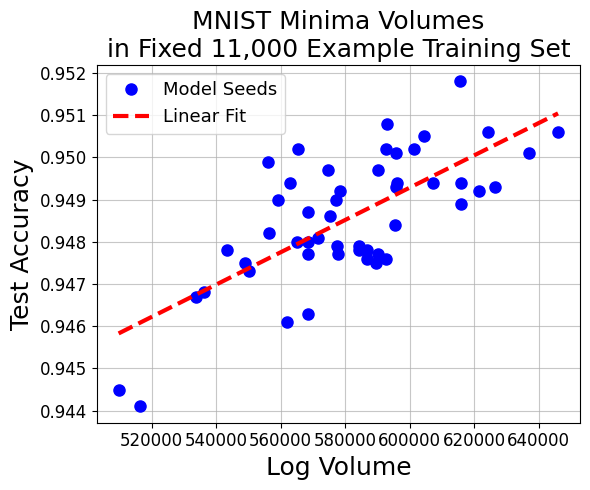}%
    \includegraphics[width=0.48\textwidth]{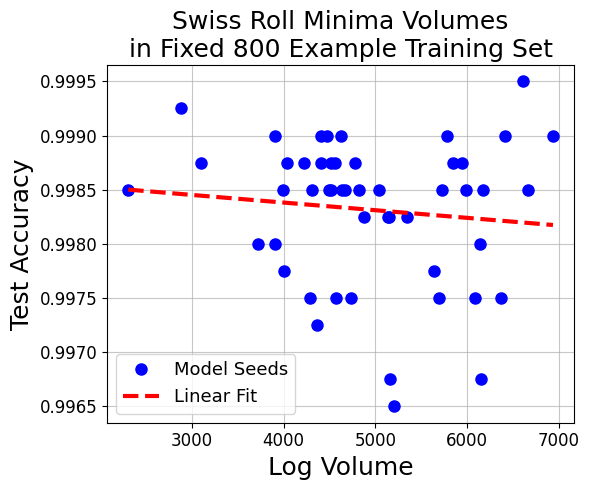}%
    \caption{Minima volume vs. test accuracy for models trained on fixed datasets---1,000 and 11,000 training examples for MNIST (left) and 50 and 800 training examples for swiss roll (right). The impact of volume on test accuracy is noisy and depends on problem and training dataset size.}% Compare to Fig~\ref{fig:Sharpness Aware Minimization}, where a similar picture arises for these problems in the accuracy gap between AdamW and SAM, varying dataset size.} 
    \label{fig:FMH}
\end{figure}

In Figure~\ref{fig:FMH}, we consider the effects of random model initializations explicitly. We fix a data landscape and plot the volume vs. test accuracy for 50 model seeds (for computational reasons, these experiments use only 50 perturbations, but Appendix~\ref{app:perturbations_50} suggests the results are similar for more perturbations).

%We find similar results to experiments with different optimizers in Fig~\ref{fig:Sharpness Aware Minimization}. The variability of test accuracies obtained with AdamW and SAM is similar to the variability from random initializations, and the gaps in volume show similar trends. Note the volumes in Fig~\ref{fig:Sharpness Aware Minimization} and Fig~\ref{fig:FMH} are not directly comparable, as the former uses 500 random perturbations to measure volume, whereas the latter uses 50. More perturbations results in larger volumes due to the nature of our measurement method.

%In MNIST,
%Figure~\ref{fig:FMH} considers this latter setting explicitly: we keep a fixed data landscape and plot volume vs. test accuracy from 50 random model initializations (note also, for computational reasons, in these experiments we use only 50 perturbations, but Appendix~\ref{app:perturbations_50} shows the difference should be minor). For MNIST
 %we see a stronger relationship in the larger training dataset setting. In swiss roll, we see the opposite. 
 %The picture is similar to Fig~\ref{fig:Sharpness Aware Minimization}, comparing SAM and AdamW. Specifically, the gap between AdamW and SAM in test accuracy at different training dataset sizes shows the potential accuracy difference between minima of varying volumes, whereas the distance between AdamW and SAM volume lines (and the width of the AdamW line) show the variance in obtainable volumes. 
%There are parallels between Fig~\ref{fig:FMH} and Fig~\ref{fig:Sharpness Aware Minimization}, comparing SAM and AdamW.
We find similar results to experiments with different optimizers in Fig~\ref{fig:Sharpness Aware Minimization}. Specifically, we observe a larger gap between SAM and AdamW in test accuracy and minima volume correlates with a wider variation in test accuracy and minima volume amongst random model seeds of AdamW. Wider variation in accuracy and minima volume correlates with a stronger relationship between the two in Fig~\ref{fig:FMH}.
Note the volumes in Fig~\ref{fig:Sharpness Aware Minimization} and Fig~\ref{fig:FMH} are not directly comparable, as the former uses 500 random perturbations to measure volume, whereas the latter uses 50. More perturbations results in larger volumes due to the nature of our measurement method.
%The picture follows the gap between SAM and AdamW in Fig~\ref{fig:Sharpness Aware Minimization}.

An overall trend suggests flat minima matter most when dataset size is large enough to be representative of the true distribution loss landscape (that is, the distribution as data size approaches infinity) but small enough that models obtain varying accuracy levels.

\begin{figure}[h!]
    \centering
    % Include the combined figure
    \includegraphics[width=0.48\textwidth]{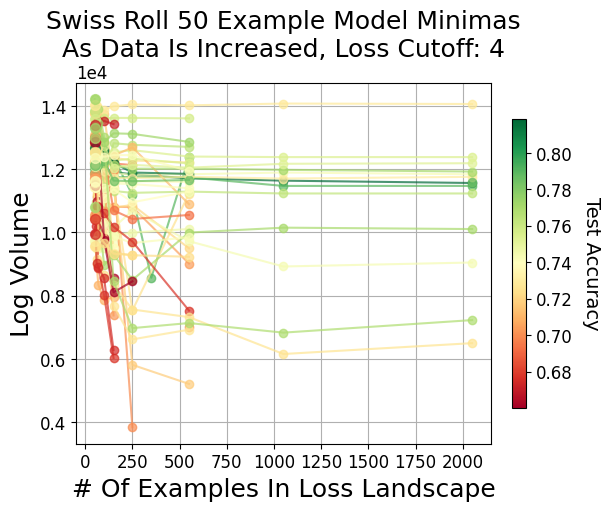}%
    \includegraphics[width=0.48\textwidth]{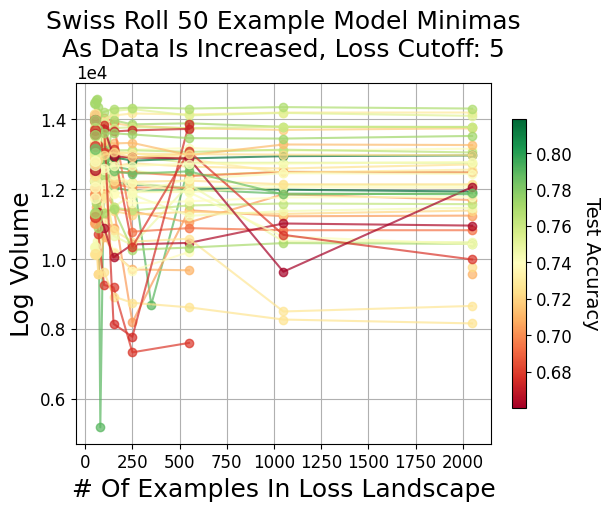}%
    \caption{Adding data to the 50 example swiss roll models show how volumes change as we approach the true distribution loss landscape. Models with worse test accuracy seem to decrease faster in volume, falling below the loss threshold (left) and are much more volatile in volume (right).
    %We see that models with worse test accuracy quickly decrease in volume, falling below the loss threshold (left) and exhibit smaller and more volatile volumes than better generalizing minima (right). 
    The loss thresholds here are significantly larger than the rest of our experiments, since loss quickly spikes when models are tested on data outside of their training set.} 
    \label{fig:FMH2_swiss}
\end{figure}

\pagebreak

\begin{figure}[h!]
    \centering
    % Include the combined figure
    \includegraphics[width=0.48\textwidth]{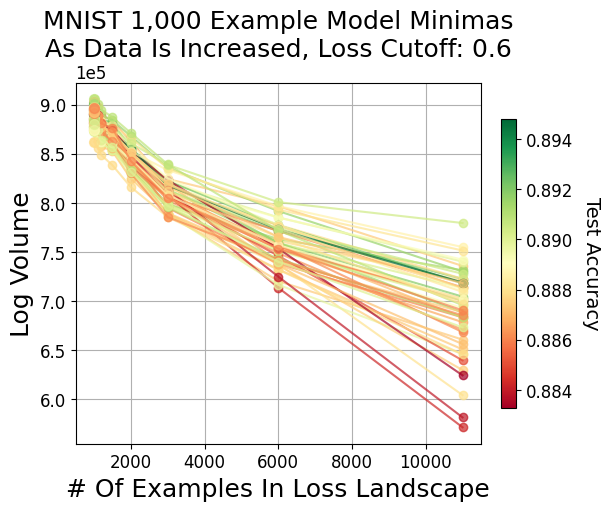}%
    \includegraphics[width=0.48\textwidth]{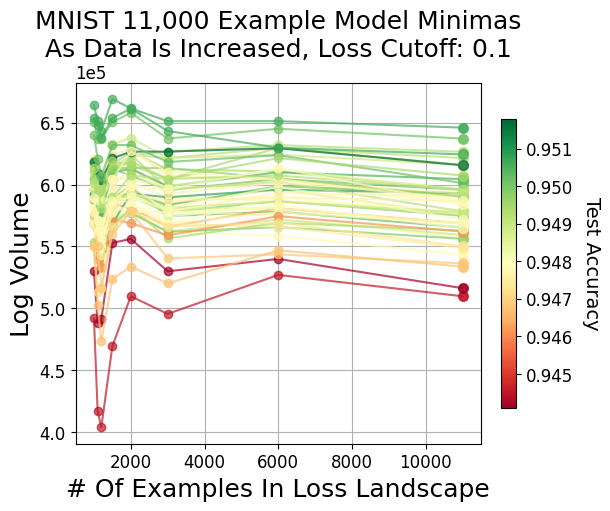}%
    \caption{\textbf{Left: }We add data to the 1,000 example MNIST models. Here we see volumes steadily decreasing, with models with higher test accuracy generally occupying slightly larger volumes. %, a trend that largely strengthens as data is added. 
    This trend becomes more apparent as additional data is added.
    \textbf{Right: }We train 11,000 example MNIST models and plot their volumes on subsets of their training set. There is an interesting volume-data trend between minima from different initializations, which we speculate originates from the minima learning similar features from the original dataset.
    %The uniformity of the lines suggest similar functions are learned despite different initializations.
    %We also see a stronger impact of the flat minima hypothesis (particularly in contrast to Fig~\ref{fig:FMH}, left), where lines are stacked according to both volume and test accuracy.
    There also seems to be a strong relationship between minima volume and test accuracy for these minima (visualized in Fig~\ref{fig:FMH}, bottom left), shown by high-accuracy green lines consistently above low-accuracy red ones.
    %We also see a strong relationship between the minima volume and the test accuracy shown by the 
    } 
    \label{fig:FMH2_mnist}
\end{figure}

We also consider the effects of different data sizes on the picture from Figure~\ref{fig:FMH}. In Figure~\ref{fig:FMH2_swiss} and Figure~\ref{fig:FMH2_mnist} (left), models trained on the same data and only varying in initializations are exposed to data they have not been trained on and rapidly shrink. In Figure~\ref{fig:FMH2_mnist} (right), randomly initialized models trained on the same large datasets are evaluated for their volumes in smaller landscapes.

%In Figures~\ref{fig:FMH2_swiss} and~\ref{fig:FMH2_mnist}, we add an extra dimension to the picture from Figure~\ref{fig:FMH}: increasing data. Volumes decrease with added data, particularly for models that achieve low test accuracy. Figure~\ref{fig:FMH2_mnist}, right, is an exception as the models are trained on all the data that is added.

\clearpage
\section{SVHN and Fashion MNIST}
\label{app:svhn_fmnist}
Here we display the results of experiments with SVHN and Fashion MNIST.

\begin{figure}[h!]
    \centering
    % Include the combined figure
    \includegraphics[width=0.45\textwidth]{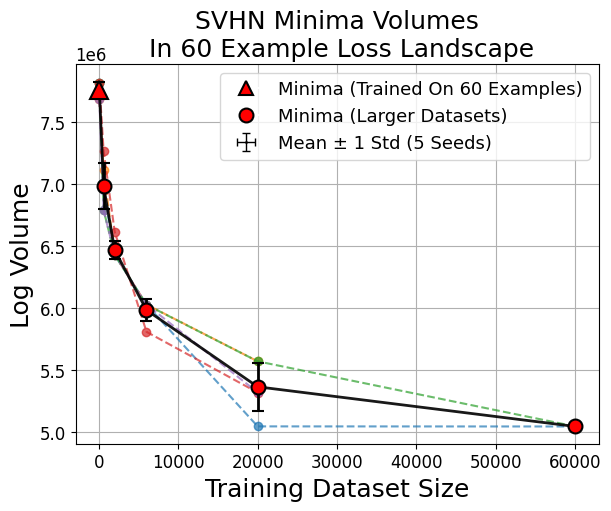}%
    \includegraphics[width=0.45\textwidth]{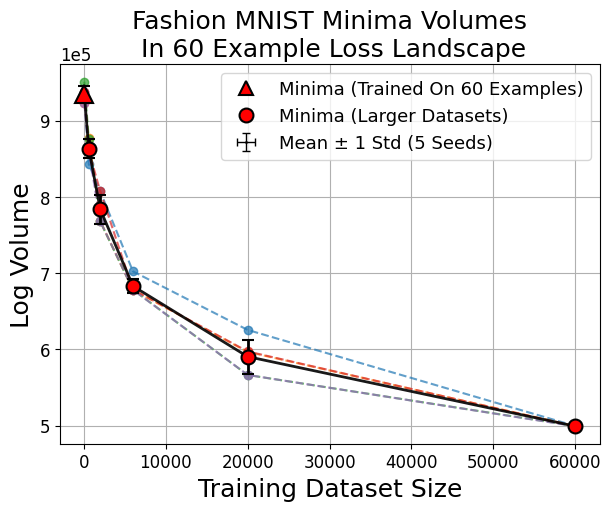}
    \includegraphics[width=0.45\textwidth]{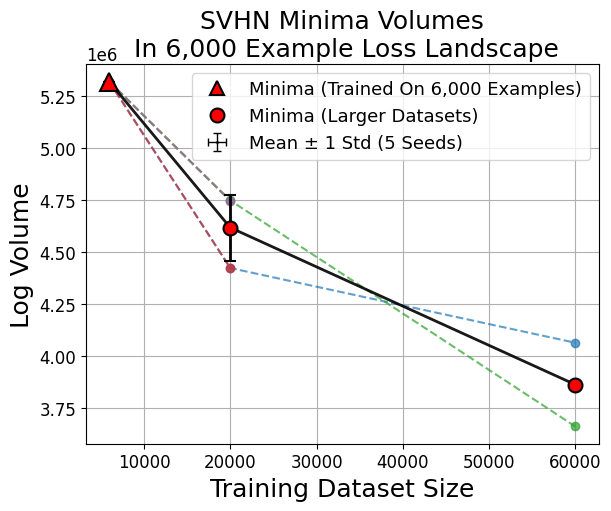}%
    \includegraphics[width=0.45\textwidth]{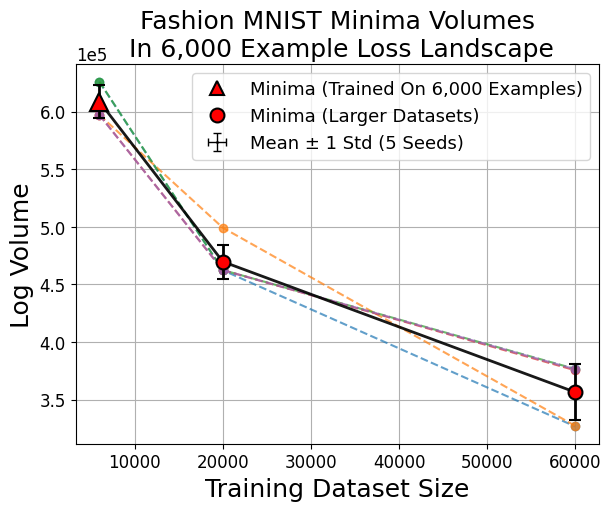}
    \includegraphics[width=0.45\textwidth]{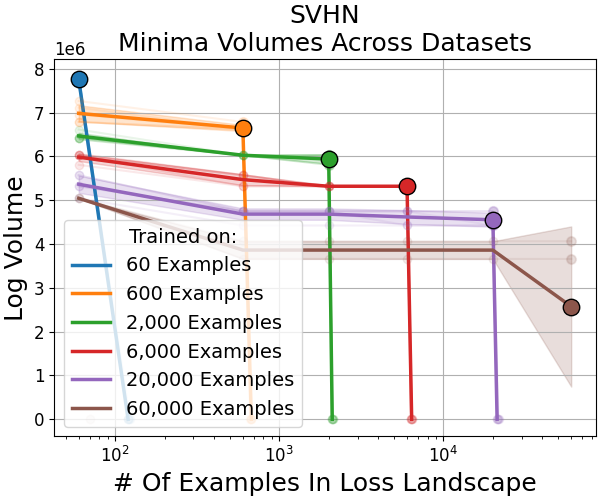}%
    \includegraphics[width=0.45\textwidth]{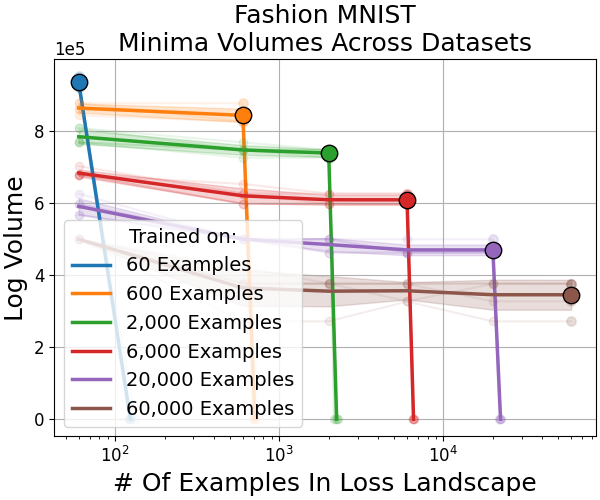}
    \caption{Results for minima volumes in SVHN (left) and Fashion MNIST (right). The same trends with our experiments from MNIST and CIFAR10 are observed, with minima volumes following a predictable power law. For SVHN at 60,000 examples, volumes are small and highly variable as only two models achieved sufficiently low training loss within the allotted epochs. This can be fixed by training for additional epochs. But we note that overall our methodology is not equipped for cases where high-loss models still generalize well.} 
    \label{fig:SVHN Fashion MNIST}
\end{figure}

\section{Scaling Laws}
\label{app:scaling_laws}

In the main text, we showed the volumes of minima naturally found at different dataset sizes obeys a simple power law across the three orders of magnitude in our experiments. Here, we provide data for the phenomenon across our different models.

Note the resulting scaling laws are highly sensitive to the chosen loss thresholds.

\begin{table}[h!]
    \centering
    \begin{tabular}{l c c c c}
        \toprule
        Model & Hidden Layers & Extra Details & $\alpha$ & Test Acc (100\%/0.1\%)\\
        \midrule
        MNIST, MLP & 256, 128 & AdamW & -0.1835 & 97.7\% / 67.5\% \\
        MNIST, MLP & 256, 128 & AdamW, 50 Perturbs & -0.1892 & 97.7\% / 67.5\% \\
        MNIST, MLP & 256, 128 & AdamW, Loss 0.01 & -0.2647 & 97.7\% / 67.5\% \\
        MNIST, MLP & 256, 128 & AdamW, 5000 Epochs & -0.2116 & 97.6\% / 66.8\% \\
        MNIST, MLP & 256, 128 & AdamW, No Filternorm & -0.0863 & 97.7\% / 67.5\% \\
        MNIST, MLP (Deep) & 256, 256, 128 & AdamW & -0.1804 & 97.6\% / 66.7\% \\%& $1.19 \times 10^6$ / $8.23 \times 10^5$ \\
        MNIST, MLP (SGD) & 256, 128 & SGD &  -0.1447 & 97.9\% / 68.1\% \\%& $9.13 \times 10^5$ / $6.80 \times 10^5$ \\
        MNIST, MLP (SAM) & 256, 128 & AdamW + SAM &  -0.1485 & 97.8\% / 67.8\% \\%& $9.13 \times 10^5$ / $6.80 \times 10^5$ \\
        MNIST, CNN & see repo* & AdamW &  -0.0853 & 99.1\% / 70.5\% \\%& $1.64 \times 10^6$ / $1.42 \times 10^6$ \\
        CIFAR10, MLP & 512, 256 & AdamW &  -0.3394 & 52.8\% / 19.7\% \\
        CIFAR10, MLP & 512, 256 & AdamW, 50 Perturbs & -0.3301 & 52.8\% / 19.7\% \\
        CIFAR10, MLP & 512, 256 & AdamW, Loss 0.01 & -0.3858 & 52.8\% / 19.7\% \\
        CIFAR10, CNN & see repo* & AdamW &  -0.2339 & 75.5\% / 23.0\% \\%& $5.16 \times 10^6$ / $3.36 \times 10^6$ \\
        Fashion MNIST, MLP & 256, 128 & AdamW &  -0.3747 & 88.2\% / 63.9\% \\%& $9.34 \times 10^5$ / $3.45 \times 10^5$ \\
        SVHN, MLP & 256, 128 & AdamW &  -0.3747 & 88.2\% / 63.9\% \\
        \bottomrule
    \end{tabular}
    \caption{Model, Scaling Constant $\alpha$, Test Accuracy, hidden layer dimensions, and extra details. This is a more comprehensive version of the table \ref{tab:scaling_constant_summary}. The observed scaling constants appear sensitive to minor tweaks like the number of epochs, choice of loss threshold, and number of perturbations. For the details of the CNNs, see our repo.}
    \label{tab:scaling_constant_details}
\end{table}

The linear fits from which the scaling coefficients were derived are shown below.

\begin{figure}[h!]
    \centering
    % Include the combined figure
    \includegraphics[width=0.33\textwidth]{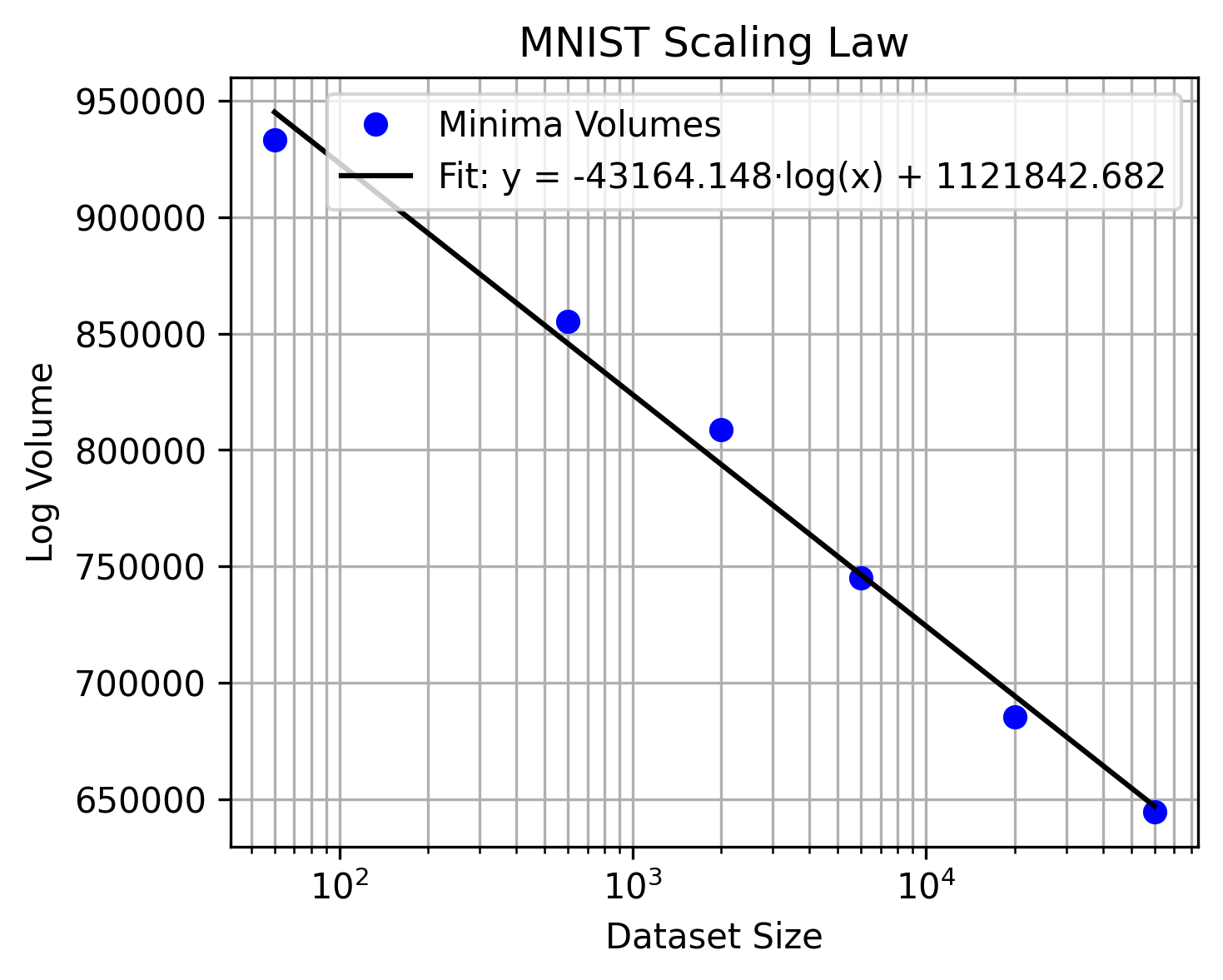}%
    \includegraphics[width=0.33\textwidth]{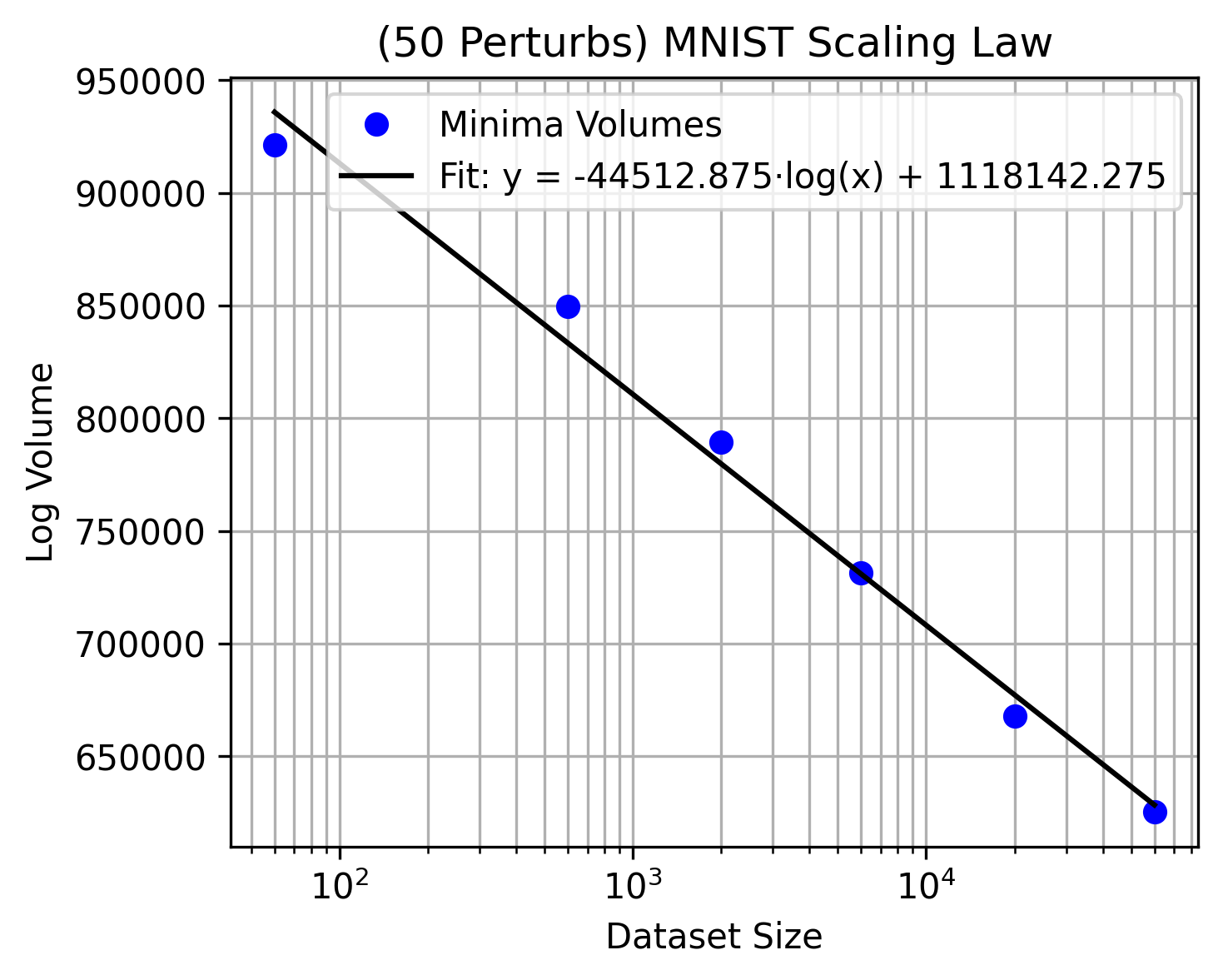}%
    \includegraphics[width=0.33\textwidth]{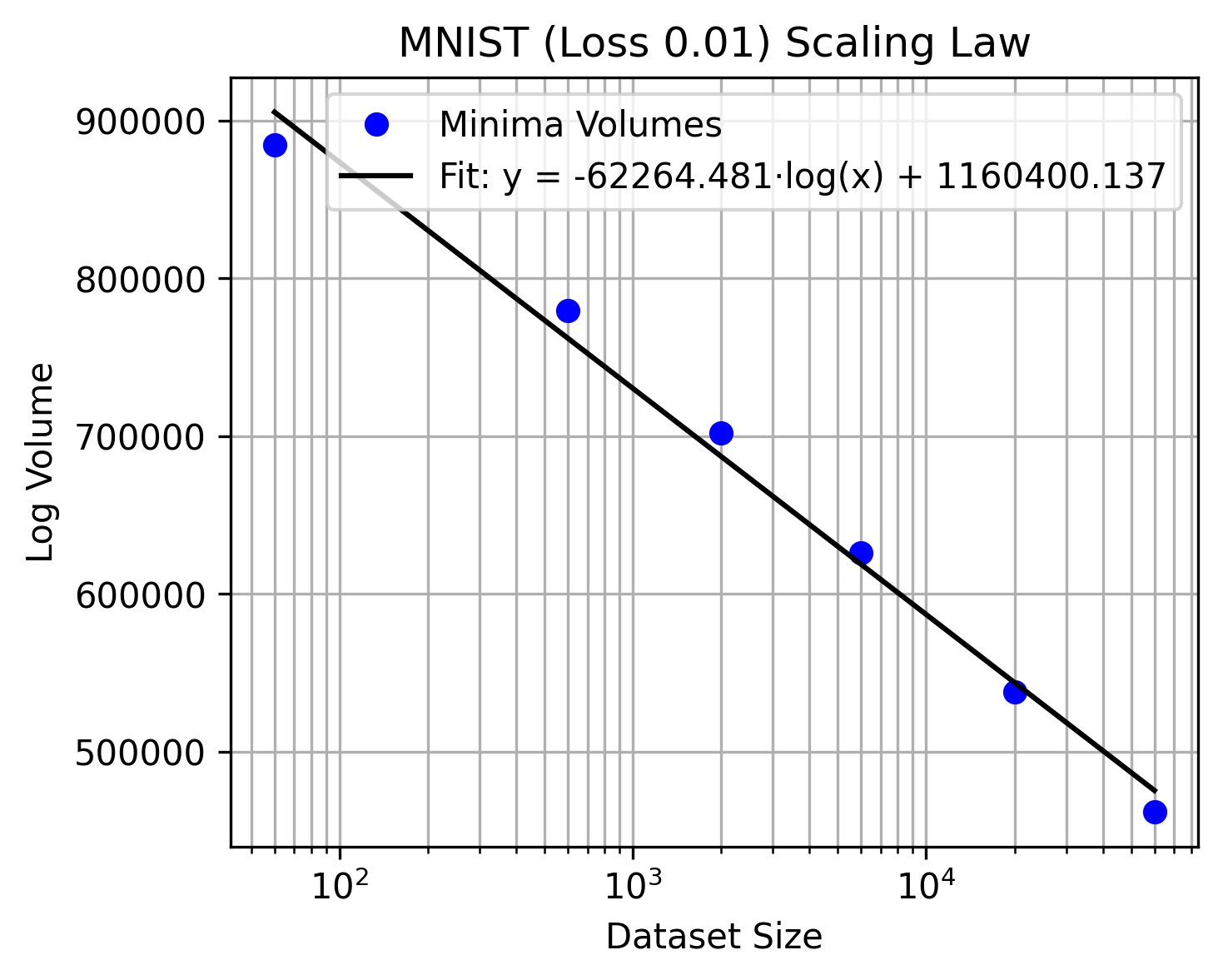}
    \includegraphics[width=0.33\textwidth]{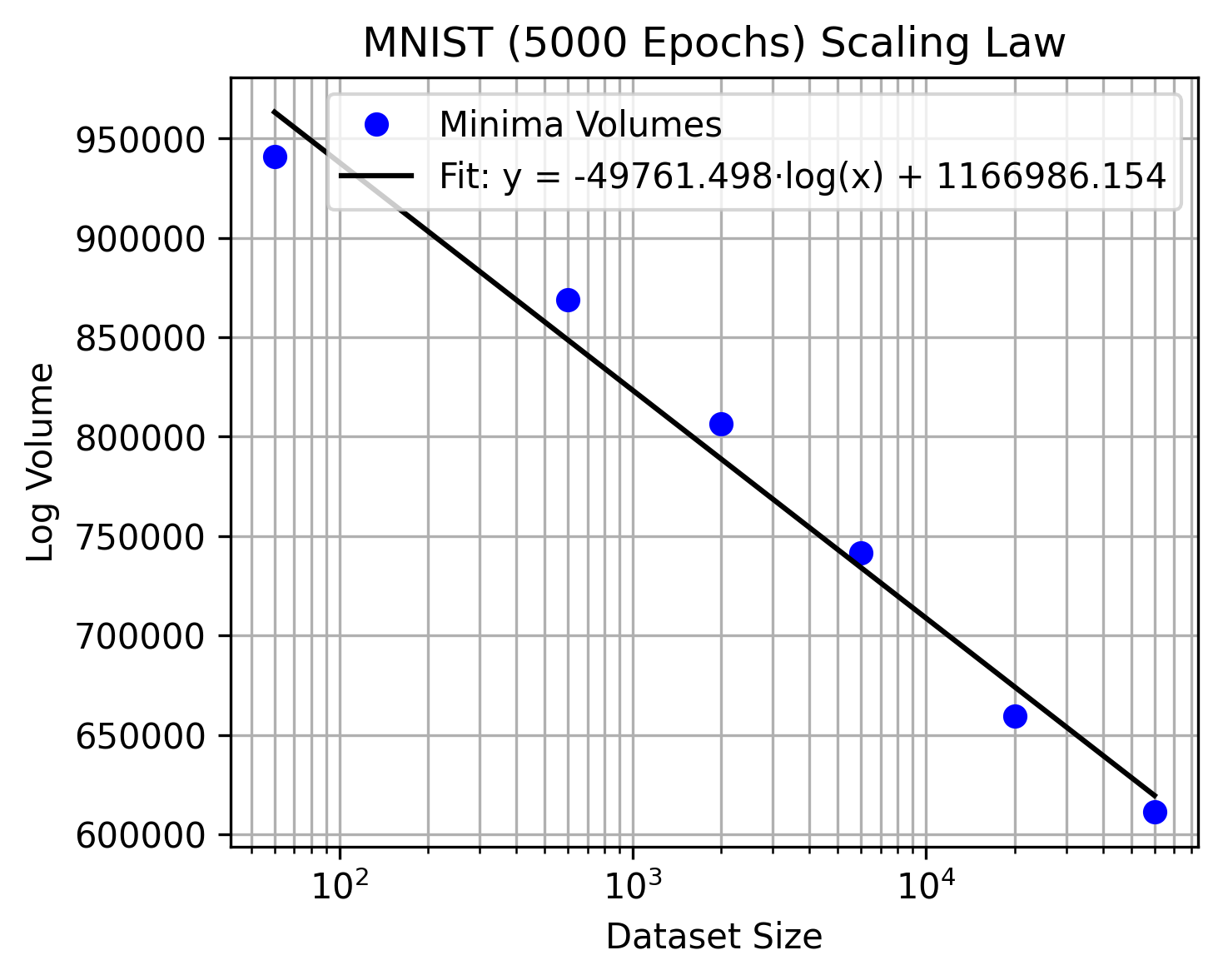}%
    \includegraphics[width=0.33\textwidth]{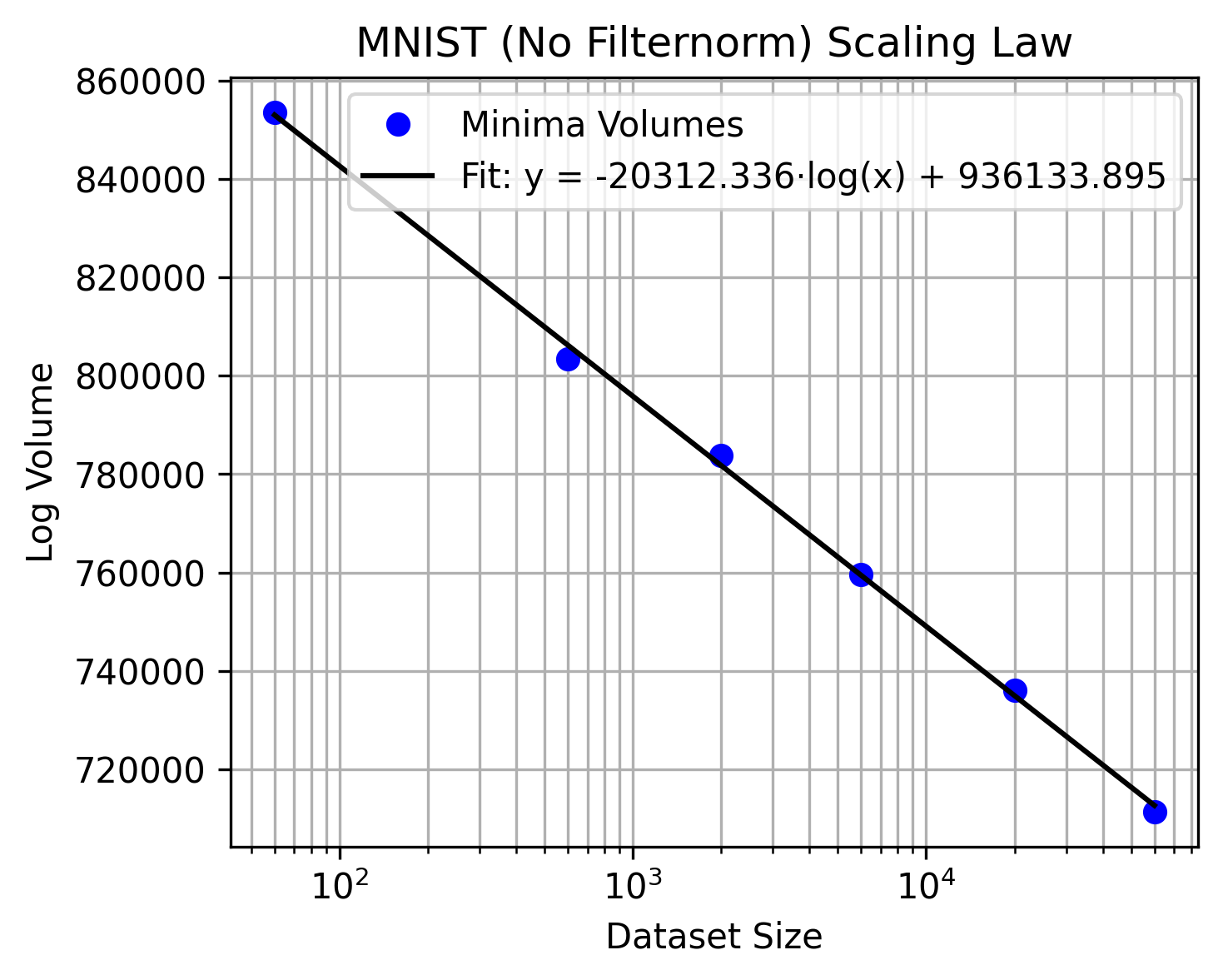}%
    \includegraphics[width=0.33\textwidth]{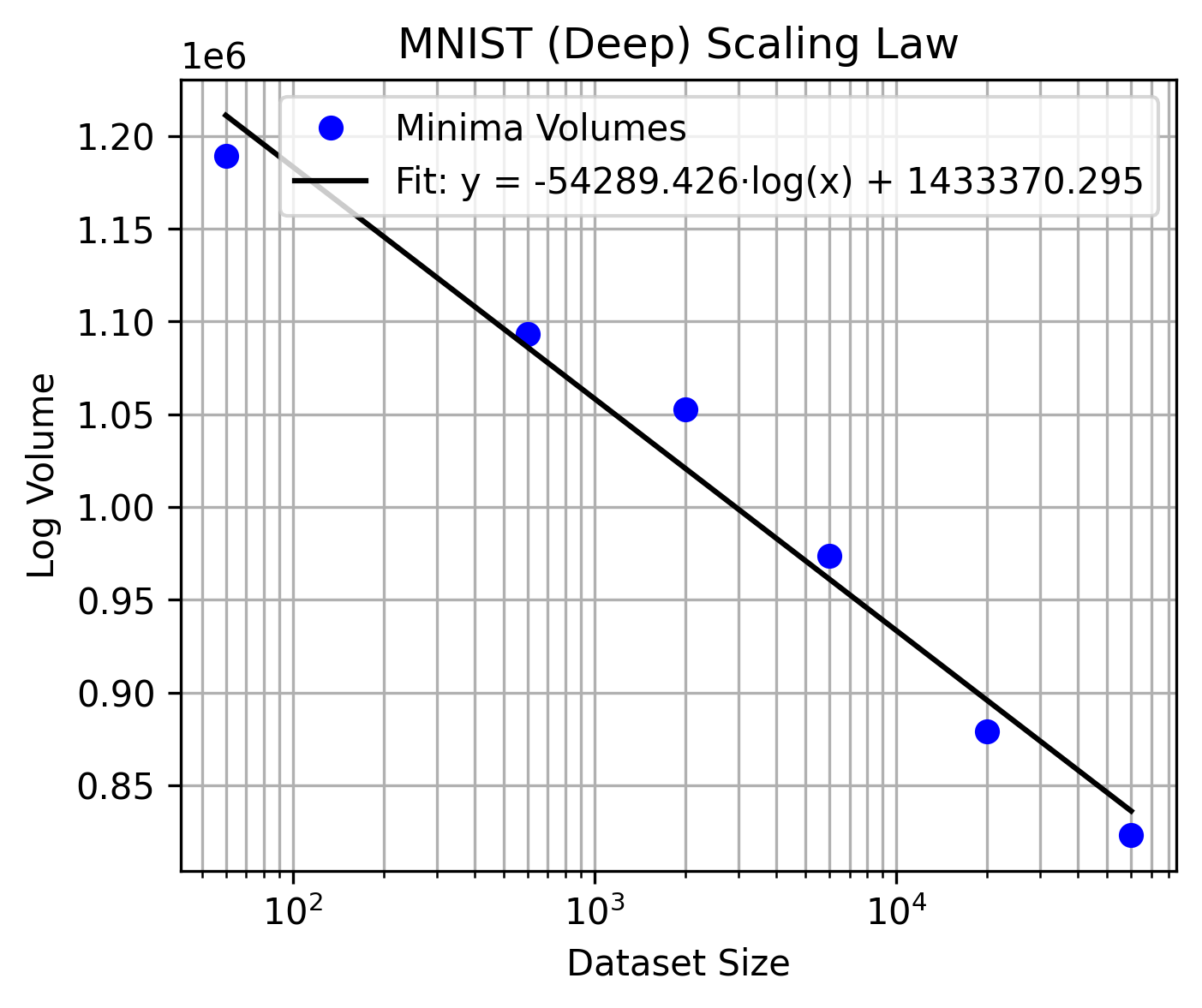}
    \includegraphics[width=0.33\textwidth]{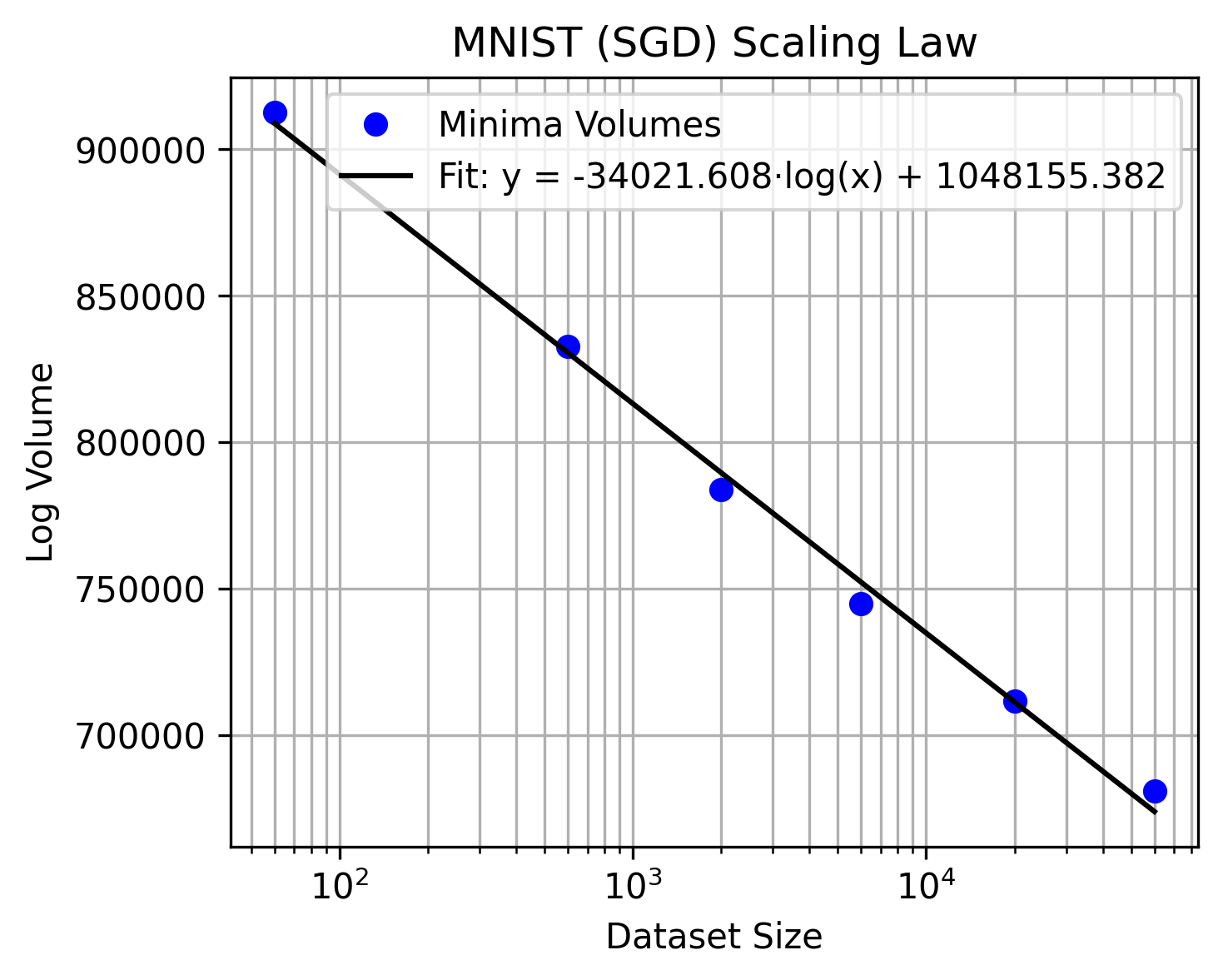}%
    \includegraphics[width=0.33\textwidth]{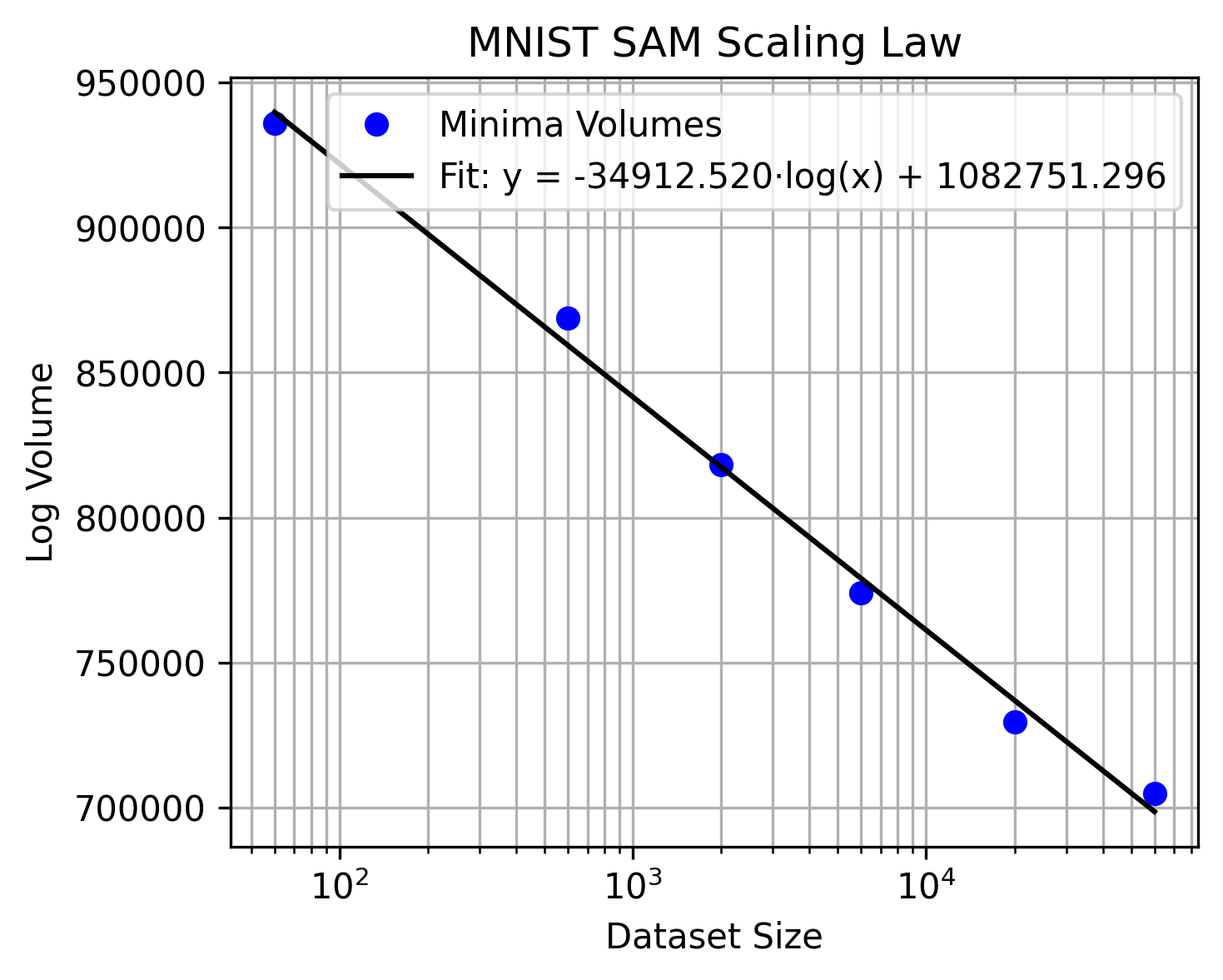}%
    \includegraphics[width=0.33\textwidth]{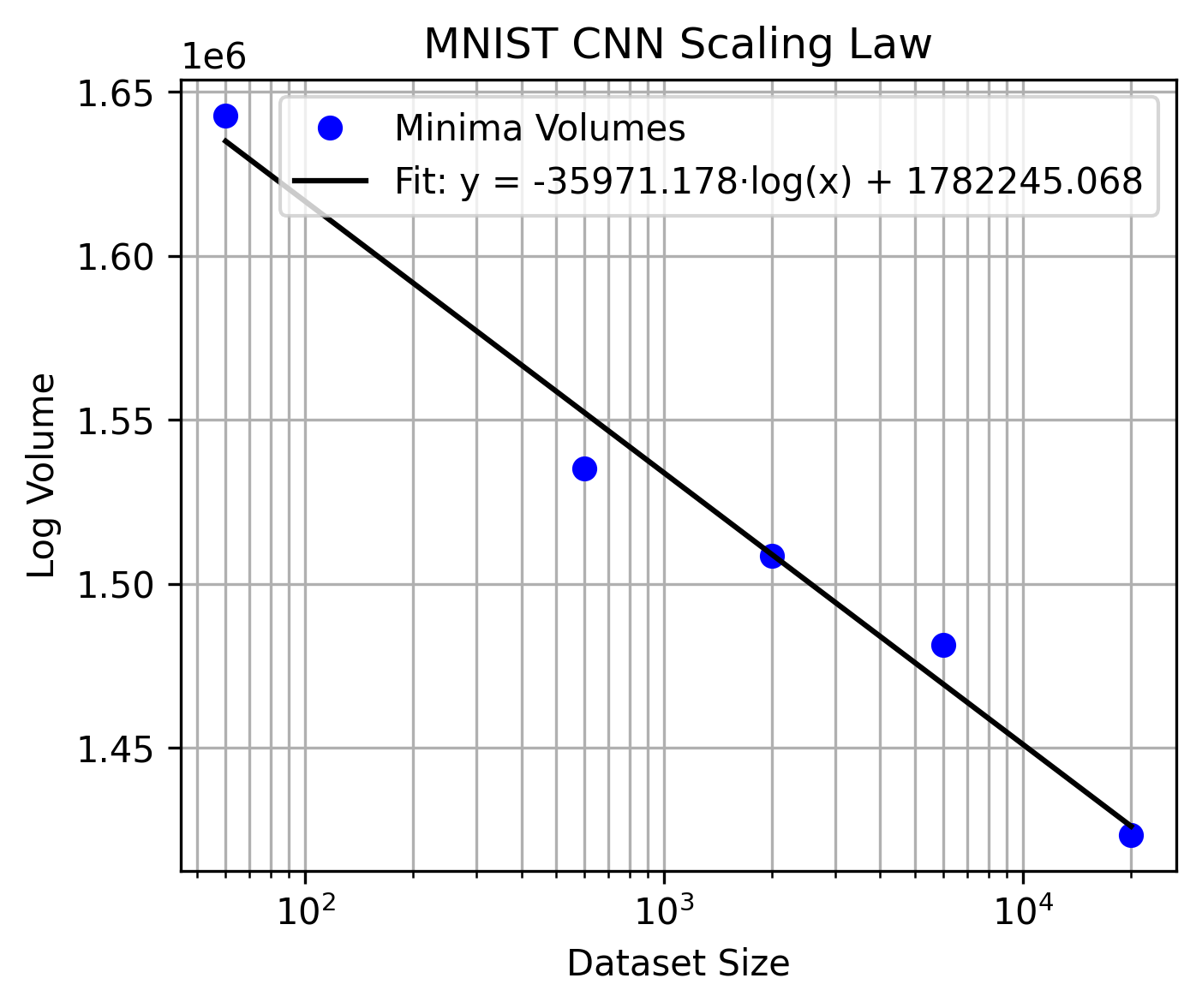}
    \includegraphics[width=0.33\textwidth]{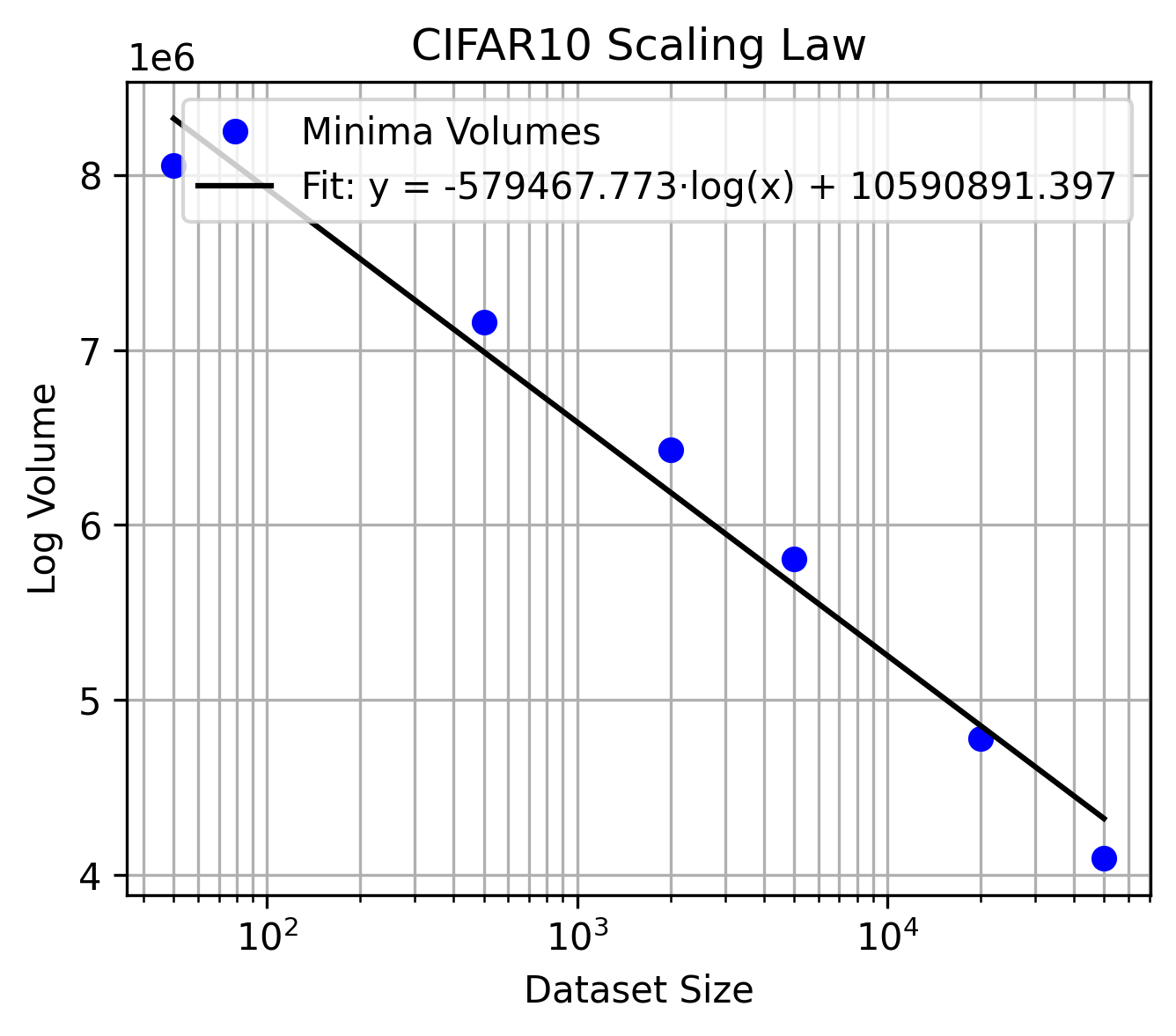}%
    \includegraphics[width=0.33\textwidth]{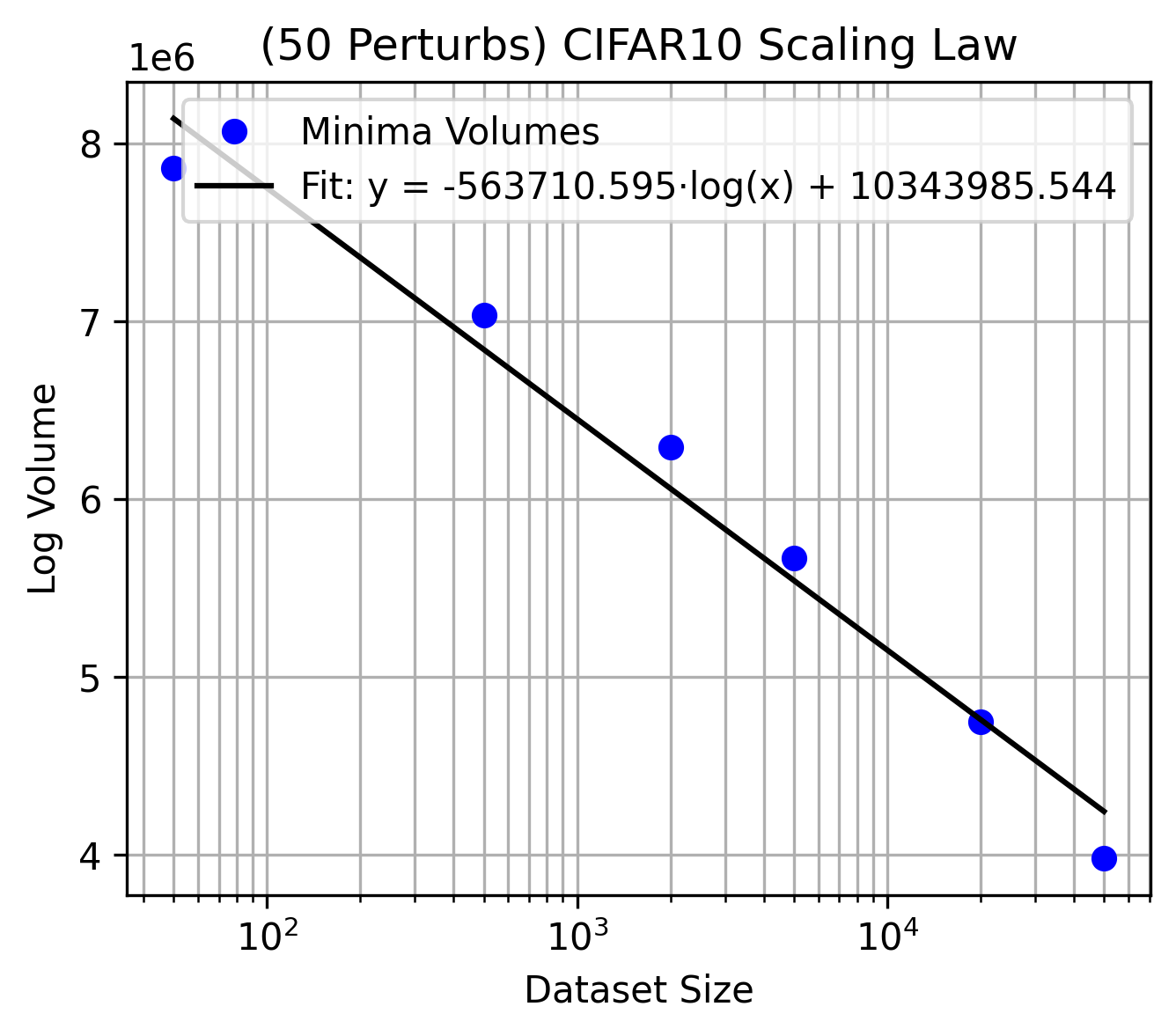}%
    \includegraphics[width=0.33\textwidth]{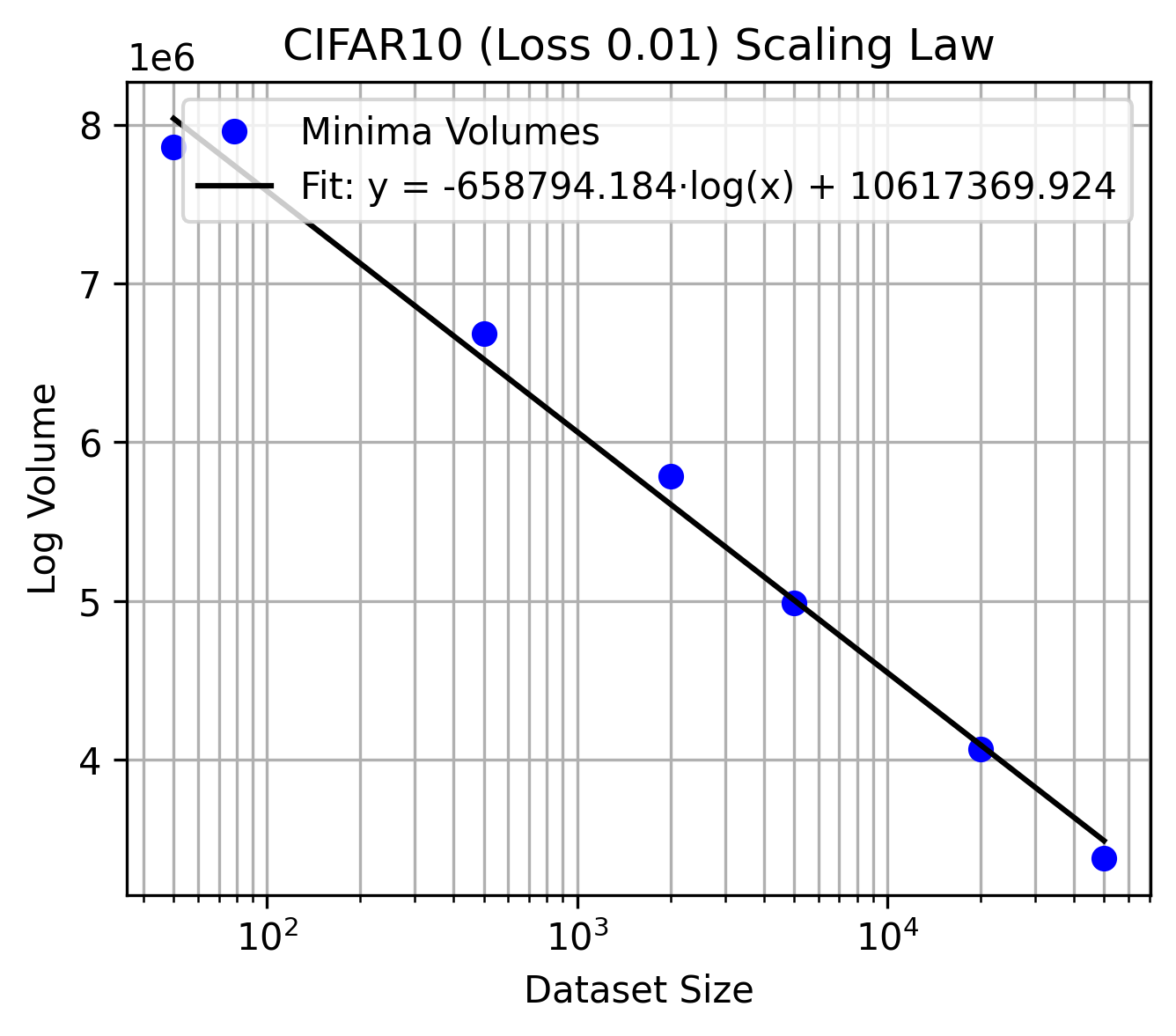}
    \includegraphics[width=0.33\textwidth]{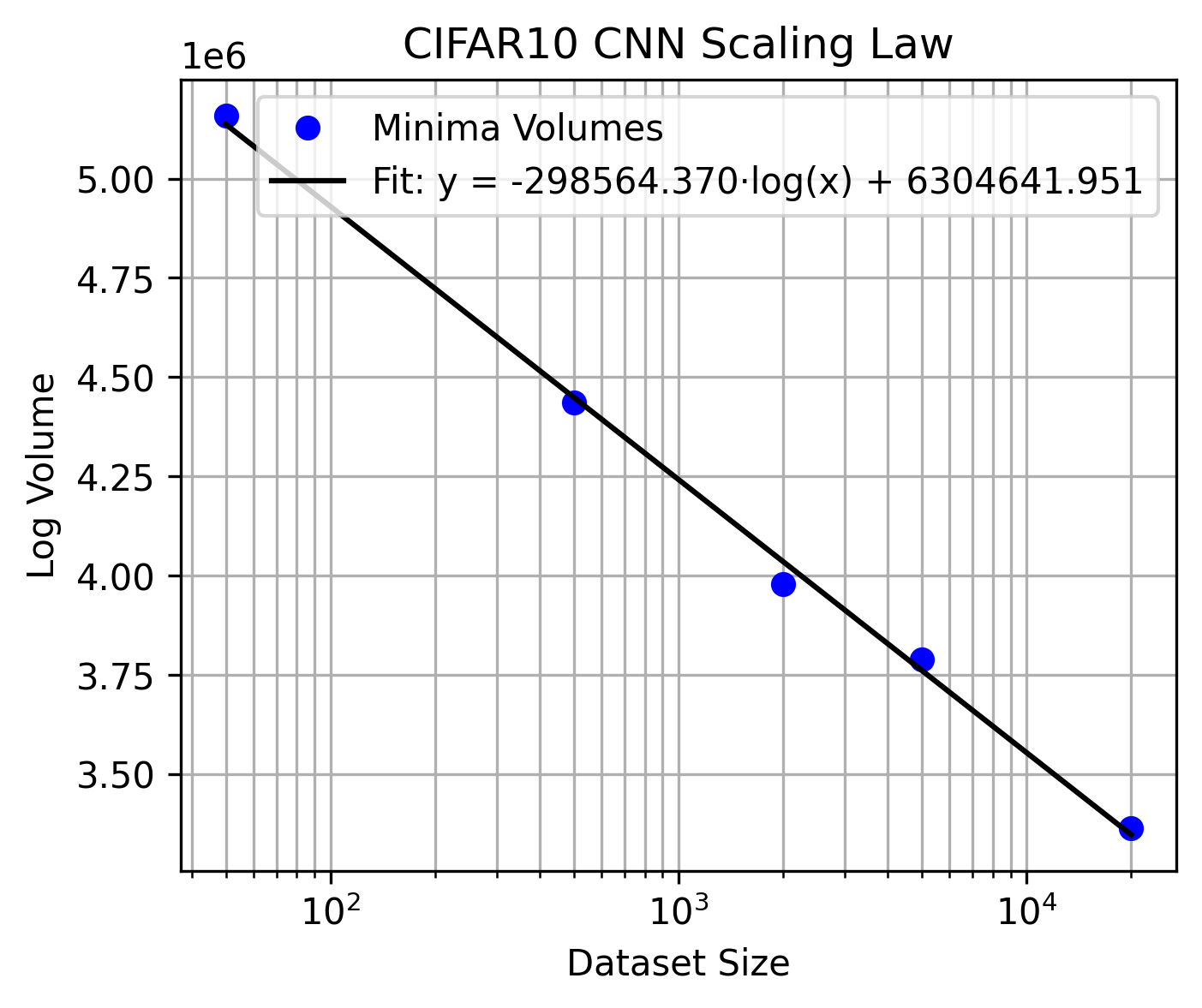}%
    \includegraphics[width=0.33\textwidth]{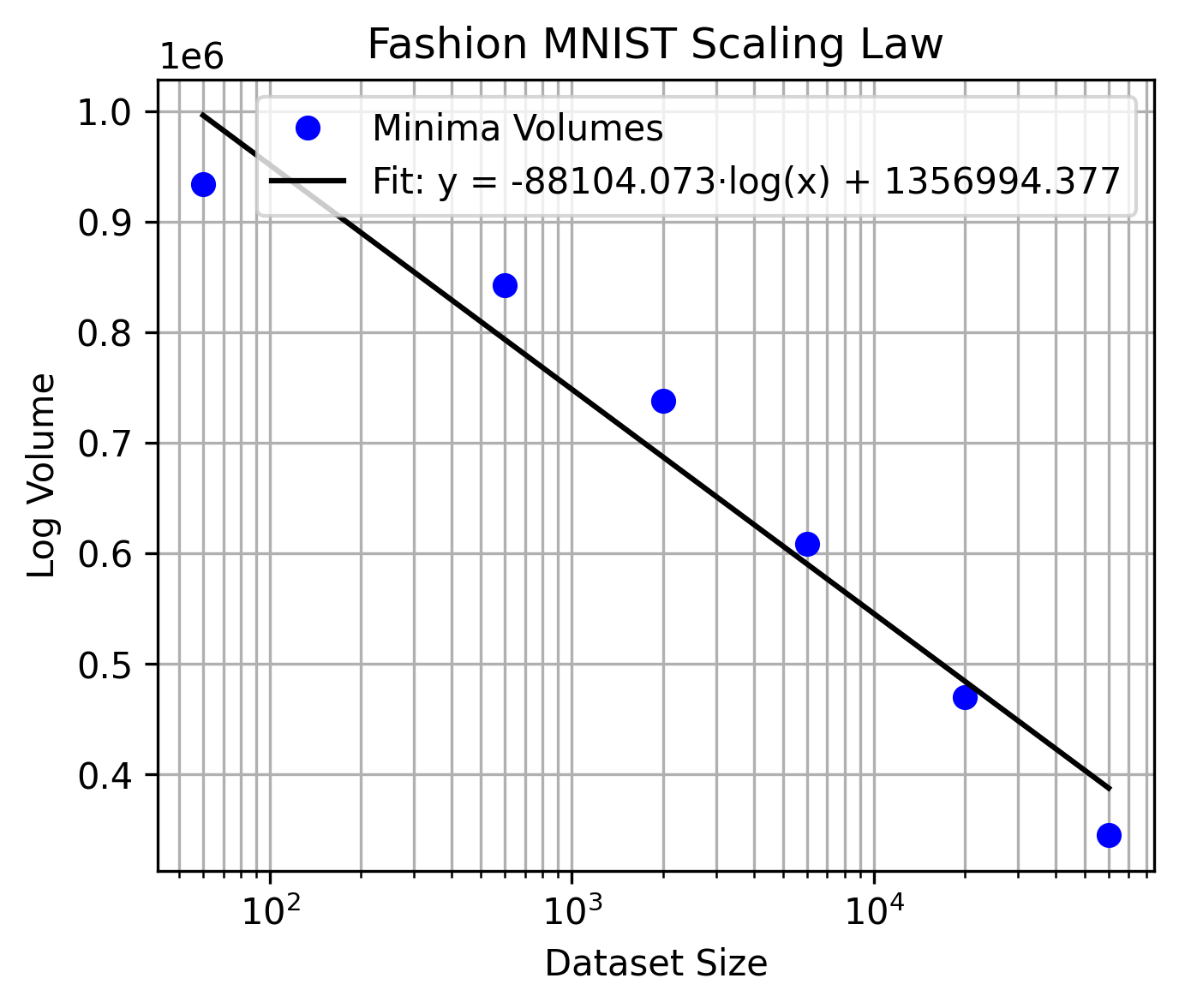}%
    \includegraphics[width=0.33\textwidth]{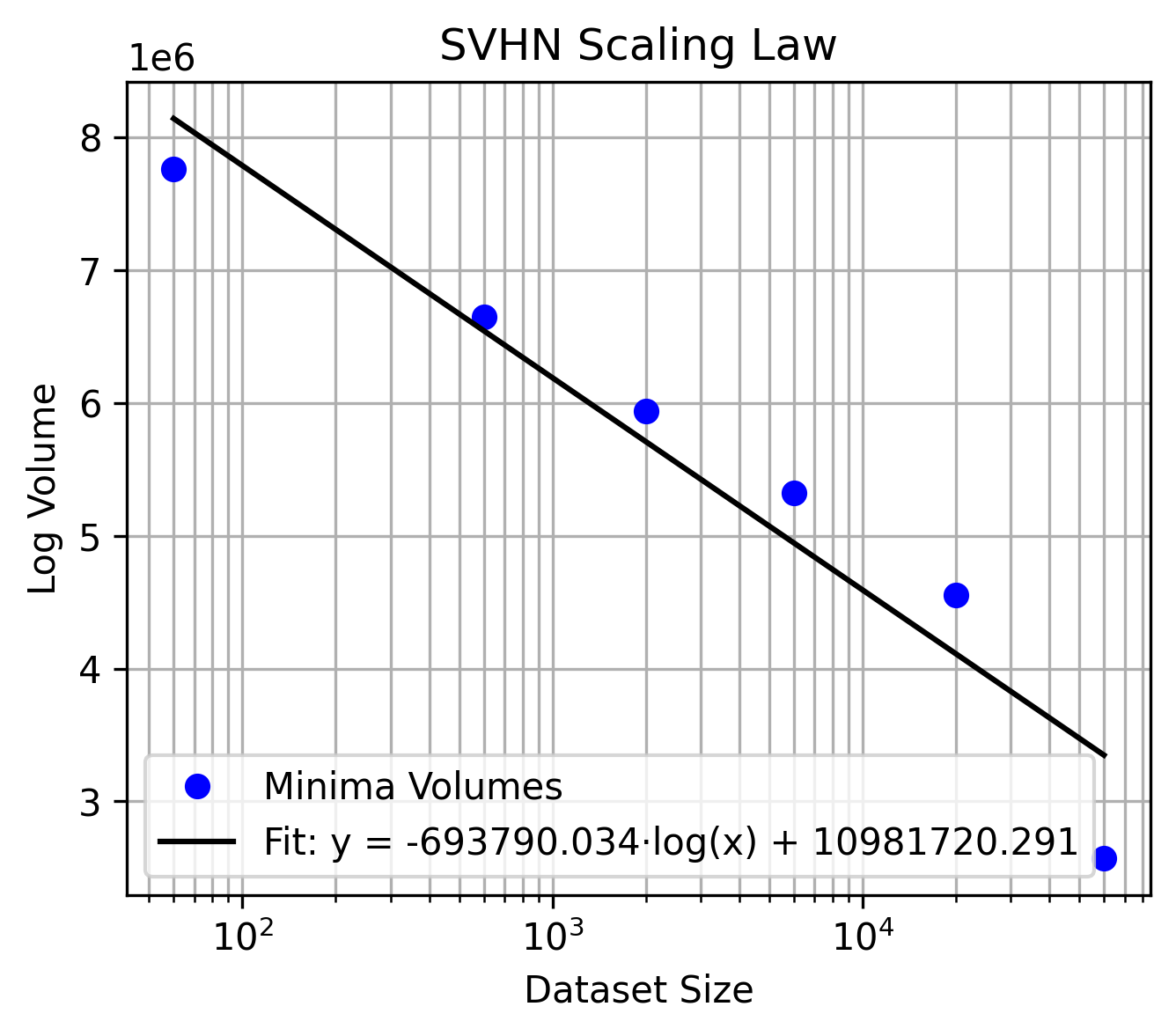}
    \caption{Linear fits of log volume to log dataset size. The resulting slope divided by the number of model parameters, yields the scaling constant.} 
    \label{fig:scaling plots}
\end{figure}   % or \include{appendix}

\end{document}